\documentclass[10pt,twocolumn,letterpaper]{article}

\usepackage{cvpr}              %

\usepackage{svg}
\usepackage[dvipsnames]{xcolor}
\usepackage{multirow} 
\usepackage{pgfplots} %
\usepackage{pgfplotstable} %
\pgfplotsset{compat=1.18}
\usepackage{etoc}

\newcommand{\ignorethis}[1]{{}}

\definecolor{green}{rgb}{0.0, 0.5, 0.0}

\newcommand{\dataset}{\emph{WikiEarth}\xspace}

\newbox\jsavebox
\newcommand{\jsubfig}[2]{%
	\sbox\jsavebox{#1}%
	\parbox[t]{\wd\jsavebox}{\centering\usebox\jsavebox\\#2}%
	}

 \newcommand{\whitetxt}[1]{{\color{white}#1}\normalfont}

 \newif\ifcomment

\usepackage{graphicx,makecell}

\newcommand{\tile}[2]{%
  \setlength{\fboxsep}{0pt}%
  \fbox{\includegraphics[width=0.13\textwidth,#1]{#2}}%
}

\definecolor{cvprblue}{rgb}{0.21,0.49,0.74}
\usepackage[pagebackref,breaklinks,colorlinks,allcolors=cvprblue]{hyperref}

\def\confName{CVPR}
\def\confYear{2026}

\title{Scene Grounding In the Wild}

\author{
Tamir Cohen$^{1}$ \quad
Leo Segre$^{1}$ \quad
Shay Shomer-Chai$^{1}$ \quad
Shai Avidan$^{1}$ \quad
Hadar Averbuch-Elor$^{2}$ \\
$^{1}$Tel Aviv University \quad 
$^{2}$Cornell University \\\\
\href{https://tau-vailab.github.io/SceneGround/}{https://tau-vailab.github.io/SceneGround/}
}

\begin{document}

\definecolor{myorange}{HTML}{f8694d}
\definecolor{mygreen}{HTML}{008000}
\definecolor{green_teaser}{rgb}{0.361, 0.839, 0.337} 
\definecolor{magenta_teaser}{rgb}{0.804, 0.2, 0.808} 

\twocolumn[{%
\renewcommand\twocolumn[1][]{#1}%
\maketitle
\jsubfig{\includegraphics[width=\textwidth, trim=4.5cm 7.4cm 4.5cm 0, clip]{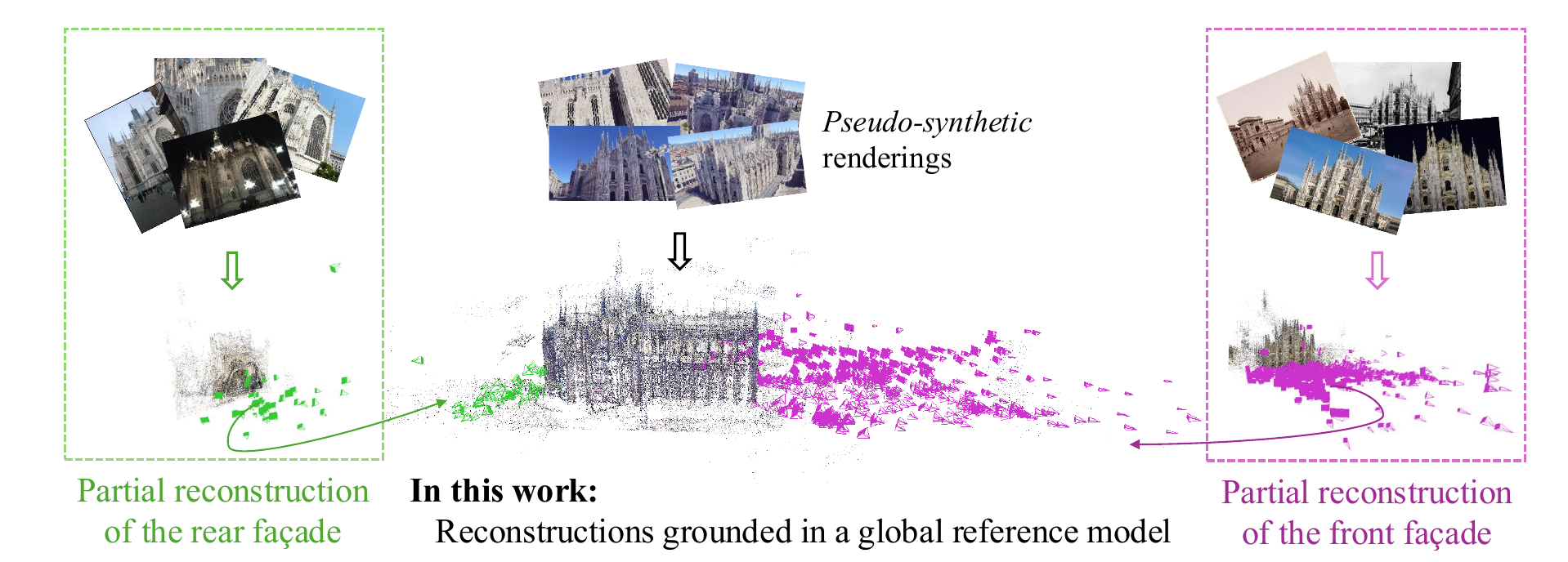}}{}
\captionof{figure}{%
Given a partial 3D reconstruction produced by running structure from motion on Internet images capturing large-scale landmarks, such as the \textcolor{green_teaser}{\bf front} or the \textcolor{magenta_teaser}{\bf rear} façade of the Milan Cathedral depicted above, we present a technique for grounding this reconstruction in a complete 3D reference model of the scene. Reference models are constructed from \emph{pseudo-synthetic} renderings extracted from Google Earth Studio. As illustrated above, our approach allows for merging partial, disjoint 3D reconstructions into a unified model. %
\vspace{5pt}
}
\label{fig:teaser}
    }]

\begin{abstract}

Reconstructing accurate 3D models of large-scale real-world scenes from unstructured, in-the-wild imagery remains a core challenge in computer vision, especially when the input views have little or no overlap. In such cases, existing reconstruction pipelines often produce multiple disconnected partial reconstructions or erroneously merge non-overlapping regions into overlapping geometry.
In this work, we propose a framework that grounds each partial reconstruction to a complete reference model of the scene, enabling globally consistent alignment even in the absence of visual overlap. 
We obtain reference models from dense, geospatially accurate pseudo-synthetic renderings derived from Google Earth Studio. These renderings provide full scene coverage but differ substantially in appearance from real-world photographs. Our key insight is that, despite this significant domain gap, both domains share the same underlying scene semantics. We represent the reference model using 3D Gaussian Splatting, augmenting each Gaussian with semantic features, and formulate alignment as an inverse feature-based optimization scheme that estimates a global 6DoF pose and scale while keeping the reference model fixed. 
Furthermore, we introduce the \dataset{} dataset, which registers existing partial 3D reconstructions with pseudo-synthetic reference models. We demonstrate that our approach consistently improves global alignment when initialized with various classical and learning-based pipelines, while mitigating failure modes of state-of-the-art end-to-end models.

\end{abstract}
    
\section{Introduction}
\label{sec:intro}

One of the grand challenges in computer vision is to reconstruct the geometry of a 3D scene from an unstructured set of photographs. Over the past several decades, remarkable progress - ranging from classical structure-from-motion (SfM) pipelines~\cite{snavely2006photo,schonberger2016structure} to modern learning-based methods~\cite{sun2022neural,wang2024dust3r,leroy2024mast3r,wang2025vggt} - has enabled increasingly accurate and detailed reconstructions, even from crowd-sourced image collections containing transient objects, varying illumination, and significant appearance changes. However, despite these advances, 3D reconstruction frameworks fundamentally rely on sufficient visual overlap between input views to establish reliable geometric correspondences. This requirement is often violated in large-scale real-world image collections, where images are heavily biased toward a sparse set of iconic viewpoints. %
For example, as illustrated in Figure~\ref{fig:teaser}, tourists photographing the Milan Cathedral overwhelmingly capture its main entrance, with a smaller set of images depicting the rear façade. Such viewpoint bias yields multiple disconnected partial reconstructions - or worse, introduces erroneous geometry by collapsing non-overlapping observations into overlapping regions. %

But what if we had access to a dense “oracle” reference model of the entire scene which could serve as a common anchor for all partial reconstructions? While crowd-sourced imagery rarely covers every region of interest, such a model can, in fact, be readily constructed from complementary sources of visual data. For example, tools like Google Earth Studio\footnote{\url{https://www.google.com/earth/studio/}} can render dense, geospatially accurate views of real-world landmarks from arbitrary camera poses, yielding complete scene coverage. However, these renderings, previously referred to as pseudo-synthetic images~\cite{vuong2025aerialmegadepth}, are generated from textured 3D meshes and therefore differ substantially in appearance from real-world crowd-sourced images. This pronounced domain gap makes it unclear how such reference models can be leveraged to align and unify partial reconstructions into a global coordinate system.

In this work, we introduce a technique that grounds partial reconstructions captured in the wild to a complete pseudo-synthetic–based reference model of the scene, effectively bridging the large appearance gap between the two domains. Our approach is motivated by the key observation that, despite substantial visual variation - both between crowd-sourced images and pseudo-synthetic renderings, and among the images themselves - all observations capture the same underlying scene \emph{semantics}. We represent the reference model using 3D Gaussian Splatting (3DGS)~\cite{kerbl20233d} and cast the alignment as an inverse optimization problem~\cite{yen2020inerf,segre2024vfnerfviewshedfieldsrigid} that estimates a global transformation for each partial reconstruction while keeping the reference model fixed. In contrast to prior work that utilize a standard photometric loss for aligning input and rendered views, we propose a semantic feature-based robust optimization scheme that operates reliably on \emph{real-world} image collections, even in the presence of outlier images that contain significant occlusions.

To evaluate our approach, we introduce the \dataset{} benchmark, pairing pseudo-synthetic reference models with thirty existing 3D reconstructions from WikiScenes~\cite{wu2021towers}. We apply our inverse optimization scheme on top of various initializations spanning classical and learning-based pipelines, and observe consistent improvements in global alignment across multiple metrics. Additionally, we assess state-of-the-art feed-forward 3D models including DUSt3R~\cite{wang2024dust3r}, MASt3R~\cite{leroy2024mast3r}, $\pi^3$~\cite{wang2025pi3} and VGGT~\cite{wang2025vggt}. Our evaluation shows that despite strong performance in other settings, on our benchmark they frequently collapse non-overlapping partial reconstructions into incorrect geometries, highlighting the need for an external reference model and our semantic-based alignment approach. Finally, we demonstrate that our approach generalizes to reference models built from additional data sources, such as unstructured frames from drone videos. %
Our contributions include:
\begin{itemize}
\item A semantic alignment framework that grounds fragmented, non-overlapping partial reconstructions to a complete pseudo-synthetic–based reference model, effectively bridging large domain gaps between real-world imagery and rendered views.
\item A robust feature–based optimization scheme, formulated as an inverse optimization problem over a 3D Gaussian Splatting representation, that replaces photometric cues with semantic features to achieve reliable alignment in challenging, in-the-wild conditions.
\item The \dataset{} benchmark, which pairs pseudo-synthetic reference models with real-world 3D reconstructions, enabling systematic evaluation of cross-domain 3D alignment techniques.
\end{itemize}

\section{Related Work}
\subsection{Sparse 3D Reconstruction} 
The goal of the sparse 3D reconstruction task is to reconstruct a scene given a sparse set of views, with little to no overlap between the images. Numerous works have been proposed for performing object-level sparse reconstruction~\cite{zhang2022relpose,fan2023pope,lin2023relpose++,wang2023posediffusion}.
Much fewer works address this problem at scene-scale. 
In particular, Chen et al.~\cite{chen2021widebaseline} propose to learn discrete distributions over the 5D pose space.

\begin{figure*}

\centering
\includegraphics[width=\textwidth]{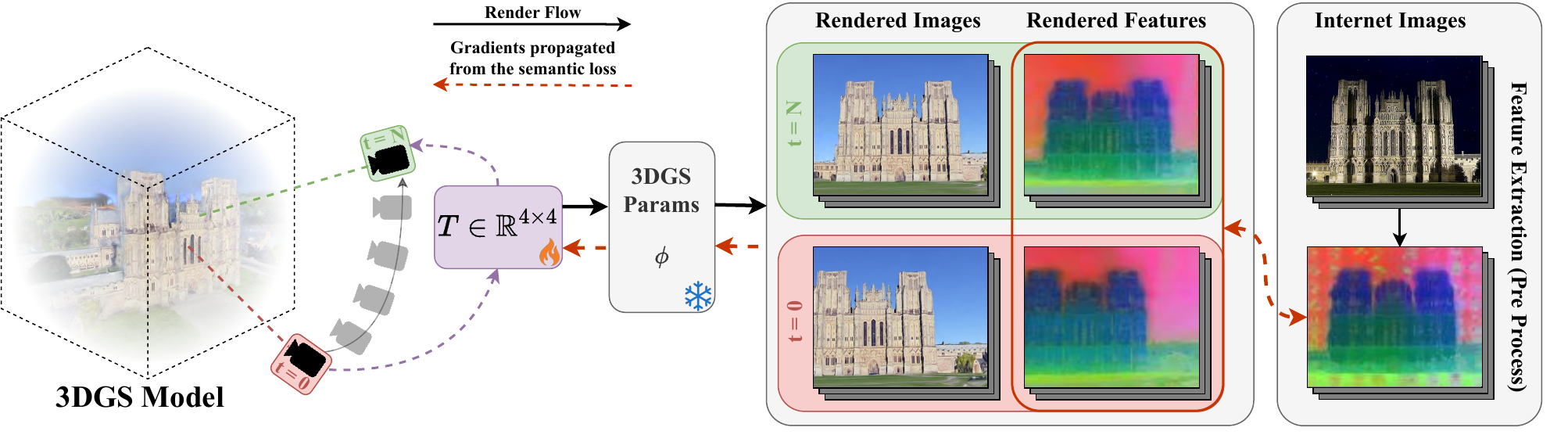}
\caption{\textbf{Scene Grounding via Semantic Feature-based Robust Optimization}. Given a 3DGS reference model ${\cal M}$ (left) and a set of Internet images ${\cal I}$ (right), we propose an inverse optimization scheme that predicts a global 6DoF+scale alignment $T$ while keeping the parameters of ${\cal M}$ fixed. We obtain an initial transformation $T$ (in red) using a traditional SfM technique. During optimization, we calculate a semantic feature loss ${L_{sem}}$ and backpropagate it to update $T$ (converging to the rendered view in green after $N$ steps). 
    }
\label{fig:method_overview}
\end{figure*}

Closely related to our problem setting, several prior work aim to reconstruct large-scale scenes that contain non-overlapping regions. Martin et al.~\cite{martin20143d} reconstruct indoor scenes given annotated floorplan maps. Cohen et al.~\cite{cohen2016indoor} propose a technique for aligning indoor and outdoor reconstruction using windows detection. By contrast to these, in our work we are interested in leveraging accessible reference model for connecting partial reconstructions obtained from large photo collections. 
Another line of work has focused on the task of predicting 3D rotations (without estimating the relative translations) between non-overlapping images~\cite{cai2021extremerotation,dekel2024estimating, bezalel2025extremerotationestimationwild}. These methods utilize only a pair of images, rather than a full collection, and haven't been explored in the context of Internet imagery.

Recently, transformer-based 3D reconstruction methods have gained popularity. A series of works including DUSt3R~\cite{wang2024dust3r}, MASt3R~\cite{leroy2024mast3r}, FASt3R~\cite{yang2025fast3r3dreconstruction1000}, and Spann3r~\cite{wang20243dreconstructionspatialmemory} have proposed using transformers to directly reconstruct sparse internet image collections. More recently, VGGT~\cite{wang2025vggt} introduced a large feed-forward transformer that can predict all key 3D attributes of a scene given input images. While these methods have achieved significant improvements in 3D reconstruction, these models are mostly trained on image sets with significant overlap, which limits their ability to handle sparse, non-overlapping collections. %
Additionally, memory constraints limit the number of input images that can be simultaneously processed. As we demonstrate in our paper, these methods cannot yet handle our challenging non-overlapping problem setting. %

\subsection{3D Scene Registration}
Registration has been studied extensively in the field of Computer Vision. Here we cover just the work most closely related to our work.
In case the two 3D scenes are represented as point clouds, then a classical algorithm such as iterative closest poin    t~\cite{icp} and its many derivatives can be used.
There are classic global methods that use 3D descriptors to aid matching~\cite{guo2016comprehensive} and then use a sparse subset of them for global alignment \cite{zhou2016fast}. When the scene is represented as a mesh, mesh-based localization methods \cite{Panek_2023_CVPR, panek2022meshlocmeshbasedvisuallocalization} perform alignment by matching sparse features. Recent works learn the alignment features utilizing deep neural networks \cite{choy2020deep, wang2019deep, hezroni2021deepbbs, zhang20233d, sarlin2021featurelearningrobustcamera}.

One method that utilizes it for NeRF registration is DReg-NeRF \cite{chen2023dregnerf}, which converts the NeRF to a voxel grid representation and trains a deep neural network for registration task. It achieves improved results over point cloud registration methods but requires a large training set.
NeRF2NeRF~\cite{nerf2nerf} registers two 3D scenes, where both scenes are represented as a NeRF. They show an improvement over point cloud registration methods, but require user input at the initialization, which is not needed in our approach. GaussReg ~\cite{chang2024gaussreg} also addresses the task of registering two 3D scene, however it operates on scenes represented as 3DGS.
Later, VF-NeRF~\cite{segre2024vfnerfviewshedfieldsrigid} extended NeRF to include Viewshed Fields, that capture visibility constraints with NeRF, and used it to register two NeRFs.

While point cloud, mesh-based and NeRF localization methods are related to our work, we address a different setting. Specifically, our approach operates within a framework that reconstructs a scene using a Gaussian Splatting base model from low-quality images.
A more closely related work is iNeRF \cite{yen2020inerf}, which registers images to a NeRF model. Their method is based on back-propagating the photometric loss through the NeRF weights to optimize the six parameters of each camera pose, defining an SE3 transformation with an exponential parameterization. By contrast, our approach optimizes a global transformation aligning a partial reconstruction obtained by a SfM technique to a reference model, represented using 3DGS. %

\subsection{Semantic 3D Neural Representations} 
With the recent rise of 3D neural representations, various work has explored the problem of embedding semantic features over these 3D representation. These embedding are primarily used for tasks like semantic segmentation \cite{Wang_2023, Zhi:etal:ICCV2021, vora2021nesfneuralsemanticfields}, object localization \cite{drost20123d, hausler1999feature, chen2020scanrefer, liu2023nerf, moreau2022lens} and object recognition \cite{hu2023nerf, guo20143d}.
For example LERF~\cite{kerr2023lerf} augment NeRFs~\cite{mildenhall2021nerf} with CLIP~\cite{radford2021learning} embeddings, predicting a semantic feature field alongside the scene's geometry.
Several recent works embed semantic features on a 3D Gaussian Splatting representation ~\cite{qin2024langsplat, shi2023languageembedded3dgaussians, zhou2024feature}. In particular, Feature 3DGS~\cite{zhou2024feature} proposed a method for distilling 3D feature fields from any 2D foundation models. They demonstrated distillation of LSeg ~\cite{li2022lseg} and SAM ~\cite{kirillov2023segment}. Several methods specifically focus on learning semantic features for large-scale scenes~\cite{dudai2024halo,krakovsky2025lang3d}, such as the ones explored in our work. 
More closely related to our setting, Pixel-Perfect SfM~\cite{pixelperfectSFM} also minimizes a feature-metric loss. They optimize it as a part of standard SfM pipeline during bundle adjustment.
By contrast, our approach uses semantic feature embeddings for aligning a 3DGS representation with a set of images. %

\section{Method}

Our goal is to globally align a set of real-world images to a reference model. We assume the images were previously bundled together using a structure-from-motion technique (\emph{e.g.}, COLMAP~\cite{colma_7780814}), and treat them as a single meta-image ${\cal I}$.
Our reference model is built from pseudo-synthetic rendered images, rendered from a mesh model such as the freely accessible\footnote{For non profit research purposes, as further detailed on their \href{https://about.google/brand-resource-center/products-and-services/geo-guidelines/}{website}.} Google Earth Studio models, which, despite their low quality, provide extensive scene coverage. Specifically, we seek the 6DoF+scale transformation $T$ to align the meta-image ${\cal I}$ with the reference model. The challenge lies in aligning these images, that are captured in uncontrolled environments with variable lighting, viewpoints, and occlusions, to a low-quality yet globally consistent reference model.

We frame the alignment problem as an inverse optimization problem as proposed in iNeRF~\cite{yen2020inerf}. However, unlike prior work, we iteratively refine a \emph{global} 6DoF+scale transformation, leveraging information from multiple views to ground the meta-image ${\cal I}$ in the reference model ${\cal M}$. Furthermore, our reference model is a 3D Gaussian Splatting (3DGS)~\cite{kerbl3Dgaussians} model. 3DGS representations achieve real-time rendering speed, significantly outperforming NeRF-based methods, and hence are much better suited for inverse optimization-based solutions. Finally, to align meta-image ${\cal I}$, composed of images captured \emph{in the wild}, with a reference model ${\cal M}$, we propose a semantic feature-based robust optimization scheme, as further detailed below.

\smallskip \noindent \textbf{Registration:}
We use an inverse-based approach for registration, optimizing the camera location based on rendering results (see \cref{fig:method_overview}). We freeze the 3DGS model parameters and introduce a 7-parameter vector: the first 6 represent rigid transformation in $SE3$, and the last parameter represents scale. Our method optimizes using semantic features as the objective function and addresses outliers with robust optimization, as detailed in the following paragraphs

\smallskip \noindent \textbf{Semantic Features:}
Prior inverse optimization-based techniques~\cite{yen2020inerf, segre2024vfnerfviewshedfieldsrigid} utilize a standard photometric loss for aligning the input images with views rendered from the neural 3D representation. In our problem setting, not only do the input images vary in appearance - for instance, due to illumination conditions and transient occlusions - but they also significantly differ from the views rendered from a (possibly) low-quality reference model. Consider the image pair in Figure \ref{fig:method_overview} (top center). A standard color loss would not provide meaningful supervision for correctly aligning the image set to the reference model. Our key insight is that the underlying \emph{semantics} is shared across the different scene observations, and thus semantic features can effectively guide the optimization procedure (Figure \ref{fig:method_overview}).  

Specifically, we distill DINOv2~\cite{oquab2024dinov2learningrobustvisual}  features on the 3DGS model, where each Gaussian has both color and feature vectors, similar to the distillation approach used in Feature 3DGS~\cite{zhou2024feature}. During optimization, we use a L1 loss on these high-dimensional features, denoted as $L_{\text{sem}}$. DINO features have been previously utilized for various tasks, such as semantic image matching and semantic scene segmentation ~\cite{oquab2024dinov2learningrobustvisual}. In our experiments, we show that they outperform other representations for our scene alignment task.

\begin{figure}
    \rotatebox{90}{\whitetxt{g}\small{ Low $ L_{\text{sem}}$ }}
    \subfloat{\includegraphics[width=1.88cm, trim=150 0 150 0, clip]{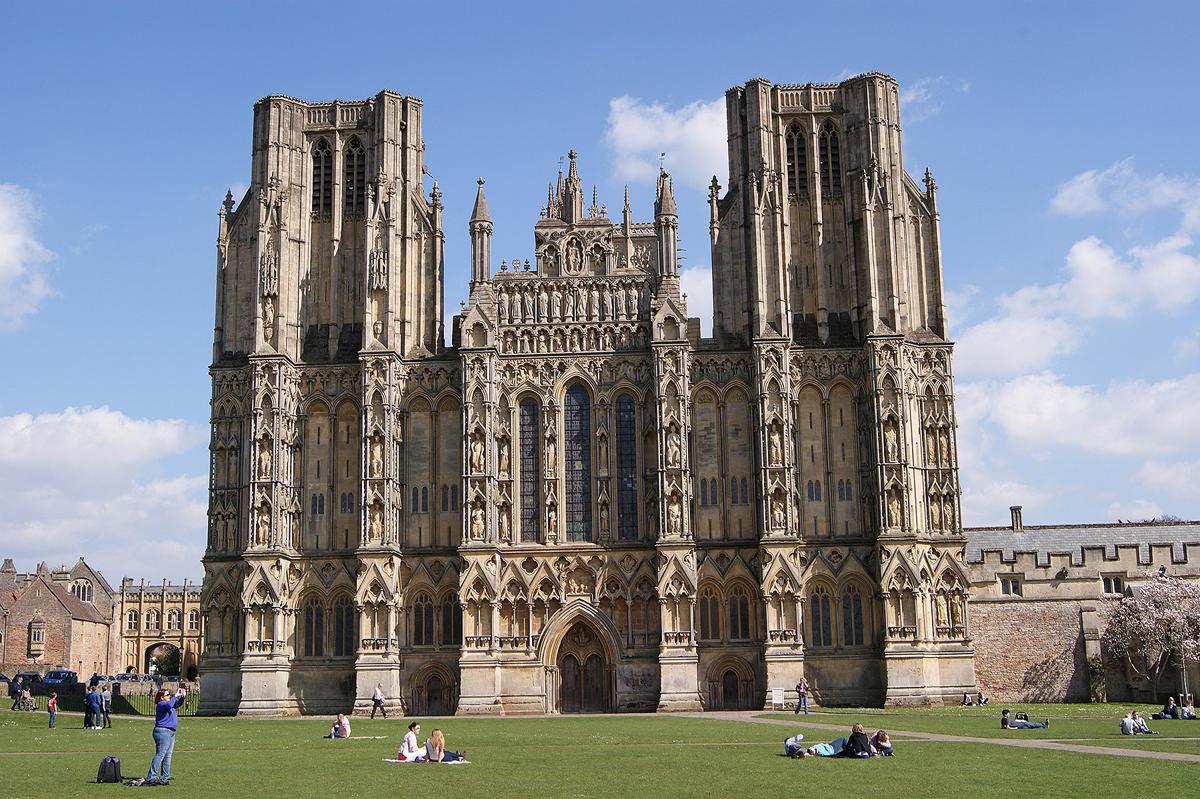}
    \includegraphics[width=1.88cm, trim=150 0 150 0, clip]{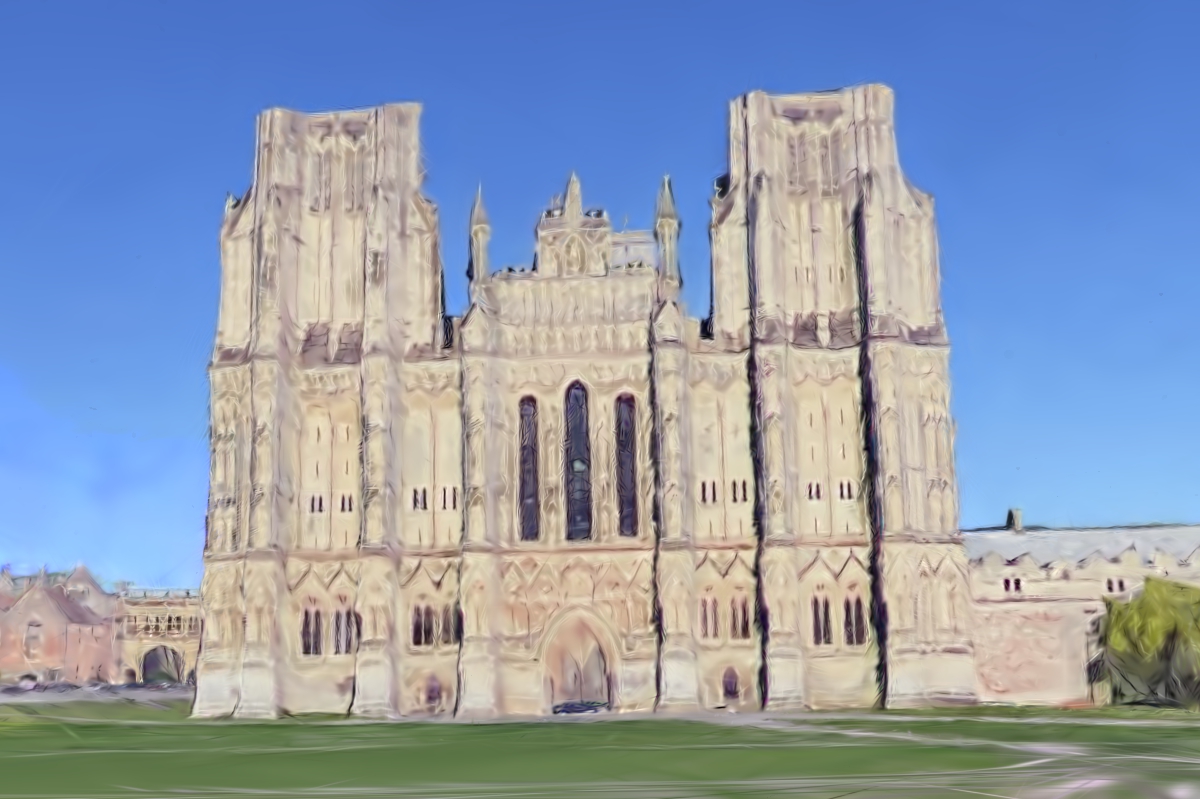}
    }
    \hfill
    \subfloat{\includegraphics[width=1.88cm, trim=80 0 80 50, clip]{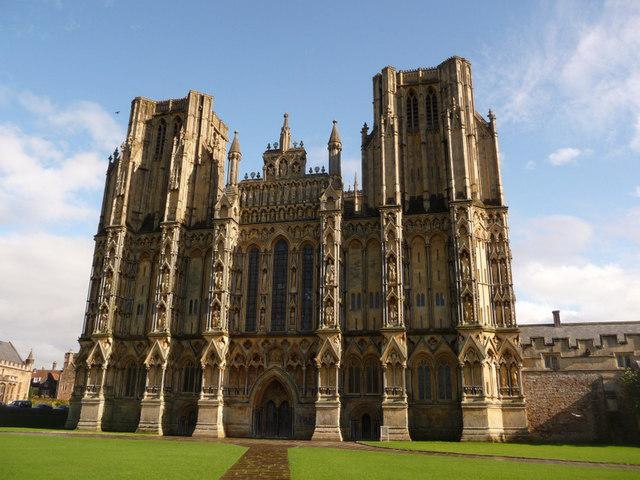}
    \includegraphics[width=1.88cm, trim=80 0 80 50, clip]{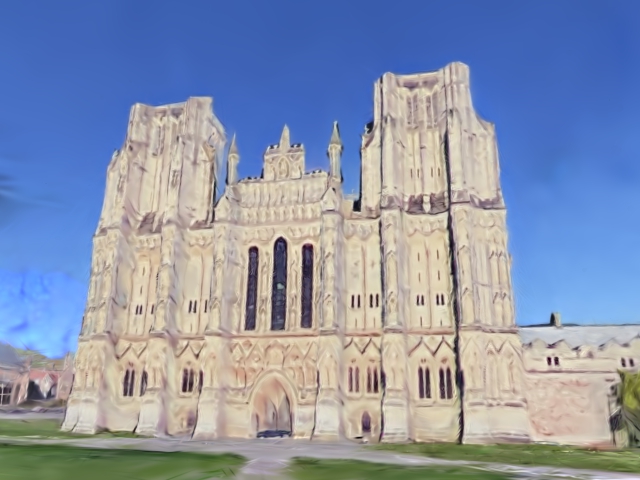}
    }
    \\
    \rotatebox{90}{\whitetxt{x}\small{ High $ L_{\text{sem}}$ }}
    \subfloat{\includegraphics[width=1.88cm, trim=50 0 50 0, clip]{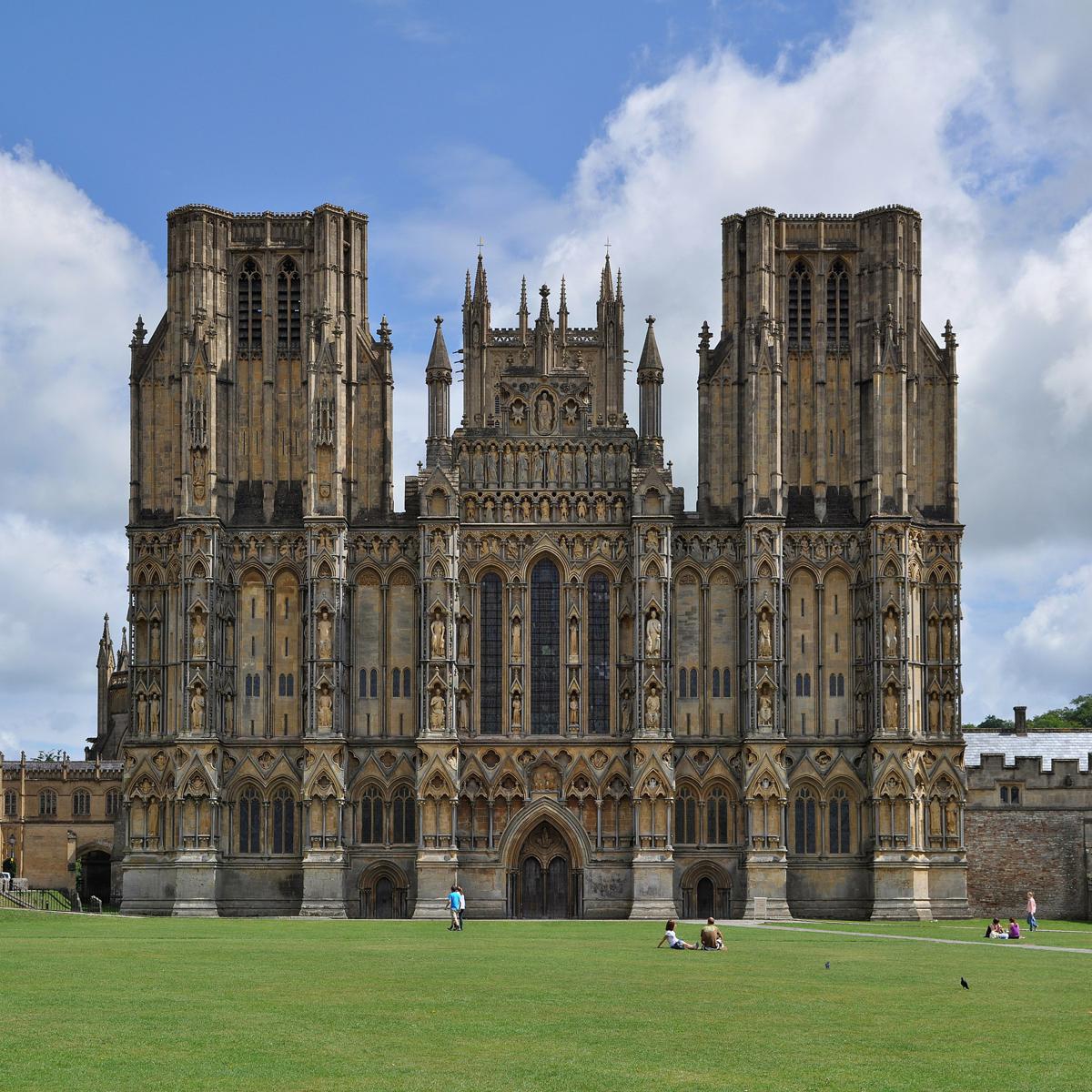}
    \includegraphics[width=1.88cm, trim=50 0 50 0, clip]{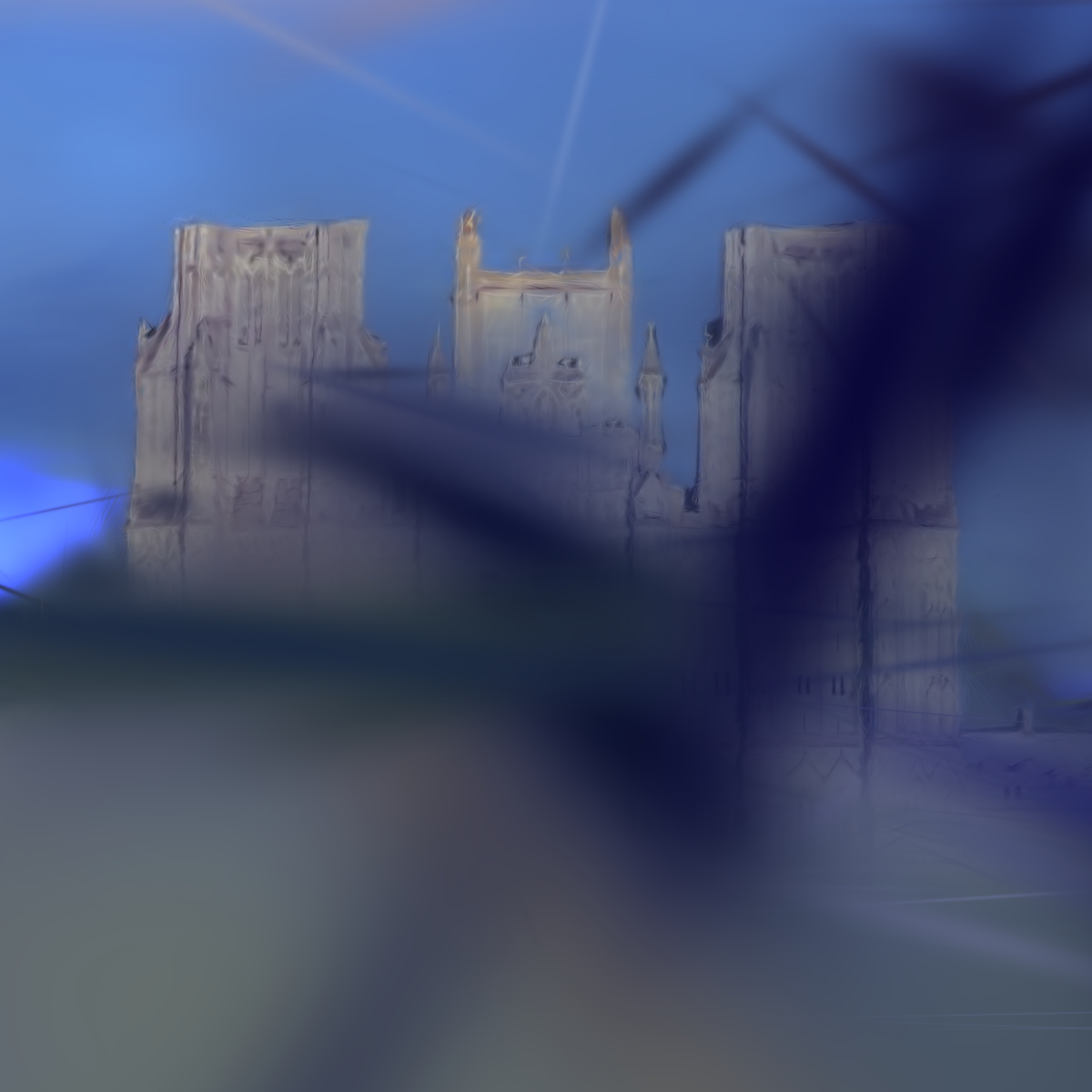}
    }
    \hfill
    \subfloat{\includegraphics[width=1.88cm, trim=0 70 0 50, clip]{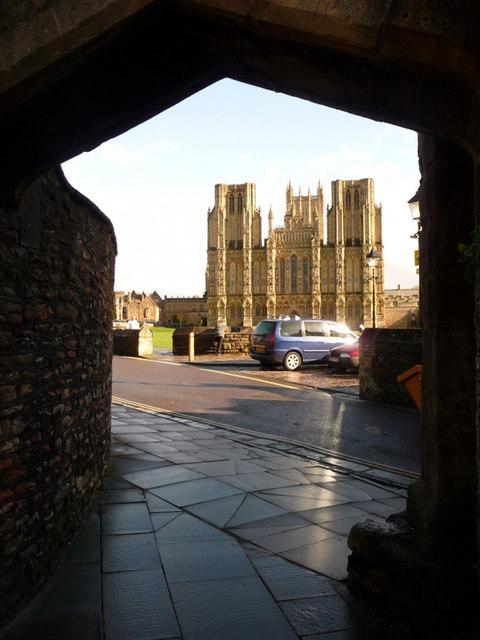}
    \includegraphics[width=1.88cm, trim=0 70 0 50, clip]{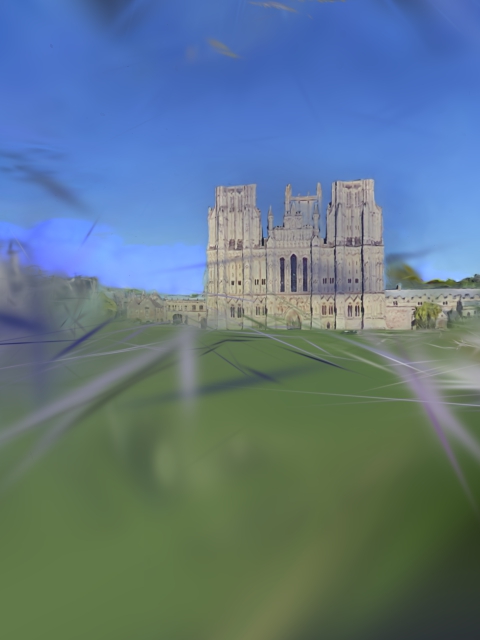}}
    \\
    \centering
    \parbox{0.23\linewidth}{\centering \whitetxt{aa} Internet}
    \parbox{0.23\linewidth}{\centering  \whitetxt{aa} Model}
    \parbox{0.23\linewidth}{\centering \whitetxt{aa} Internet}
    \parbox{0.23\linewidth}{\centering \whitetxt{aa} Model}

    \caption{
    \textbf{Challenges of aligning internet photos to       the reference model.}
    Visualization of input Internet images (first and third columns) and views rendered from the reference model at the ground-truth locations (second and fourth columns). As illustrated above, high $L_{\text{sem}}$ values (bottom row) often indicate outlier images, which our approach overcomes via a robust optimization scheme, as further detailed in Section \ref{sec:robust}.
    }
    \label{fig:outliers}
    \end{figure}

\smallskip \noindent \textbf{Robust Optimization:} 
\label{sec:robust}
Even with our semantic loss, naively optimizing the transformation $T$ between the meta-image ${\cal I}$ and the 3DGS model ${\cal M}$ fails because of outliers. See, for example, the image pairs illustrated in the bottom row of Figure \ref{fig:outliers}. As shown in the figure, rendered views may appear behind floaters in the reference model (left example), and real-world images may contain occlusions (right example). 
These outliers increase the loss and impede the optimization's convergence. 

To handle these outliers, we use robust optimization:
\begin{equation}
    {\hat T} = \arg \min_{T} \varphi( {\cal L}(T | {\cal I}, {\cal M}) ),
\end{equation}
where $\cal L$ is a summation of $L_{\text{sem}}$ values, considering all the images in $\cal I$, and $\varphi$ represents a robust loss function.  
In particular, we use the method of Least Trimmed Squares (LTS)~\cite{least_trimmed_squares}. In each optimization iteration we ignore the images with $L_{\text{sem}}$ values larger than the median loss of the previous iteration.

\label{Initialization}
\smallskip \noindent \textbf{Initialization:} We evaluate various initialization methods for global alignment between the meta-image ${\cal I}$ and the reference model ${\cal M}$. Specifically, we initialize our method with COLMAP \cite{colma_7780814}, gDLS+++ which finds the global 6Dof+scale meta-image transform by treating it as a single distributed camera \cite{sweeney2016}, and the combination of SuperPoint \cite{superpoint} as the feature extractor and LightGlue \cite{light_glue} as the feature matcher (henceforth denoted as SP+LG). Additional details for these initialization methods are provided in the supplementary material.

\smallskip \noindent \textbf{Global Alignment and Model Assembly:}
Our proposed semantic feature-based optimization scheme is applied iteratively across all meta-images, gradually aligning each to the reference model.
Each time, we independently align one meta-image to the model. 
This "puzzle-solving" process results in a cohesive, large-scale scene model that combines the partial reconstructions from each meta-image into a unified whole, overcoming limitations seen in traditional SfM methods that produce disjoint or incomplete reconstructions (as illustrated in Figure \ref{fig:teaser}).

\begin{figure}
    \centering
    \jsubfig{\setlength{\fboxsep}{0pt} 
    \fbox{\includegraphics[width=0.48\linewidth, trim=300 0 100 0, clip]{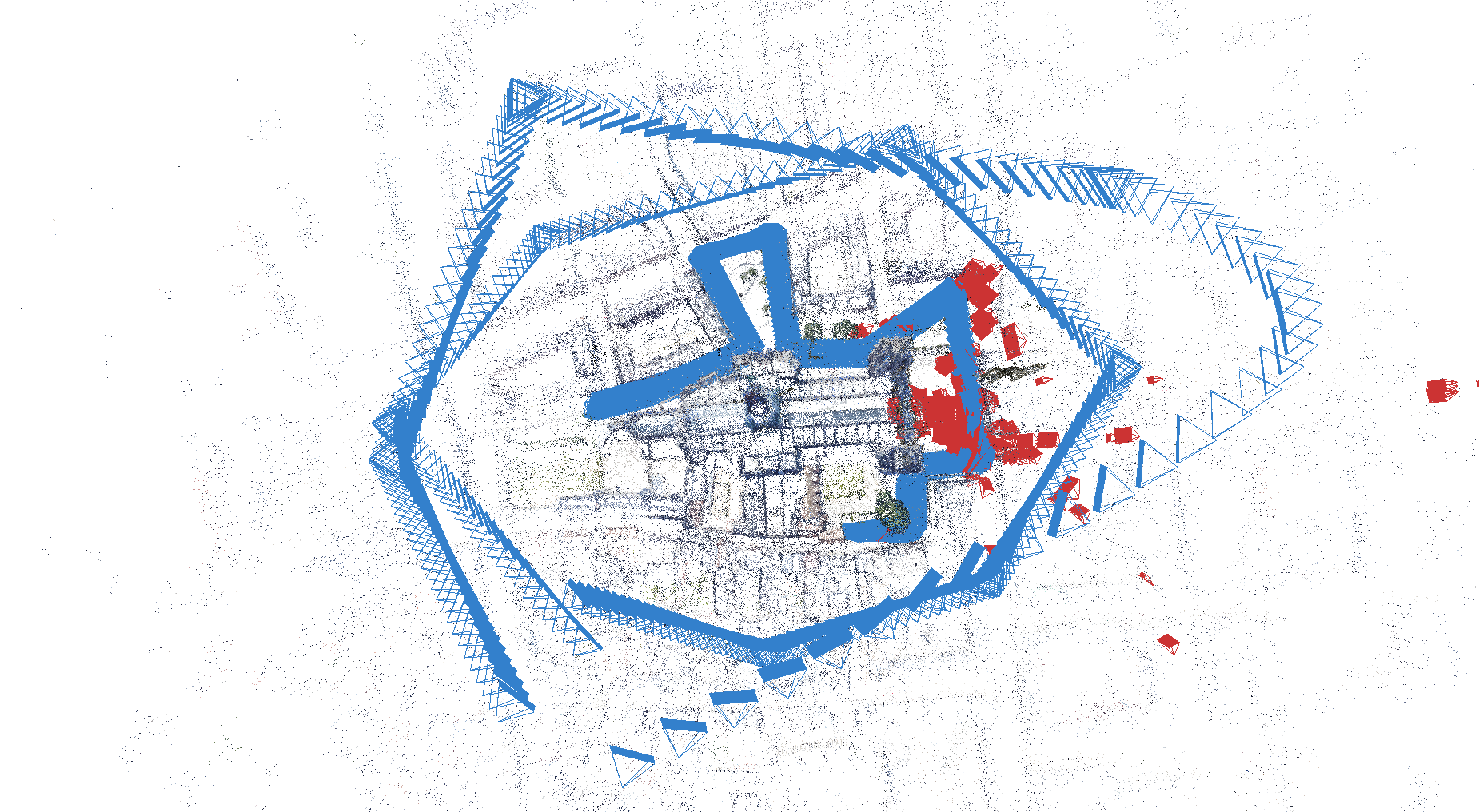}}}{Rouen Cathedral}
    \jsubfig{\setlength{\fboxsep}{0pt}
    \fbox{\includegraphics[width=0.48\linewidth, trim=200 140 400 0, clip]{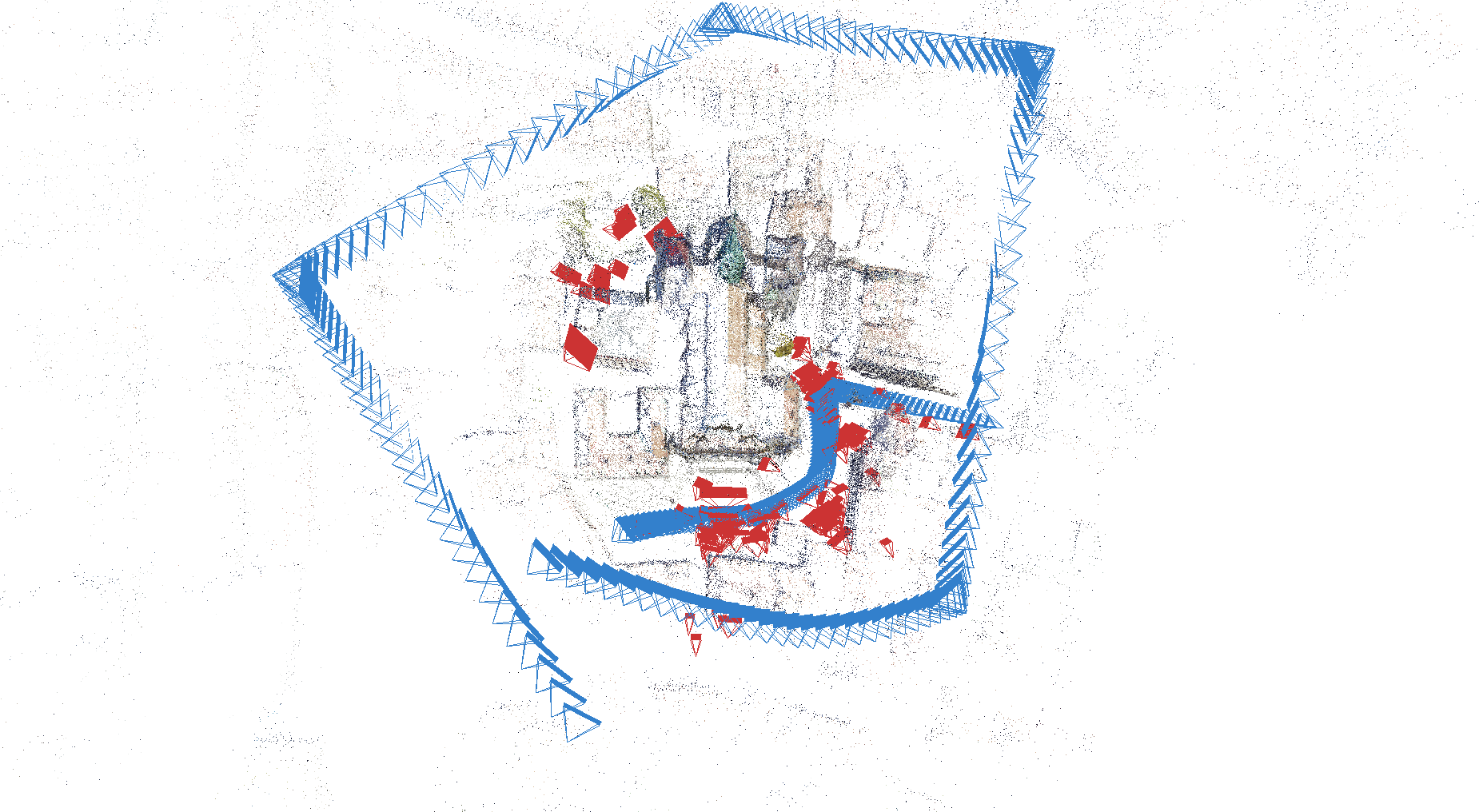}}}{Geneva Cathedral}
    \vspace{0.2cm}
    \\
    \jsubfig{\setlength{\fboxsep}{0pt}
    \fbox{\includegraphics[width=0.48\linewidth, trim=250 0 400 175, clip]{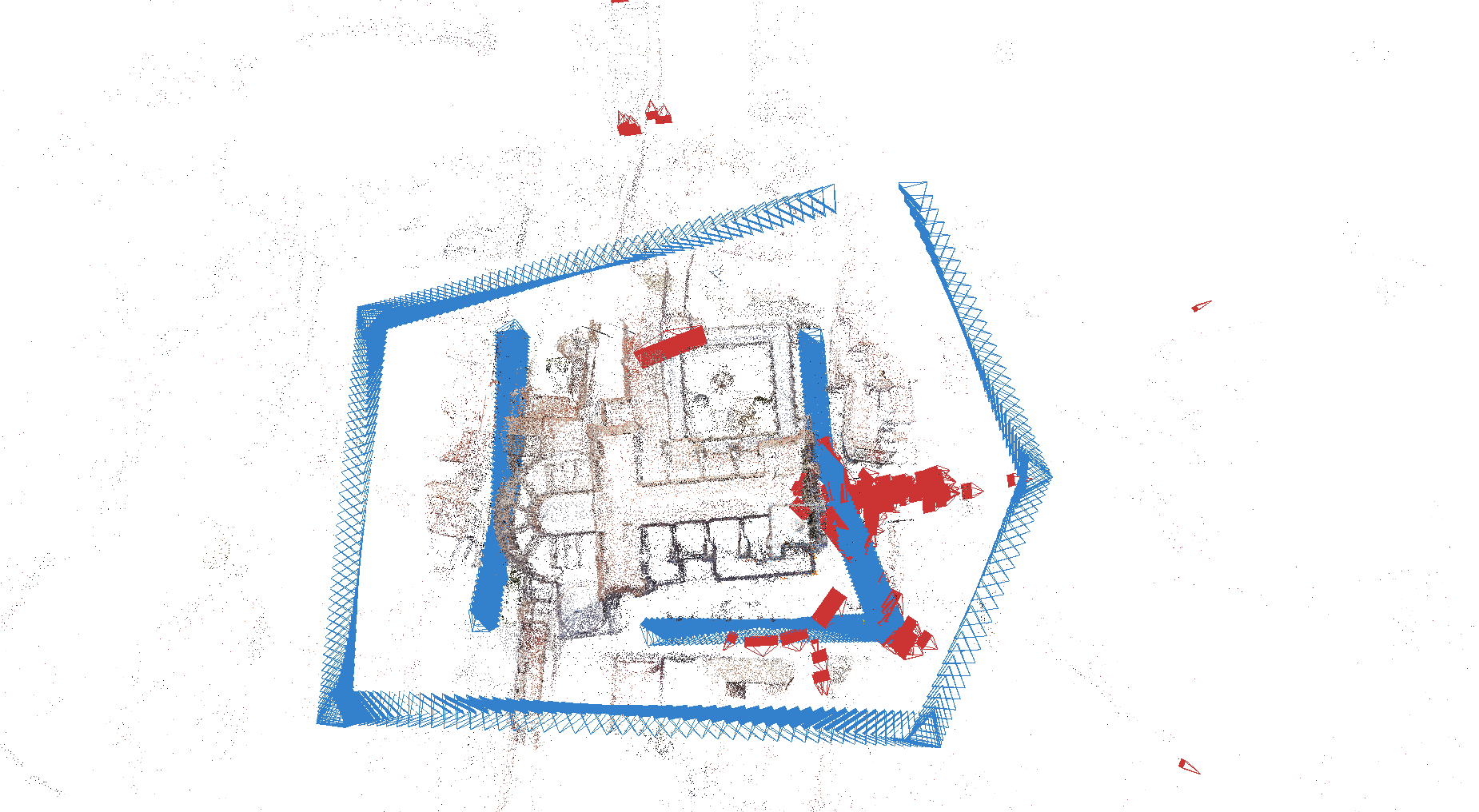}}}{Ávila Cathedral}
    \jsubfig{\setlength{\fboxsep}{0pt}
    \fbox{\includegraphics[width=0.48\linewidth, trim=150 0 250 0, clip]{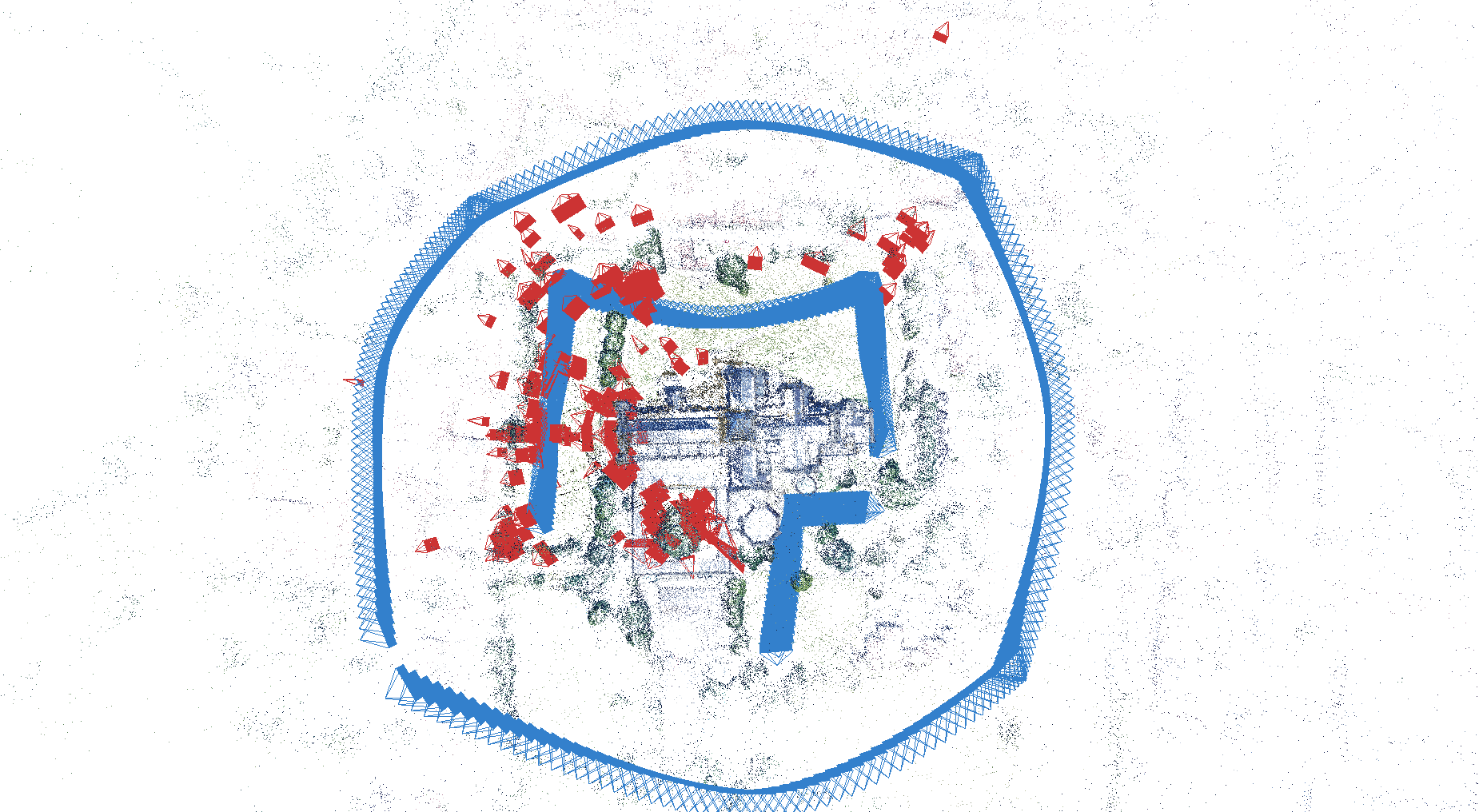}}}{Salisbury Cathedral}
\caption{\textbf{The \dataset Benchmark}. Reconstruction of four landmarks from \dataset. The blue frustums depicts the rendered images from Google Earth Studio, and the red frustums the images from WikiScenes.}
\label{fig:ground_truth_vis}
\end{figure}

\section{The \dataset{} Benchmark}
To evaluate methods for grounding partial 3D reconstructions to complete scene models, we introduce the \dataset{} benchmark. While large collections of partial in-the-wild reconstructions exist (e.g., WikiScenes~\cite{wu2021towers}, MegaDepth~\cite{li2018megadepth}, MegaScenes~\cite{tung2024megascenes}), and many landmarks also have 3D models, there is no standardized dataset that provides ground-truth correspondences or alignments between the two. Moreover, existing in-the-wild reconstructions typically cover only limited portions of a scene, making accurate evaluation particularly challenging. Our benchmark fills this gap by supplying precise alignments between partial reconstructions and full scene geometry, enabling fair and reproducible quantitative comparison across alignment methods.

We augment 3D reconstructions from the WikiScenes dataset (meta-images) with reference models derived from Google Earth Studio. Google Earth studio is an animation tool for Google Earth’s satellite and 3D imagery, which can be used for research purposes.

To create the reference model we first render camera trajectories from Google Earth Studio, while ensuring sufficient scene coverage. To achieve this, we render both aerial and ground level camera trajectories. An example of these camera trajectories is shown in Figure \ref{fig:ground_truth_vis}.
The reference model is constructed by applying COLMAP to the rendered images of the landmark from Google Earth Studio, while utilizing the GPS coordinates of the rendered images.

We create ground-truth alignments between the WikiScenes meta-images and the Google Earth reference models through a fully supervised process. Specifically, we apply COLMAP with manually selected parameters to obtain an initial registration (exact parameters per scene are listed in the supplementary). We then visually inspect and filter out any misaligned images, retaining only those with high-quality alignment. A meta-image is included in our benchmark only if at least four of its images are successfully aligned to the reference model. Sample alignments are shown in the supplementary material.

The resulting \dataset{} benchmark consists of 32 different meta-images across 23 scenes from the WikiScenes dataset. On average, each meta-image contains 97 images, with a maximum of 713 and a minimum of 8. Additional statistics are available in the supplementary.

Figure \ref{fig:ground_truth_vis} illustrates the reconstructions within the \dataset{} benchmark, where the blue frustums depict images rendered from Google Earth Studio and the red frustums Internet images from WikiScenes.

\section{Experiments}
\label{sec:results}
In this section, we present our main results and comparisons. We compare performance to the COLMAP~\cite{colma_7780814}, GDLS+++~\cite{sweeney2016} and SP+LG ~\cite{superpoint,light_glue} baselines in Section \ref{sec:eval}. 
These baselines also serve as our initializations, as further detailed in Section \ref{Initialization}. Additionally, we compare against recent feed-forward 3D models (specifically DUSt3R~\cite{wang2024dust3r}, MASt3R~\cite{leroy2024mast3r}, $\pi^3$~\cite{wang2025pi3}, and VGGT~\cite{wang2025vggt}) in Section \ref{sec:mastr}. Implementation details, additional experiments and interactive visualizations are provided in the supplementary material. %

\begin{figure}[t]
  \centering
  \setlength{\tabcolsep}{1pt}  %
  \renewcommand{\arraystretch}{0.1}

 \resizebox{\columnwidth}{!}{%
  \begin{tabular}{c c | c c | c c}
  COLMAP &  + Ours & gDLS+++ &
   + Ours &
  SP+LG &
   + Ours \\[4pt]

    \tile{}{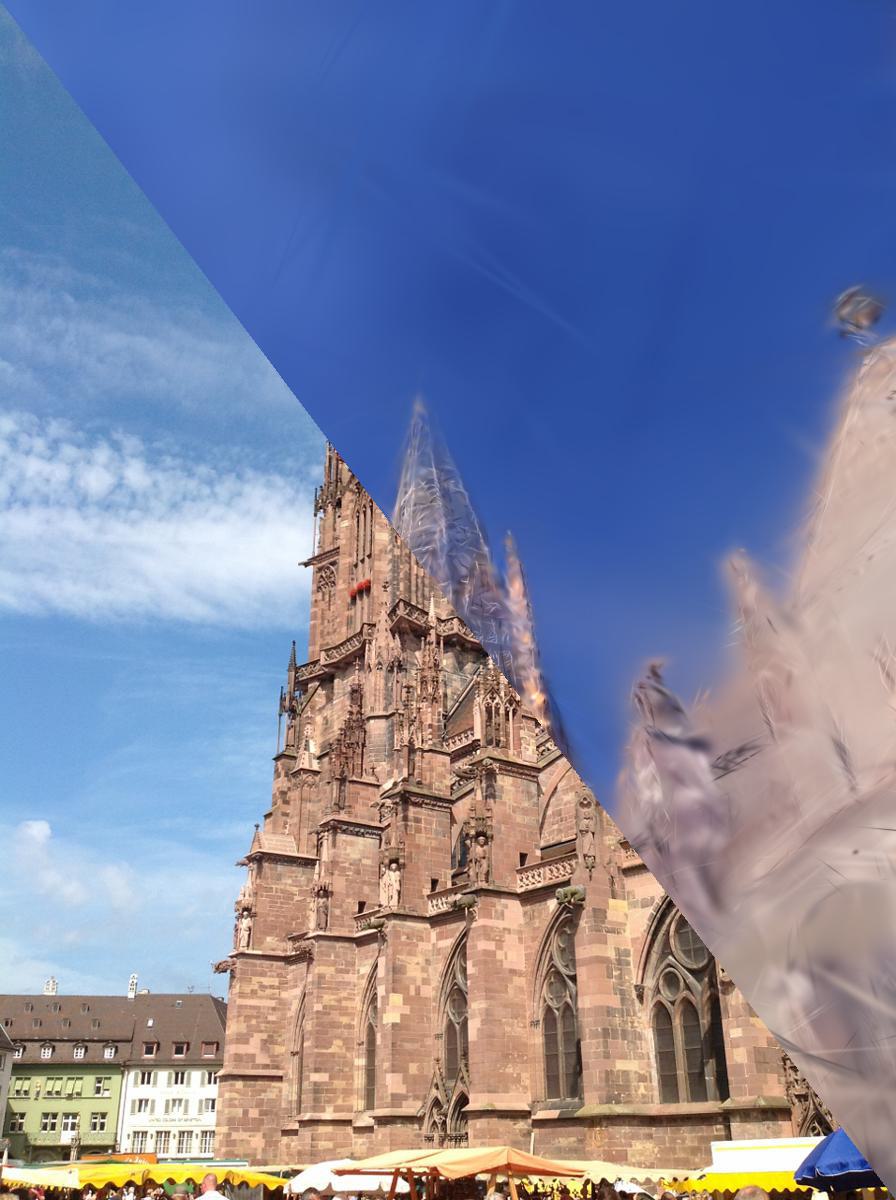} &
    \tile{}{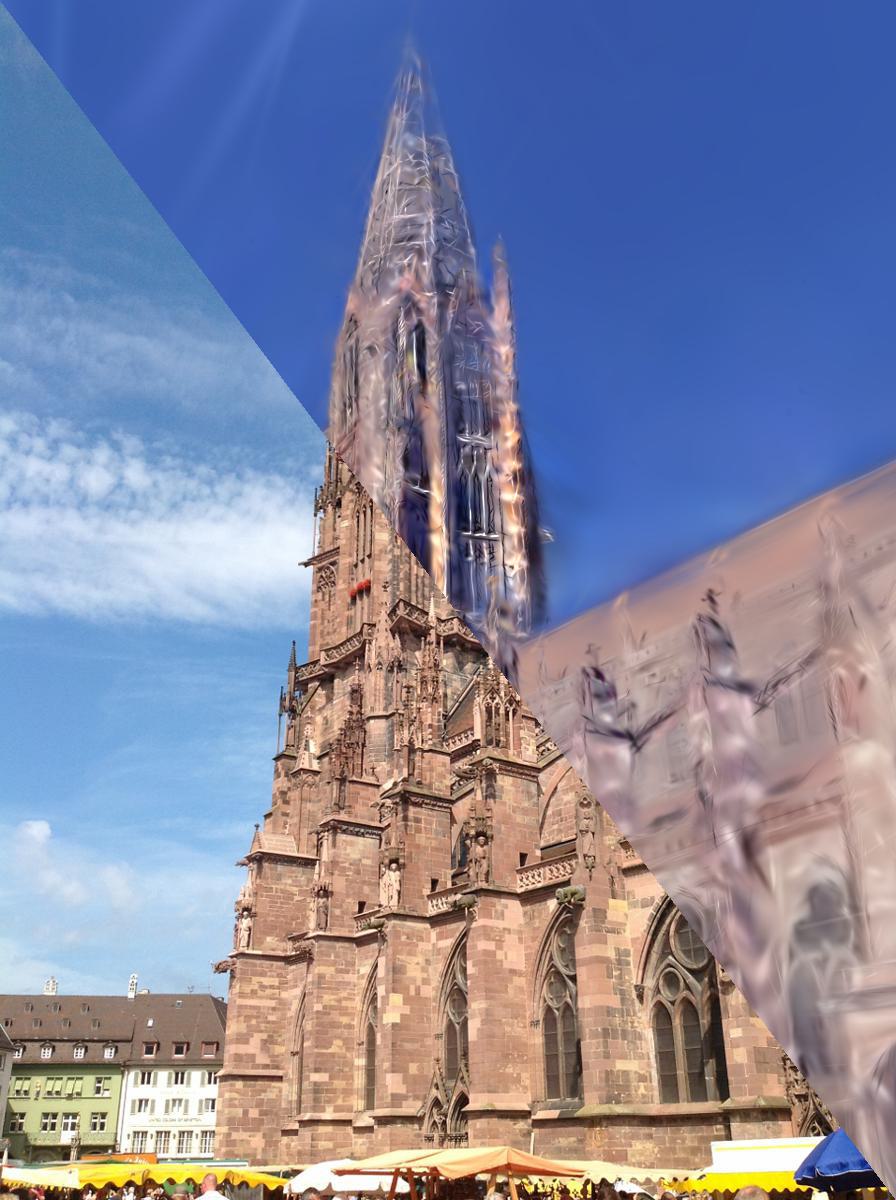} &
    \tile{}{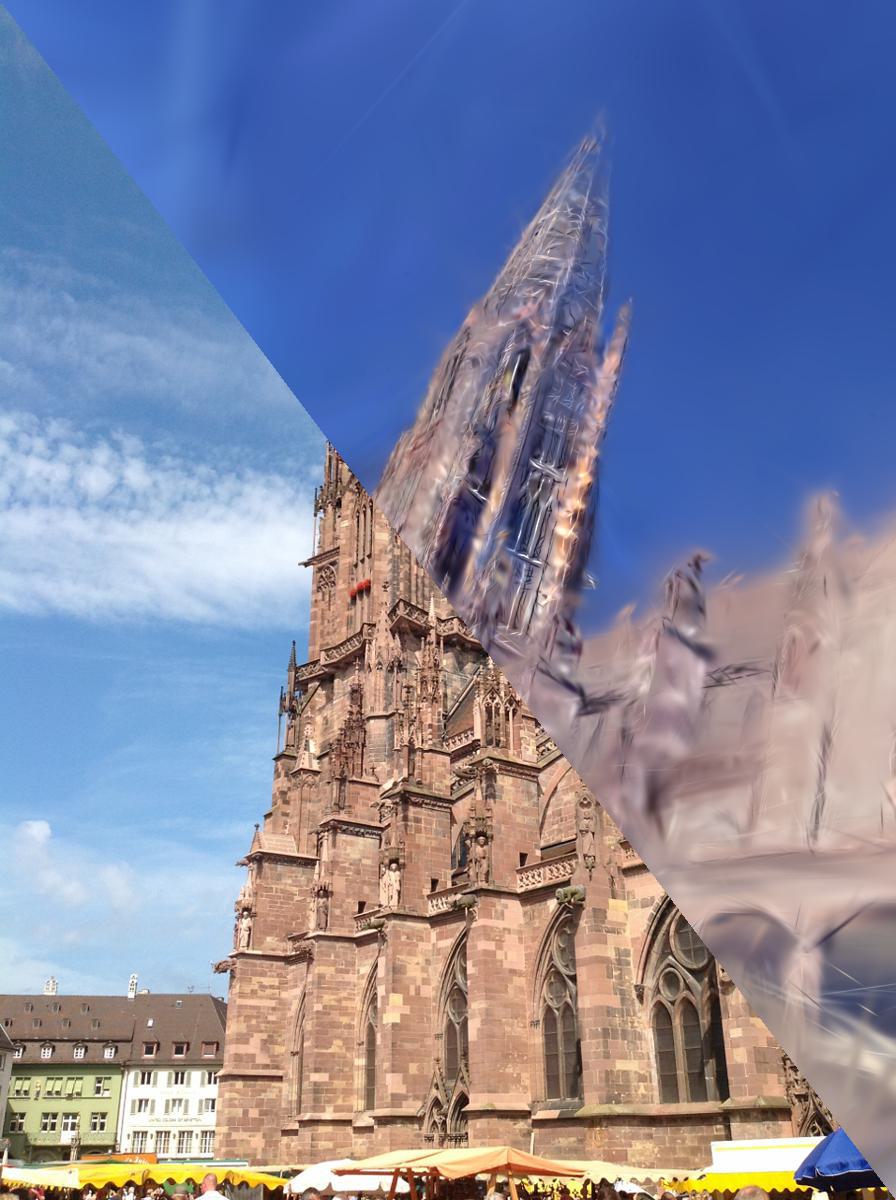} &
    \tile{}{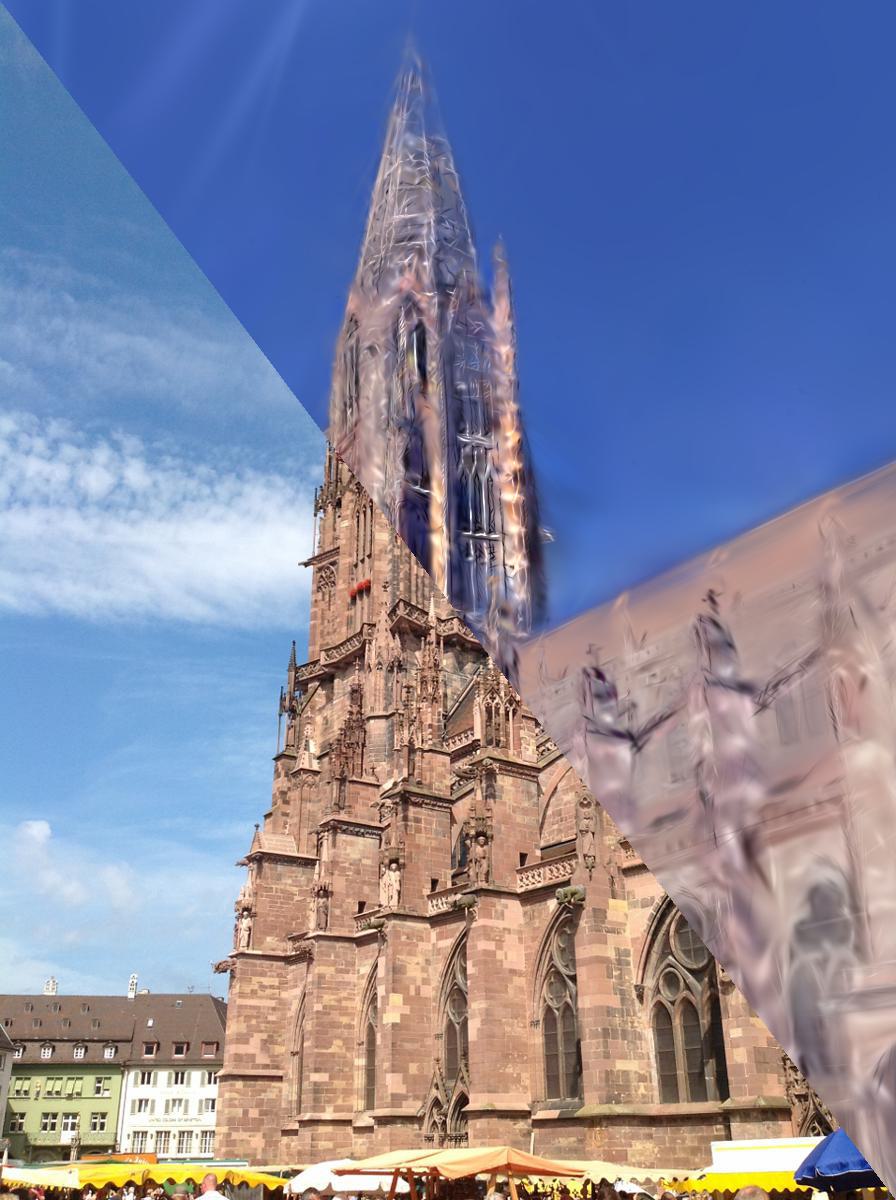} &
    \tile{}{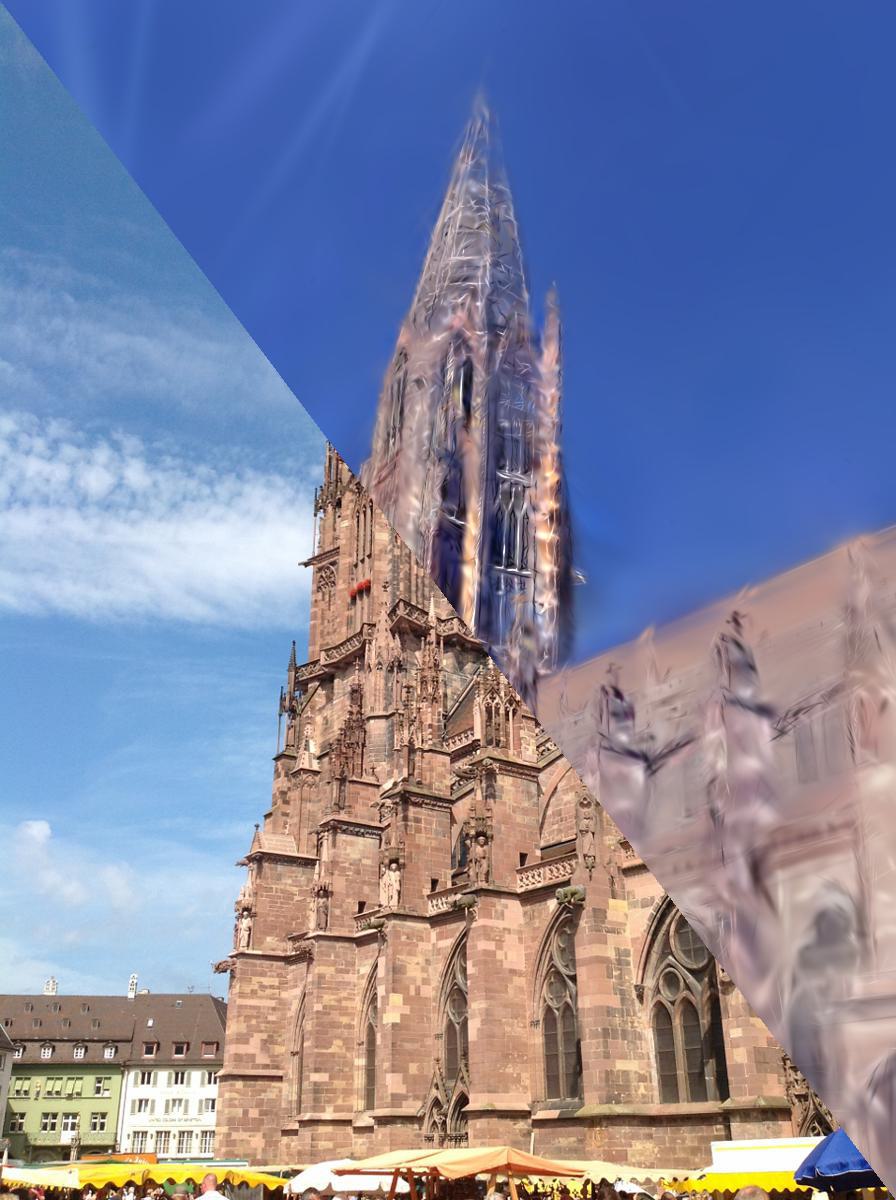} &
    \tile{}{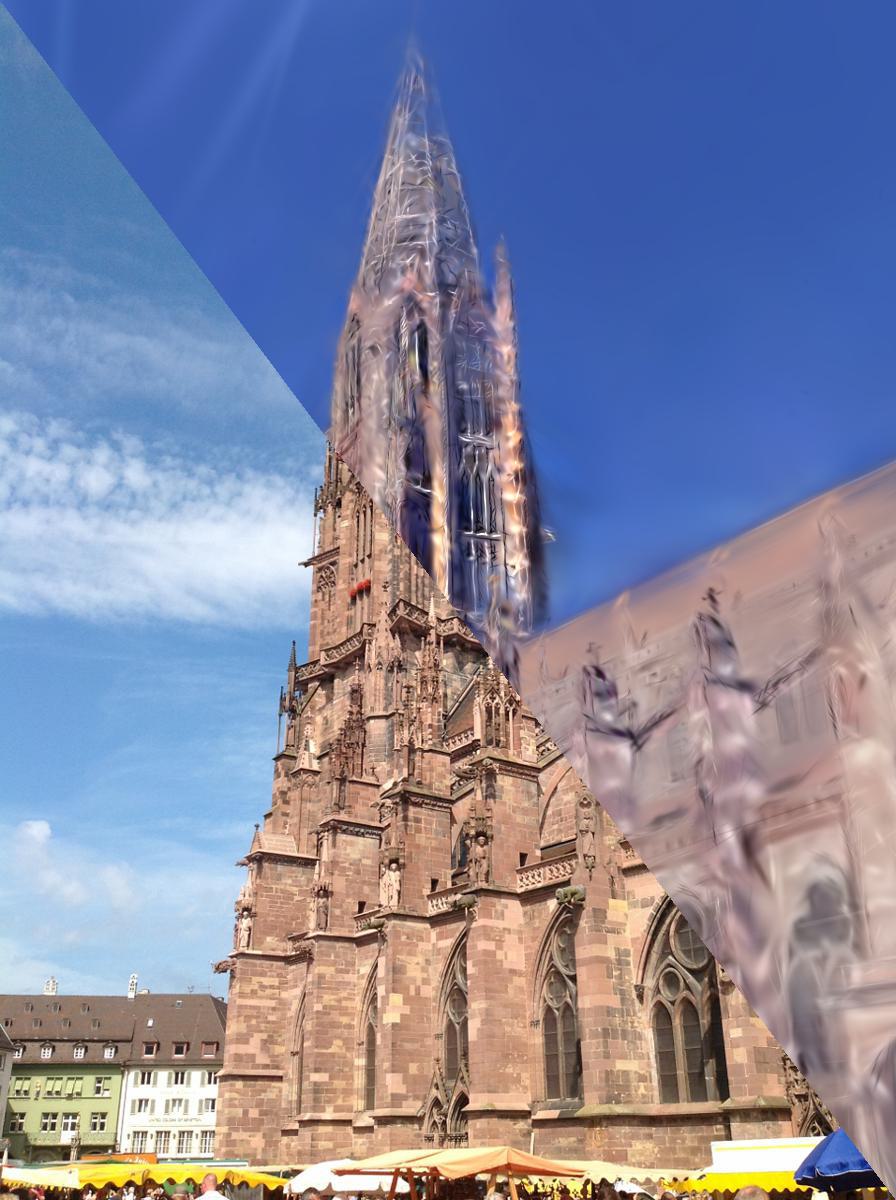}
    \\[6pt]

    \tile{trim=200 0 200 0,clip}{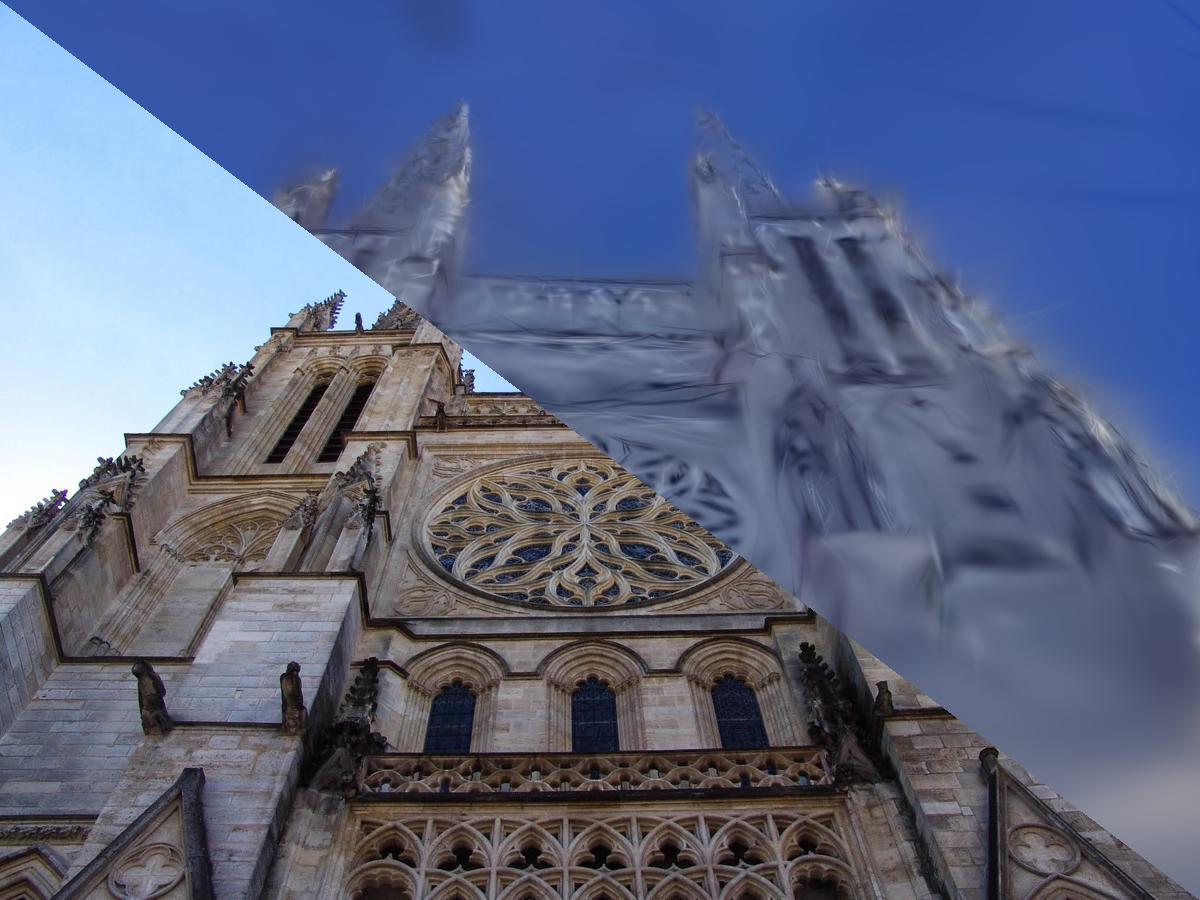} &
    \tile{trim=200 0 200 0,clip}{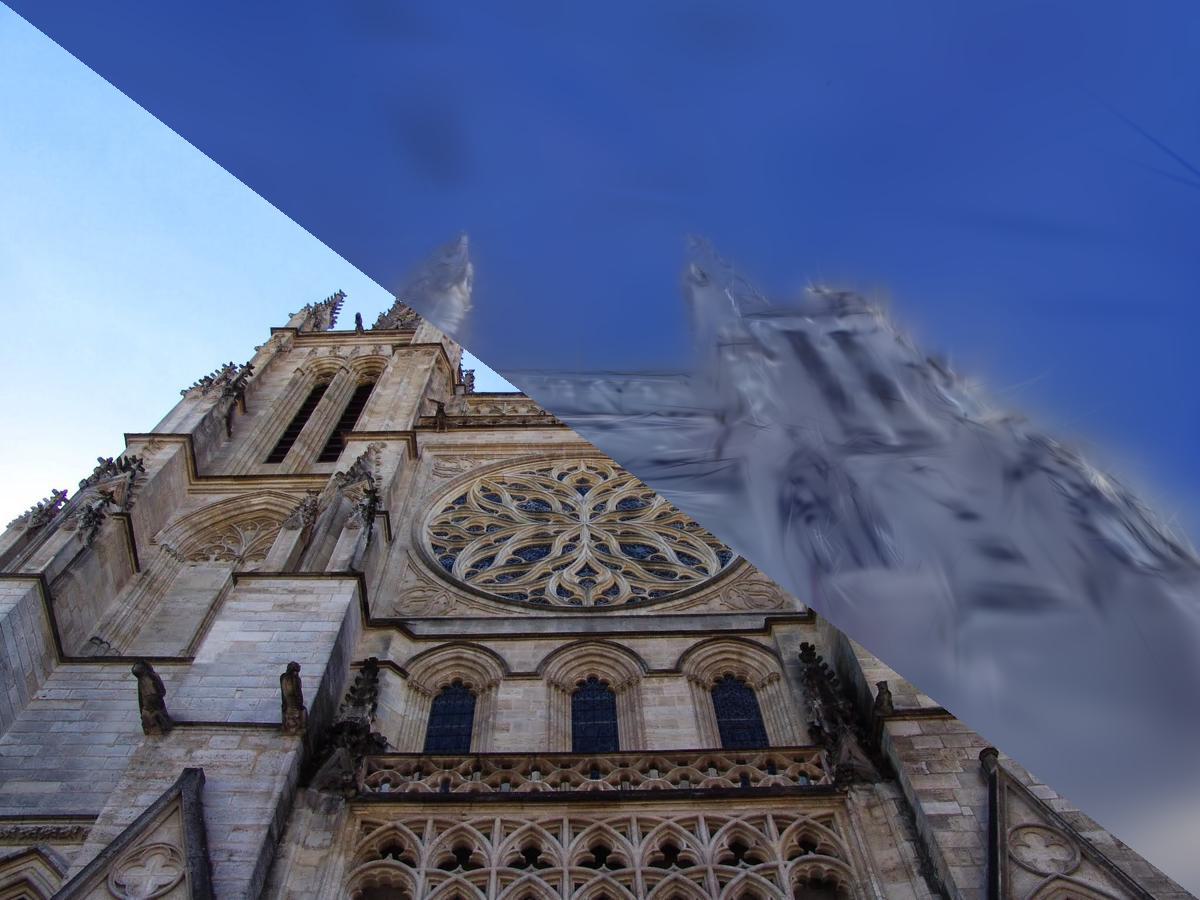} &
    \tile{trim=200 0 200 0,clip}{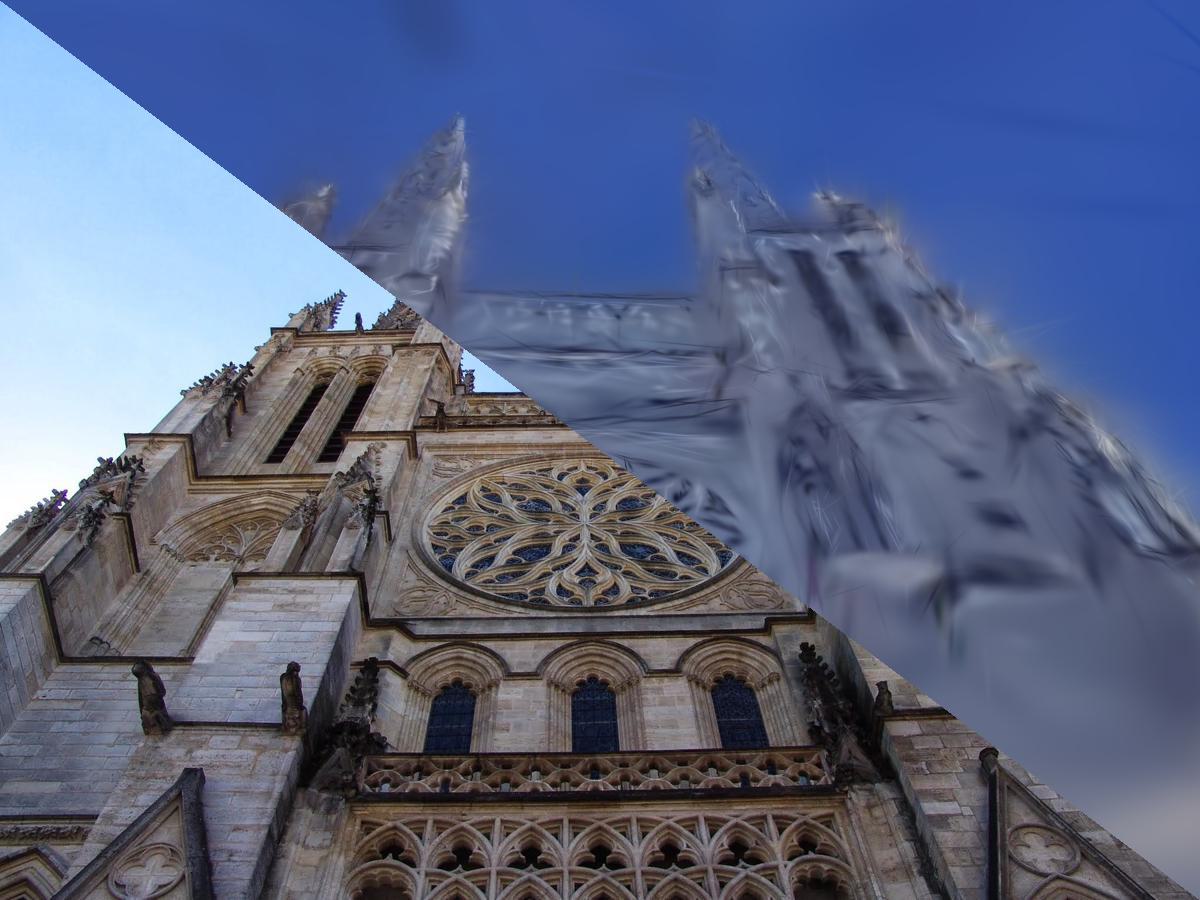} &
    \tile{trim=200 0 200 0,clip}{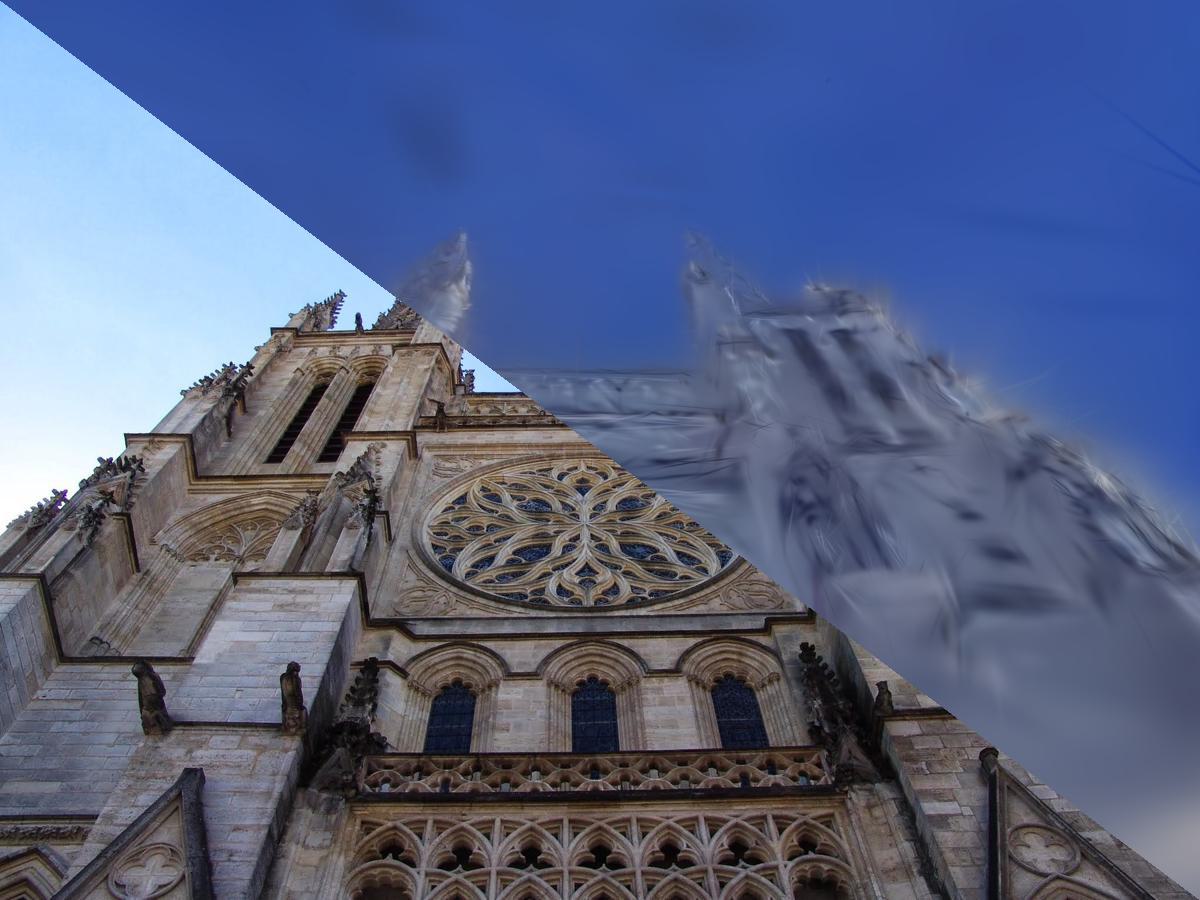} &
    \tile{trim=200 0 200 0,clip}{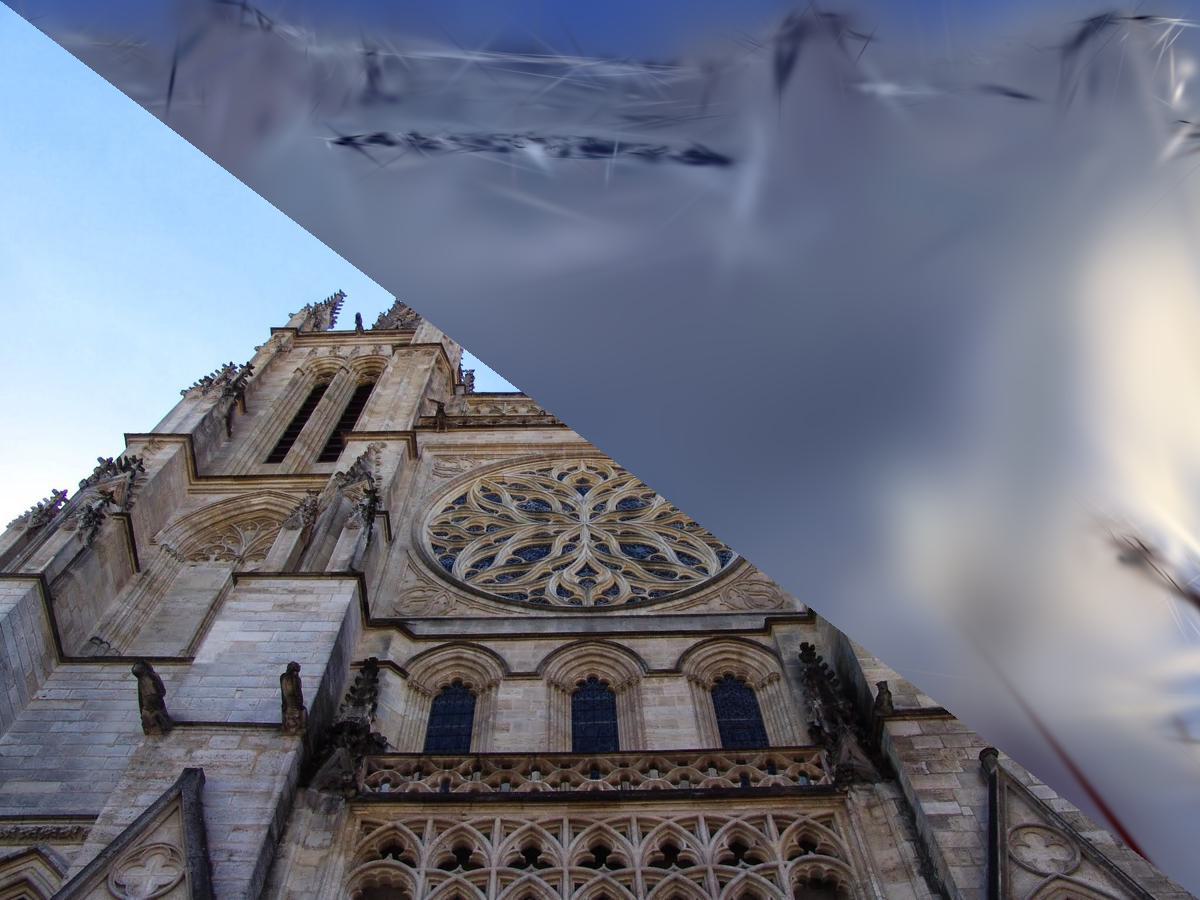} &
    \tile{trim=200 0 200 0,clip}{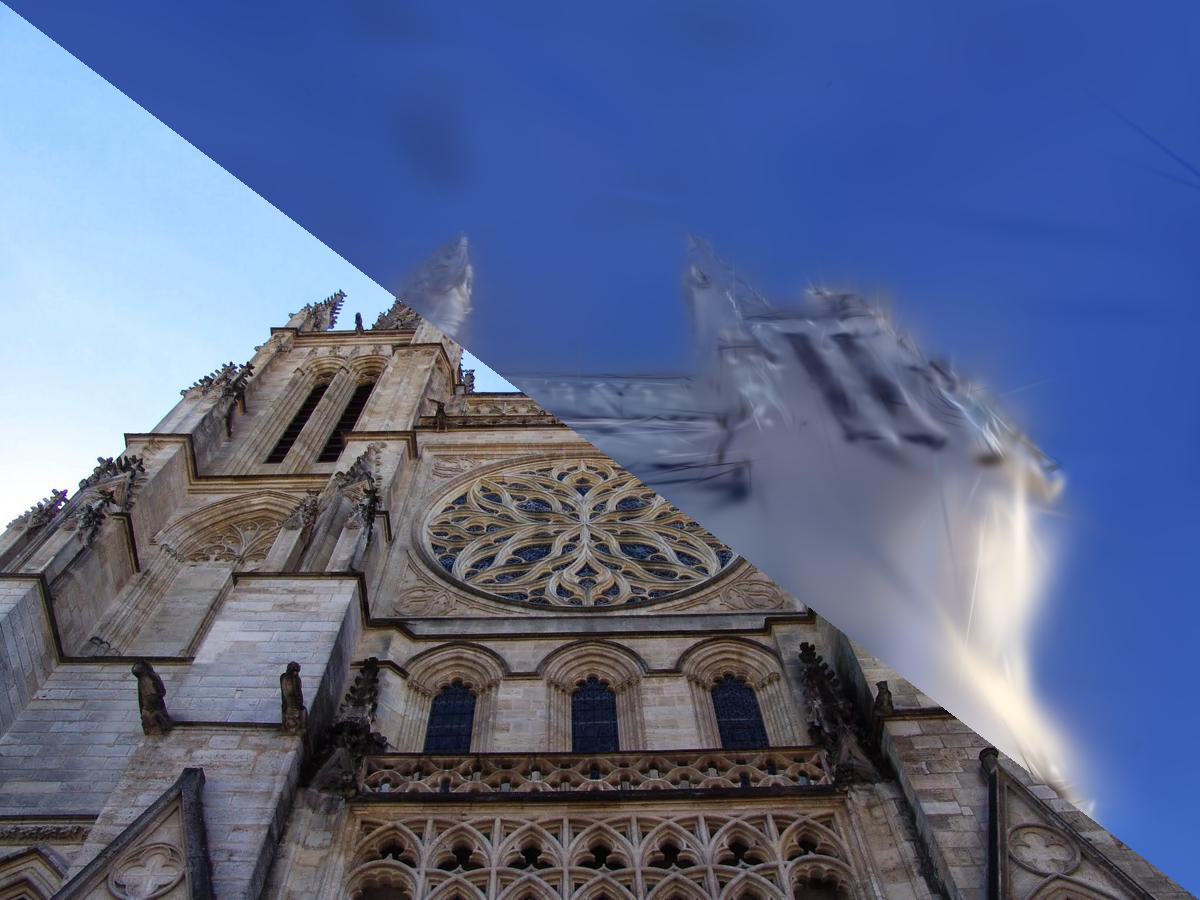}
    \\[6pt]

    \tile{}{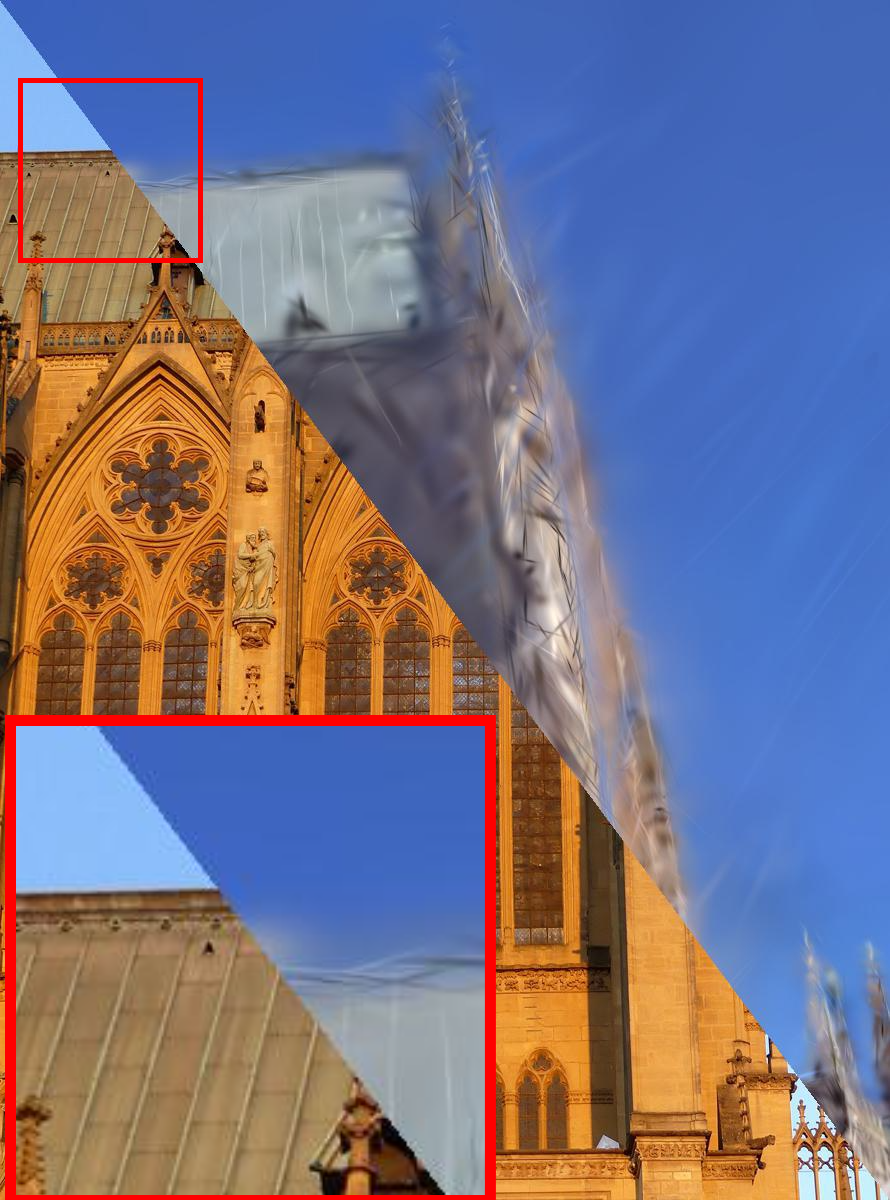} &
    \tile{}{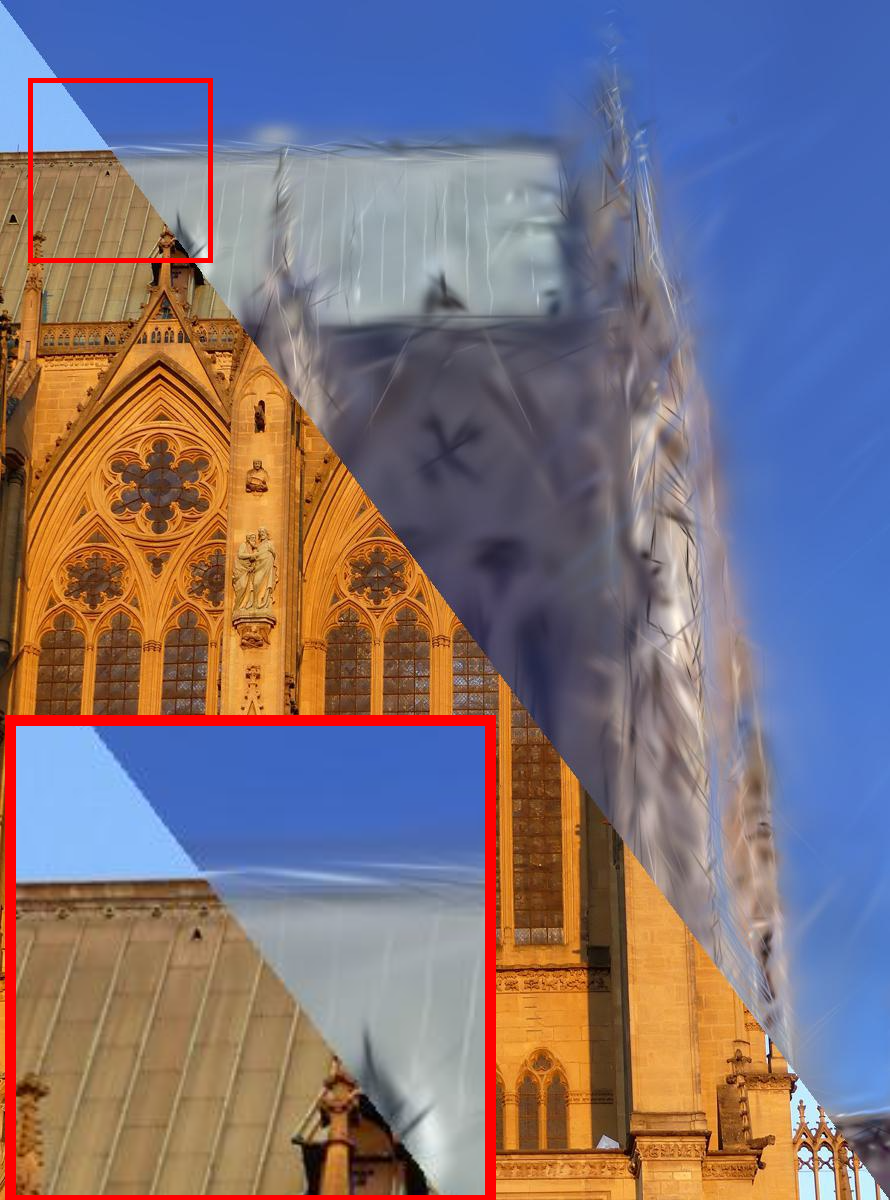} &
    \tile{}{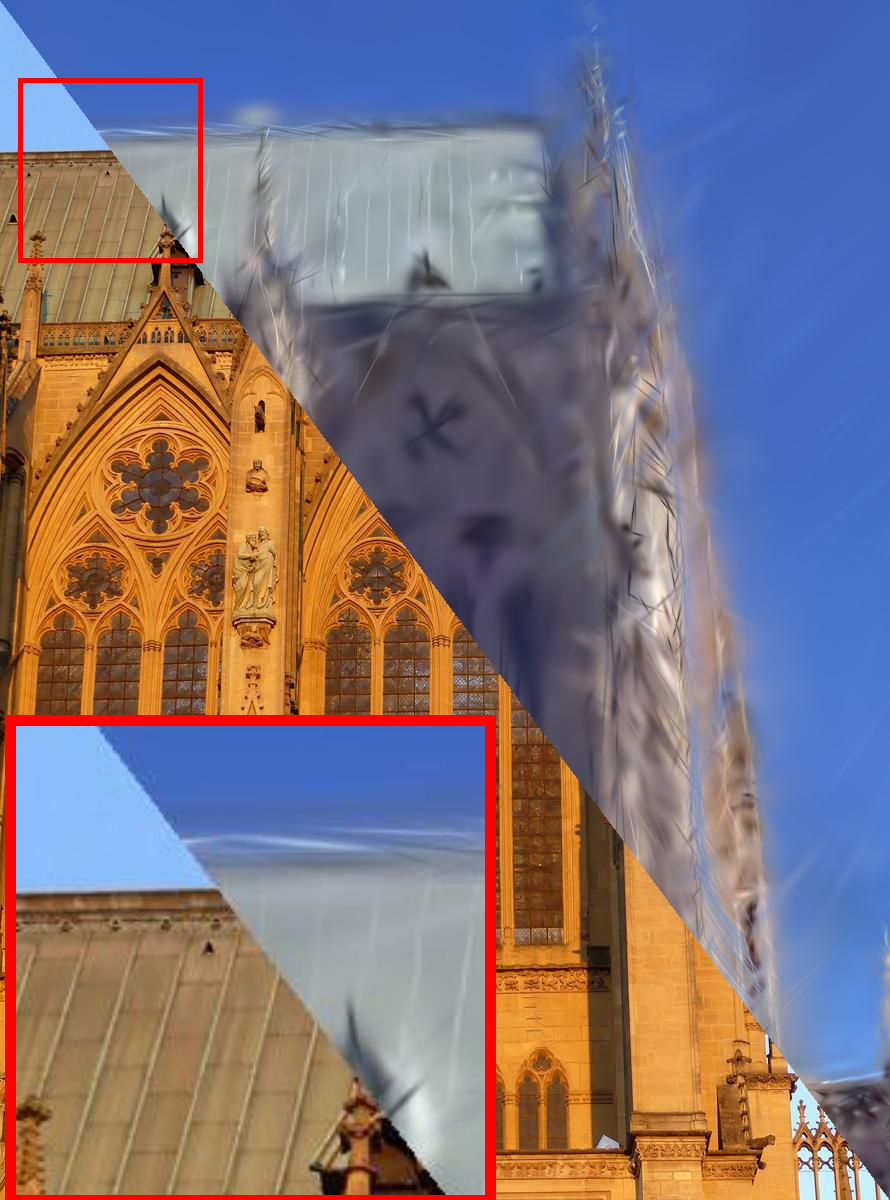} &
    \tile{}{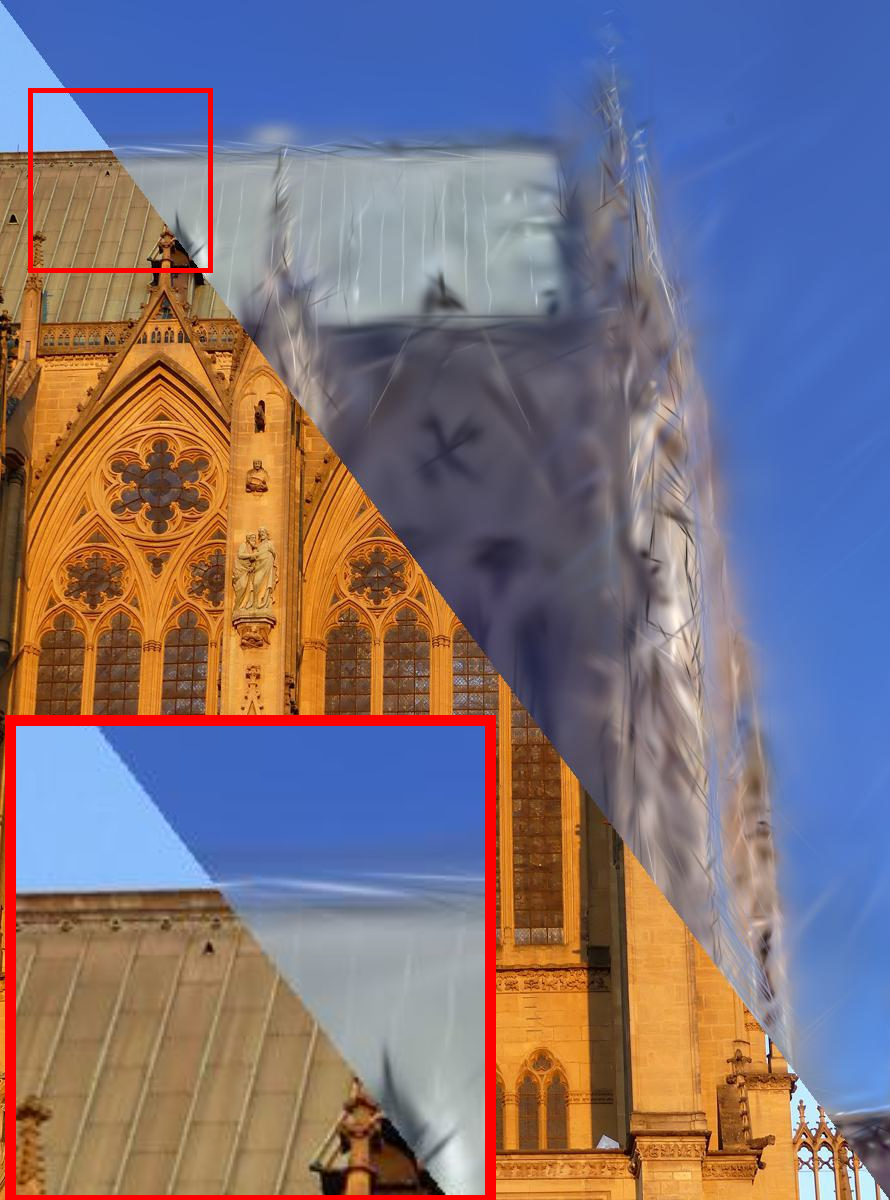} &
    \tile{}{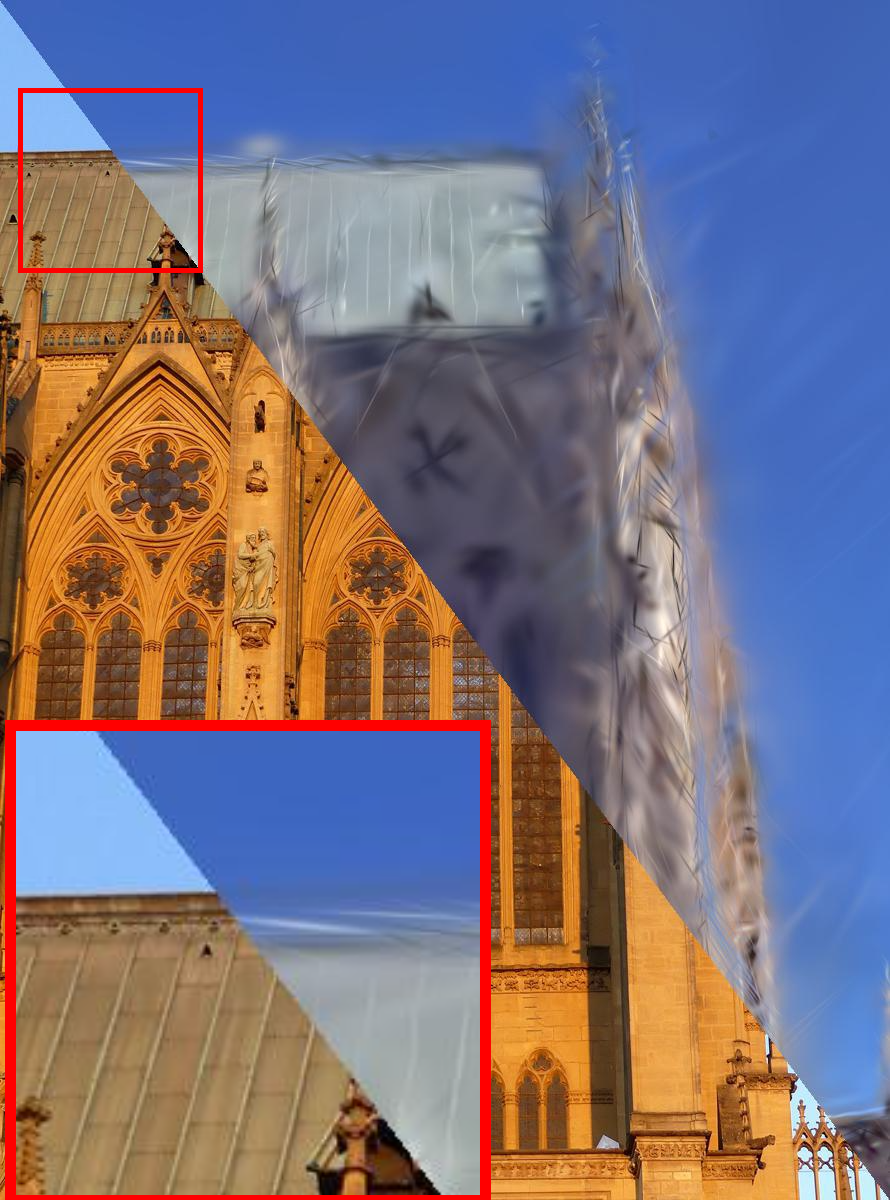} &
    \tile{}{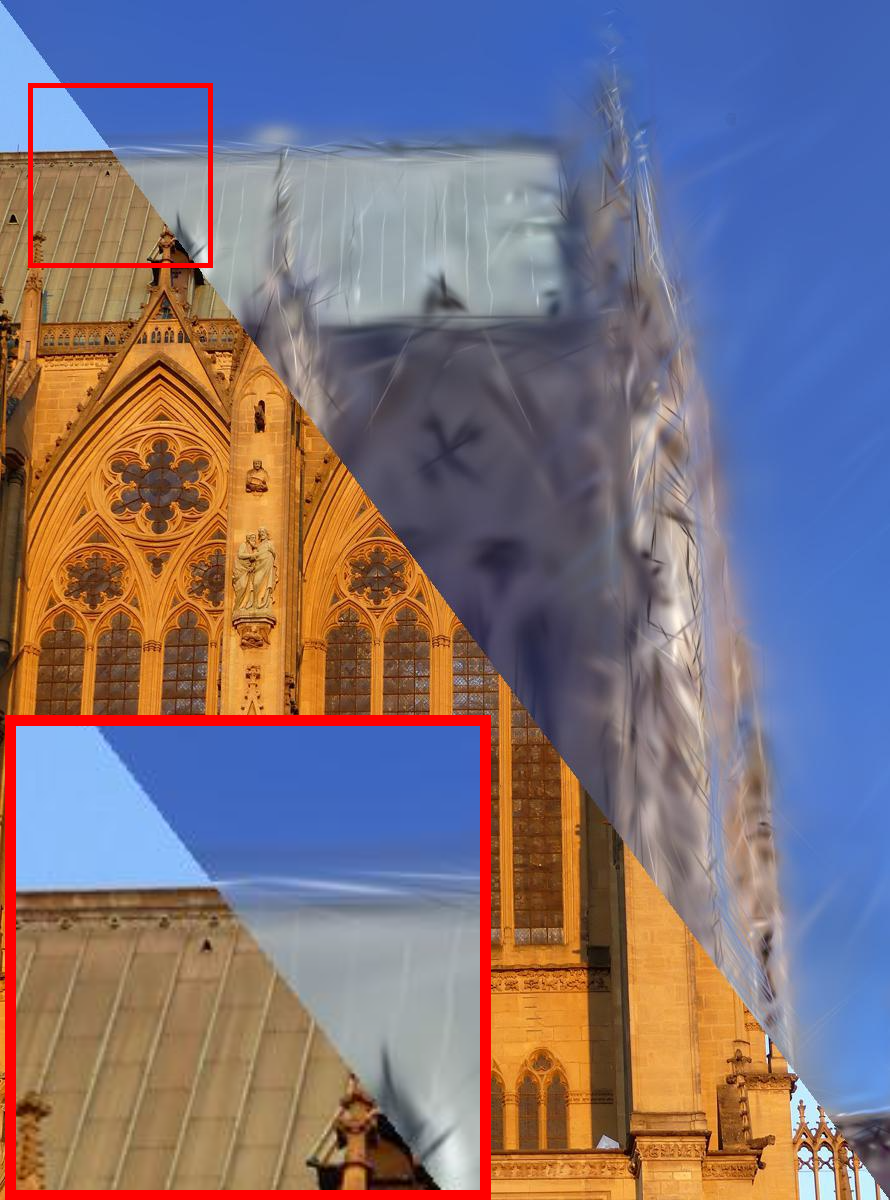}
  \end{tabular}
  }

  \caption{\textbf{Qualitative Comparison.} A visualization of the alignment results for our method compared to the three baselines. Each image shows the ground truth in the lower half and the rendered image from the reference model $\mathcal{M}$ after alignment in the top half. As demonstrated, our inverse optimization-based approach predicts precise transformations, even in the presence of challenging, inaccurate initializations. %
  }
  \label{baseline_comparison}
\end{figure}

\subsection{Evaluation Metrics}
\label{sec:metrics}
We evaluate the predicted meta-image transformations on the \dataset{} benchmark using $\Delta T_{\cal I}$ and $\Delta R_{\cal I}$, which denote the average RMS translation and rotation errors of the registered images belonging to each meta-image ${\cal I}$.
$\Delta R_{\cal I}$ is measured in degrees and $\Delta T_{\cal I}$ is expressed in the scale of the scene. All scenes are similarly scaled by COLMAP, using the rendered cameras as the reference scale. 

We report $\Delta T$ and $\Delta R$, which denote the average errors over all the meta-images (performance breakdown per meta-image is reported in the supplementary). Averages are computed only over successful meta-images, for which the method outputs a transformation. 

In addition to the average errors, we report the meta-image transformation accuracy (MTA) and the percentage of outliers $O\%$, similarly to iNeRF~\cite{yen2020inerf}. MTA and $O\%$ are reported over predefined ratios. That is, a meta-image transformation is considered accurate if $\Delta R <5^{\circ}$ and $\Delta T< 0.2$, and categorized as an outlier if $\Delta R >10^{\circ}$ or $\Delta T > 0.5$. In the supplementary, we report MTA and $O\%$ over additional threshold values.

\subsection{Comparison to Baselines}

\begin{table}
\setlength{\tabcolsep}{3pt}
 \def\arraystretch{0.95}
\centering
\caption{\textbf{Comparison with Baselines}. We compare performance over the \dataset{} benchmark against multiple baselines. We report performance of each baseline (top rows), and our method initialized with the baselines (bottom rows). \#Failures denotes the number of meta-images for which the initialization method failed to produce an alignment; errors are computed only over successful instances. As shown, our approach can be paired with a range of initializations, yielding significant improvements in most cases.
}

\resizebox{\linewidth}{!}{%
\begin{tabular}{llllll}
\toprule
Methods & ${\Delta R}$ $\downarrow$ & ${\Delta T}$ $\downarrow$ & 
MTA$\uparrow$ & $O\%\downarrow$ & \#Failures \\
\midrule
gDLS+++ & 2.86 & \textbf{0.12} & 78 & 6 & 1/32\\
Ours (gDLS+++ init)  & \textbf{2.69} & 0.13 & \textbf{84} & \textbf{3} & -\\
\midrule
SP+LG & 3.74 & 0.25 & 74 & 15 & 5/32\\
Ours (SP+LG init) & \textbf{3.13} & \textbf{0.24} & \textbf{81} & \textbf{7} & -\\
\midrule
COLMAP & 4.99 & 0.12 & 66 & 12 & 0/32\\
Ours (COLMAP init) & \textbf{2.48} & 0.12 & \textbf{81} & \textbf{0} & -\\
\bottomrule

\end{tabular}
}
\label{table:registration_results_table_baselines}
\end{table}

\label{sec:eval}

\begin{figure}
 \rotatebox{90}{\small{COLMAP}}
 \jsubfig{\includegraphics[width=0.30\linewidth,trim=200 0 200 200, clip]{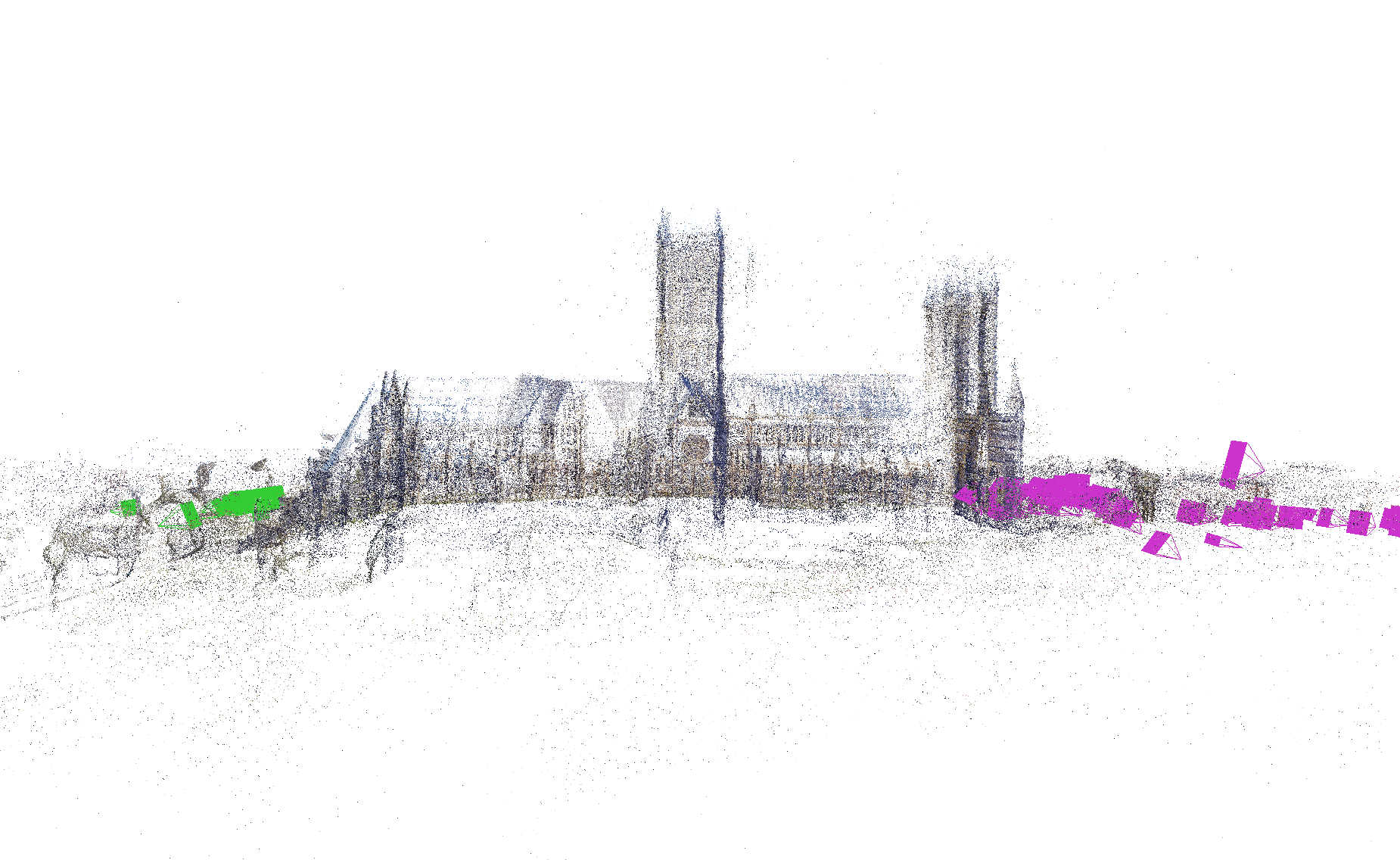}}{}
 \jsubfig{\includegraphics[width=0.30\linewidth,trim=200 0 100 200, clip]{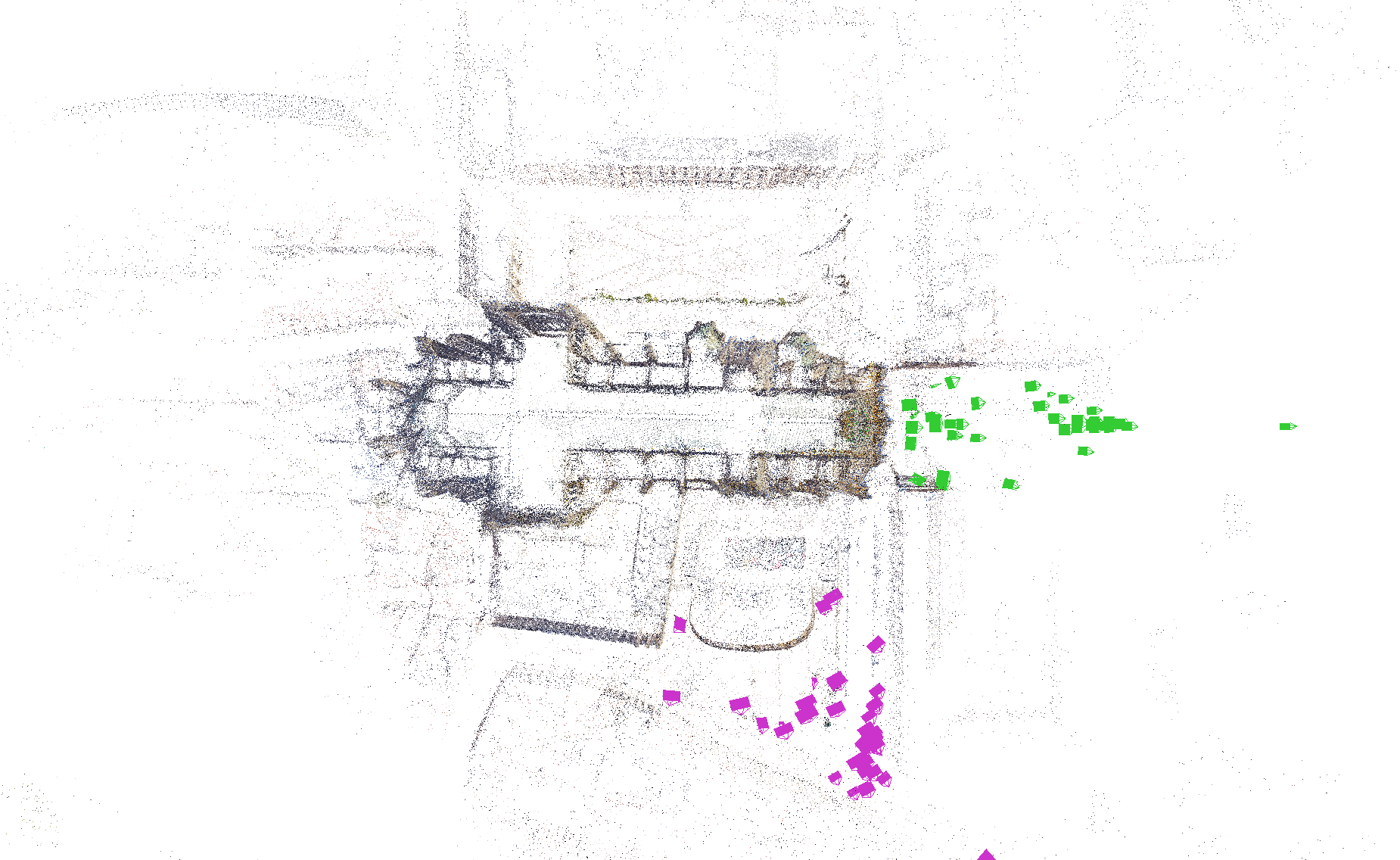}}{}
    \jsubfig{\includegraphics[width=0.30\linewidth,trim=500 200 300 100, clip]{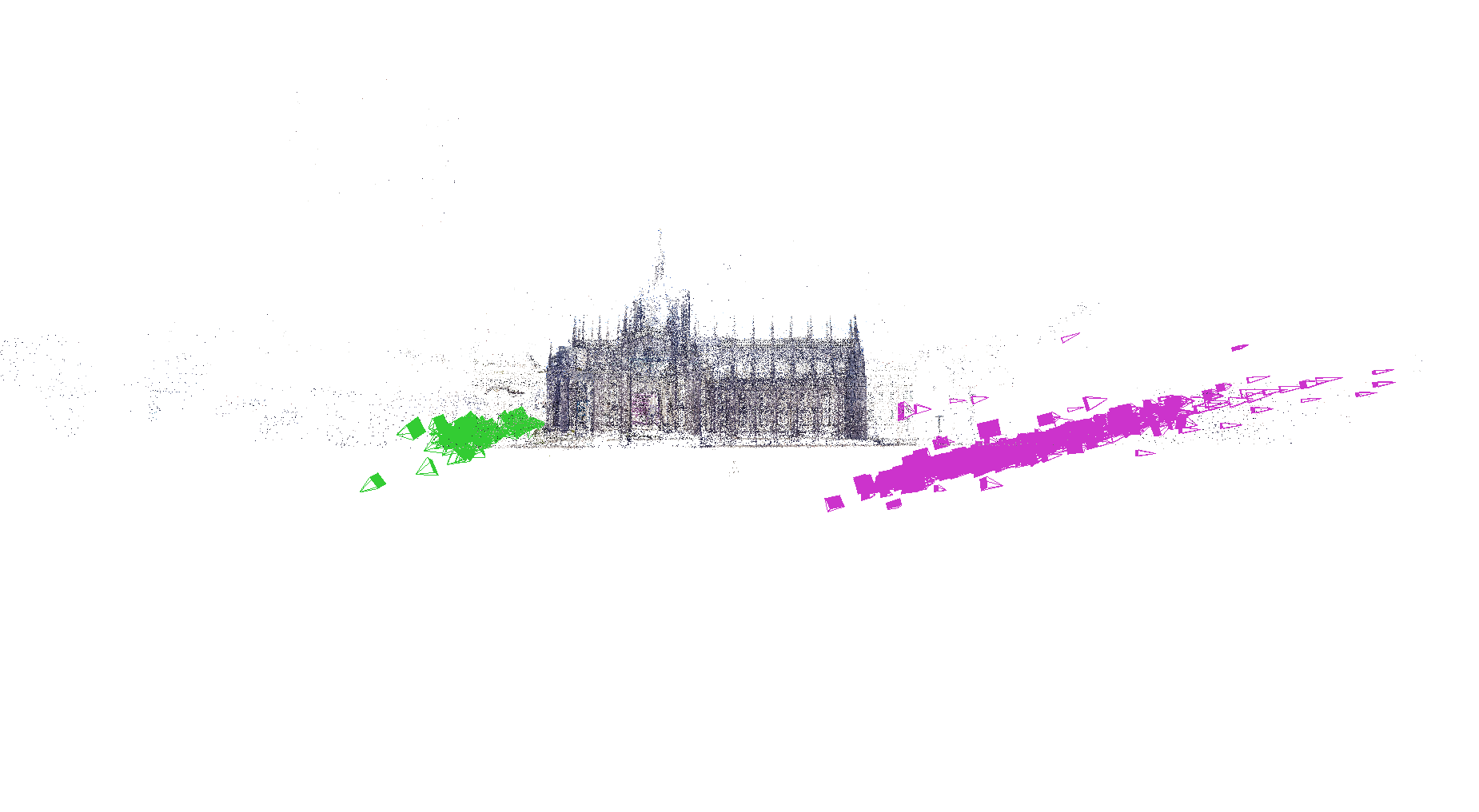}}{}
    \\ 
    \rotatebox{90}{\whitetxt{bb} \small{Ours}}
    \jsubfig{\includegraphics[width=0.30\linewidth,trim=200 0 200 200, clip]{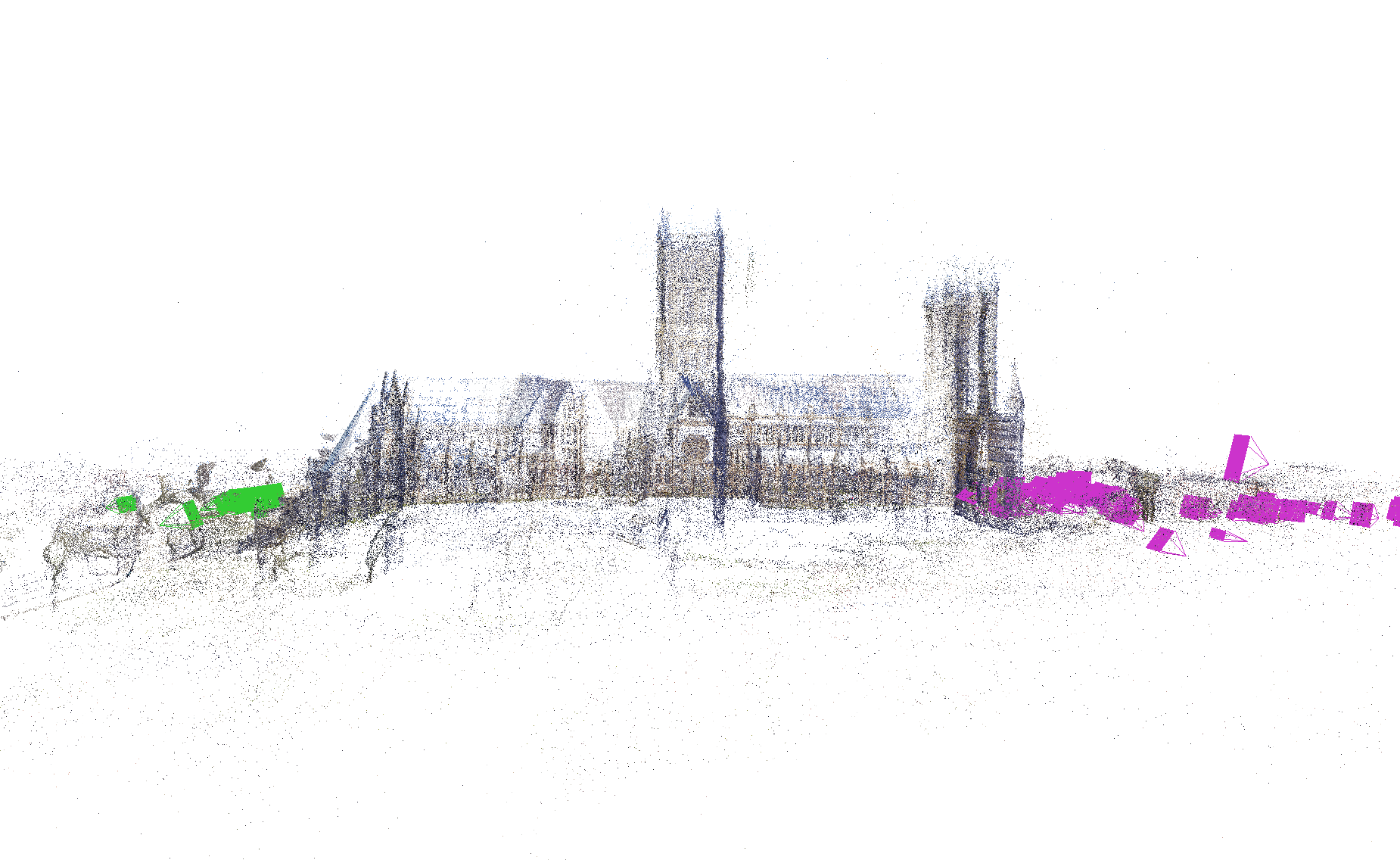}}{}
    \jsubfig{\includegraphics[width=0.30\linewidth,trim=200 0 100 200, clip]{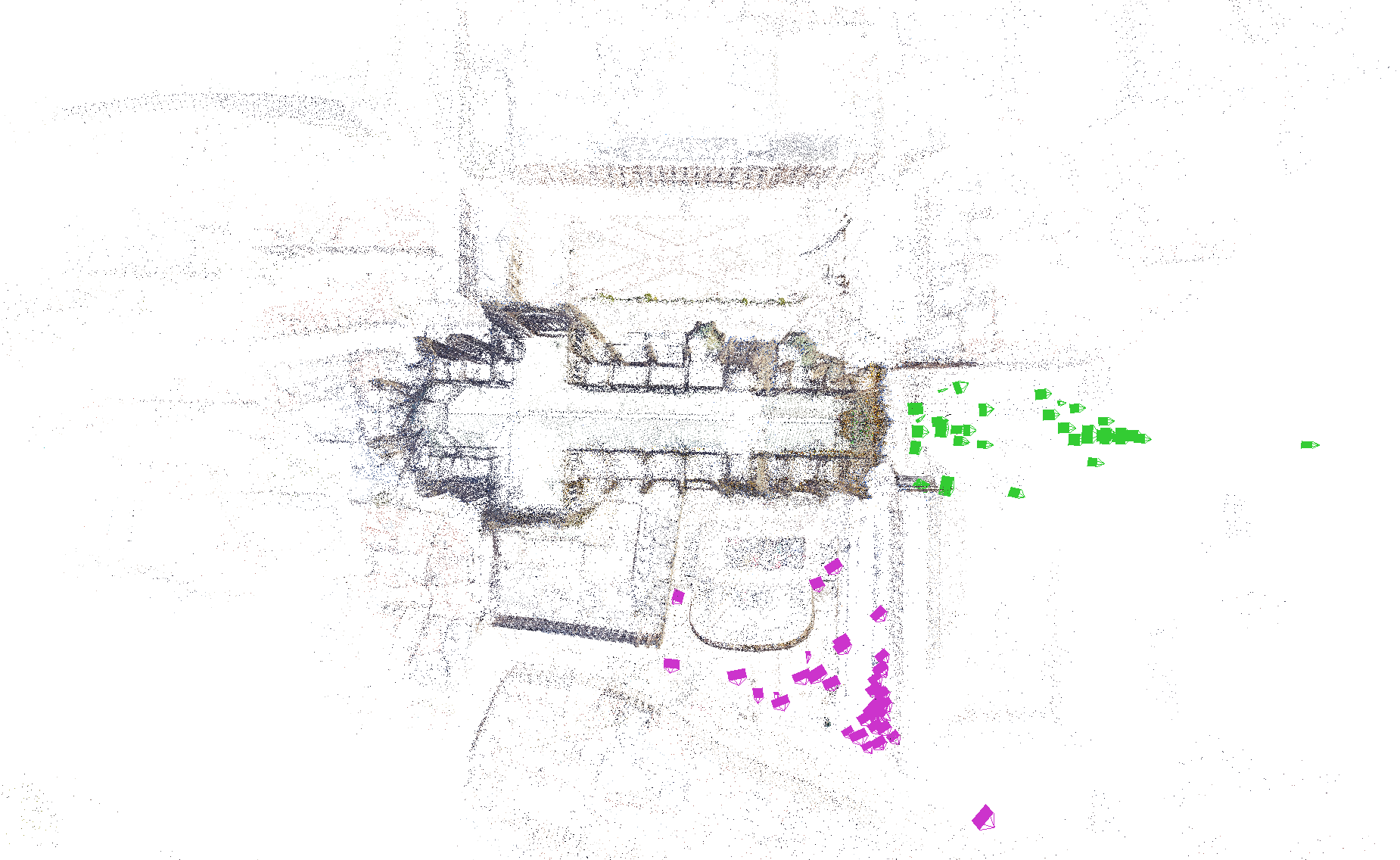}}{}
    \jsubfig{\includegraphics[width=0.30\linewidth,trim=500 200 300 100, clip]{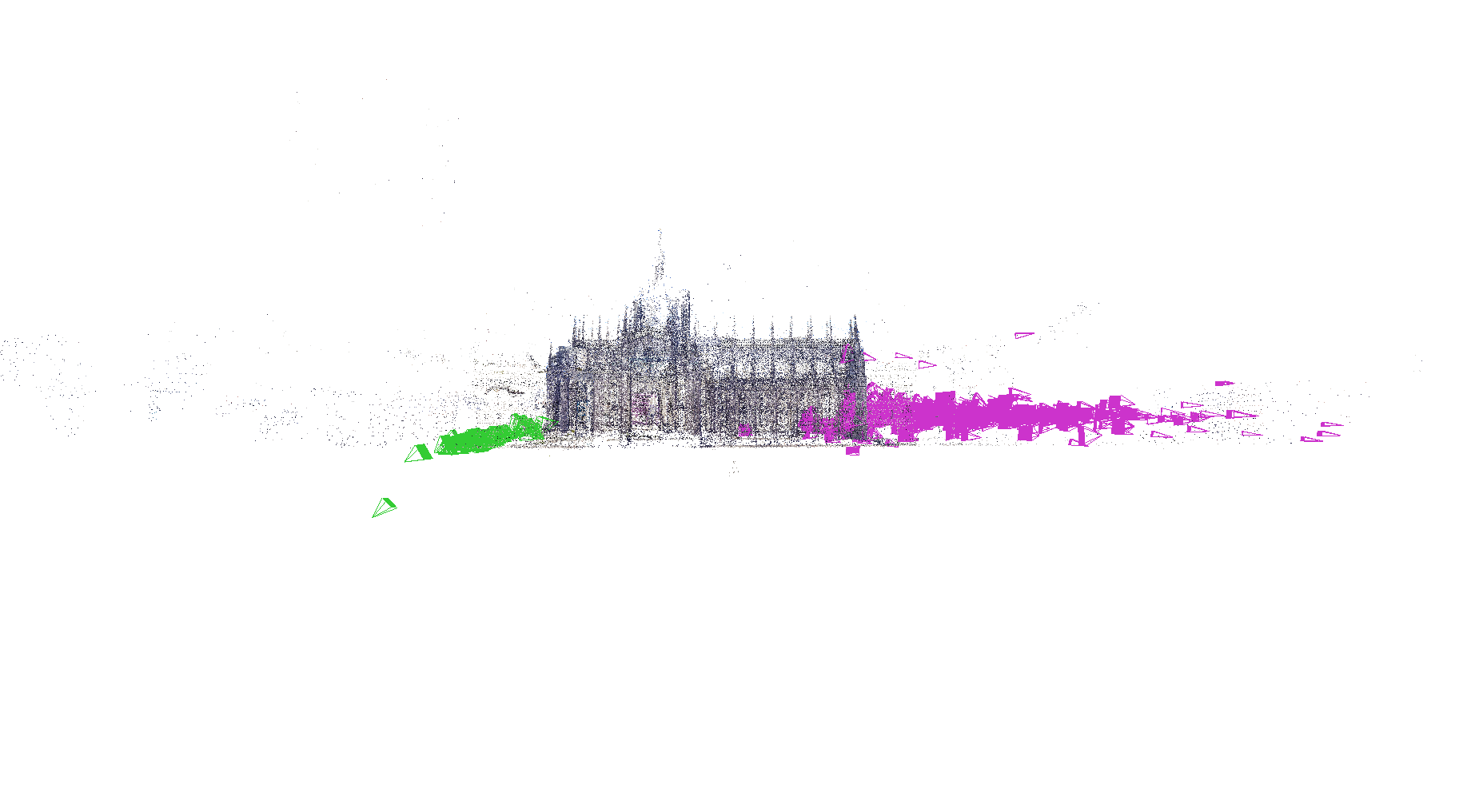}}{}
    \\ 
    \rotatebox{90}{\small{Ground Truth}}
    \jsubfig{\includegraphics[width=0.30\linewidth,trim=200 0 200 200, clip]{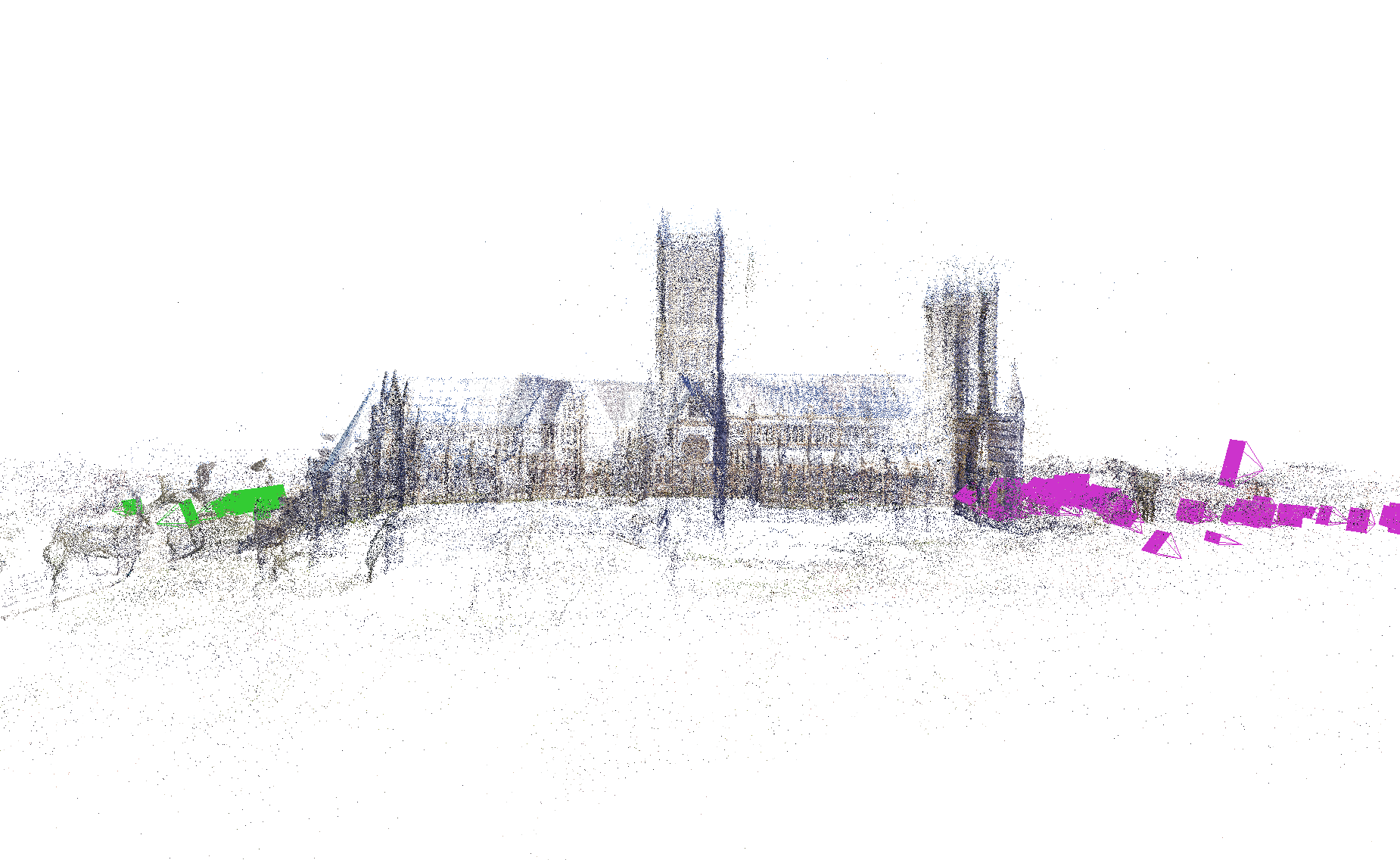}}{\small{Lincoln Cathedral}}
    \jsubfig{\includegraphics[width=0.30\linewidth,trim=200 0 100 200, clip]{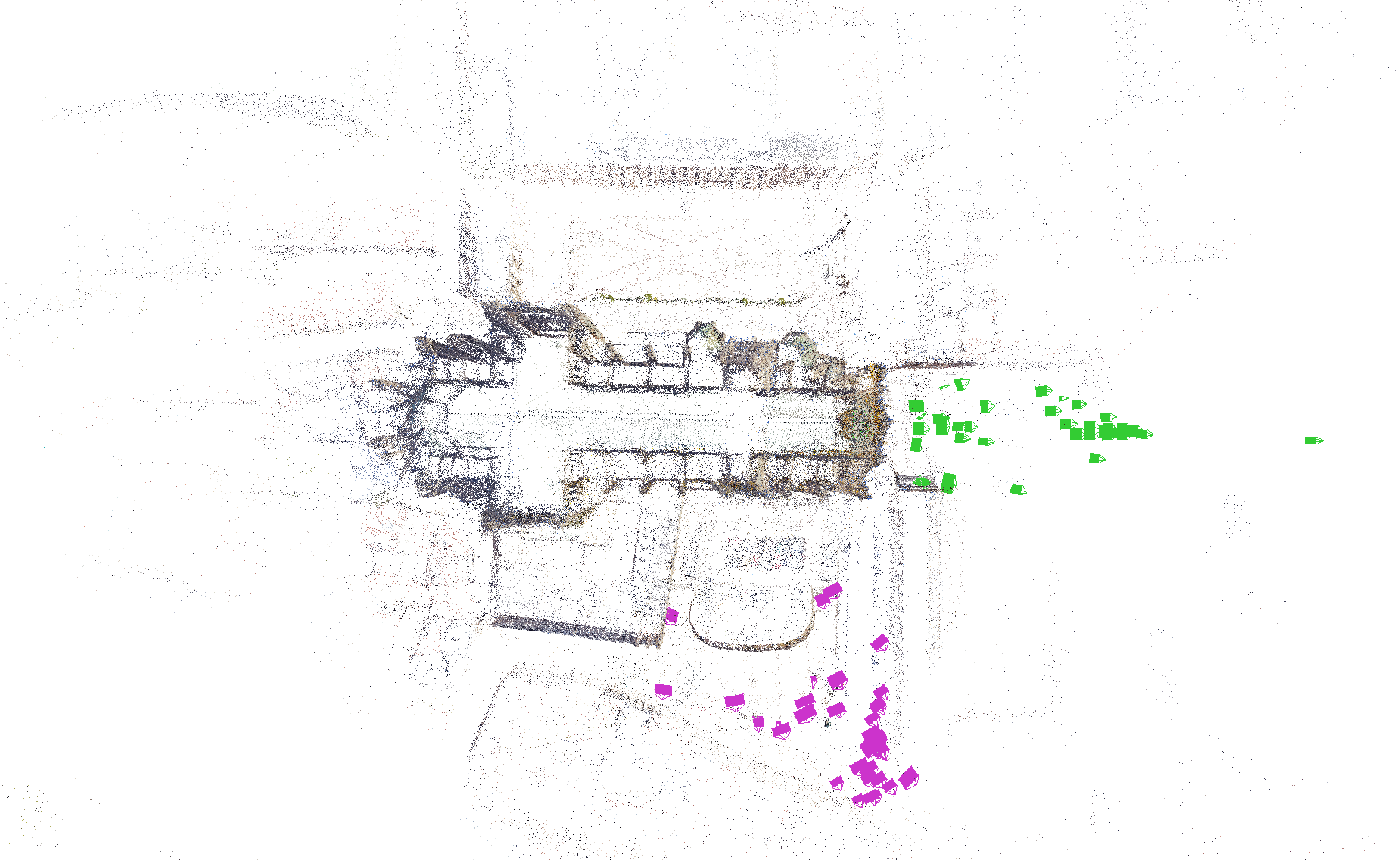}}{\small{Metz Cathedral}}
    \jsubfig{\includegraphics[width=0.30\linewidth,trim=500 200 300 100, clip]{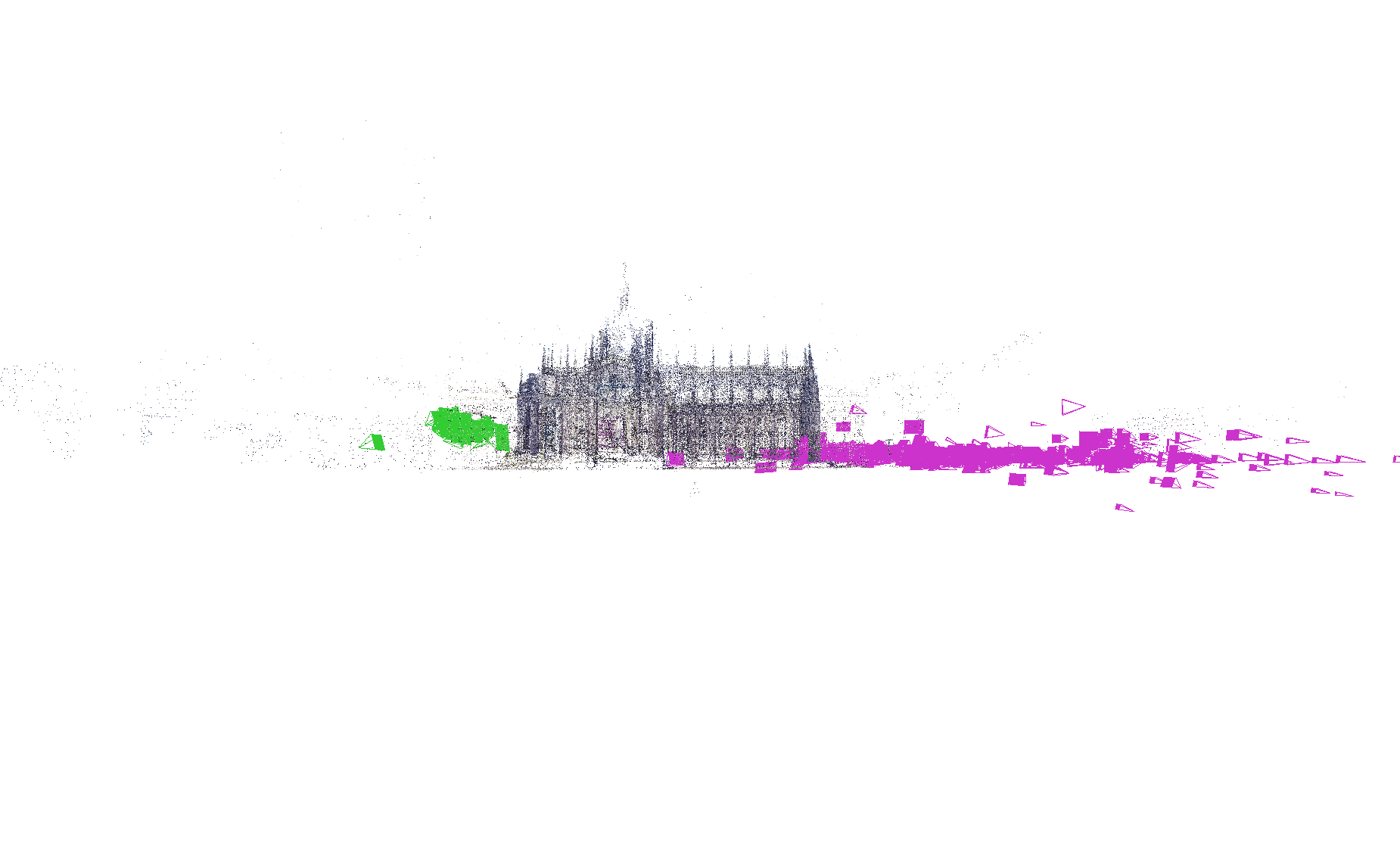}}{\small{Milan Cathedral}}

    \caption{\textbf{Grounding Multiple Meta Images}. 
    Above we show three scenes containing two meta-images per scene (visualized in \textcolor{green_teaser}{\bf green} and \textcolor{magenta_teaser}{\bf purple}), both grounded to the scene's global reference model. We show both the COLMAP initialization, and our final result. Ground truth reconstructions are provided on the bottom. %
    }
\label{fig:grounded_reconstructions}
\end{figure}

Quantitative results are reported in Table \ref{table:registration_results_table_baselines}. As illustrated, our approach outperforms all baselines in all metrics and consistently improves the initialization performance. %
Performance breakdown over all scenes and configurations are provided in the supplementary material.

We also provide a qualitative comparison over different initializations in Figure \ref{baseline_comparison}. %
In Figure \ref{fig:grounded_reconstructions}, we present alignment results of multiple meta-images across several landmarks, comparing our method and the COLMAP baseline to the ground truth. As can be observed from these visualizations, our method can successfully align meta-images with significantly erroneous initializations.

\subsection{Comparison to Feed-Forward 3D Models}
\label{sec:mastr}
\begin{figure}
    \rotatebox{90}{\small{\whitetxt{xxxxx}Input Images for $\pi^3$ Reconstruction: }}
    \rotatebox{90}{\small{Google Earth + Internet \hspace{7pt} only Google Earth} }
    \hfill
\jsubfig{{\includegraphics[width=0.90\linewidth] {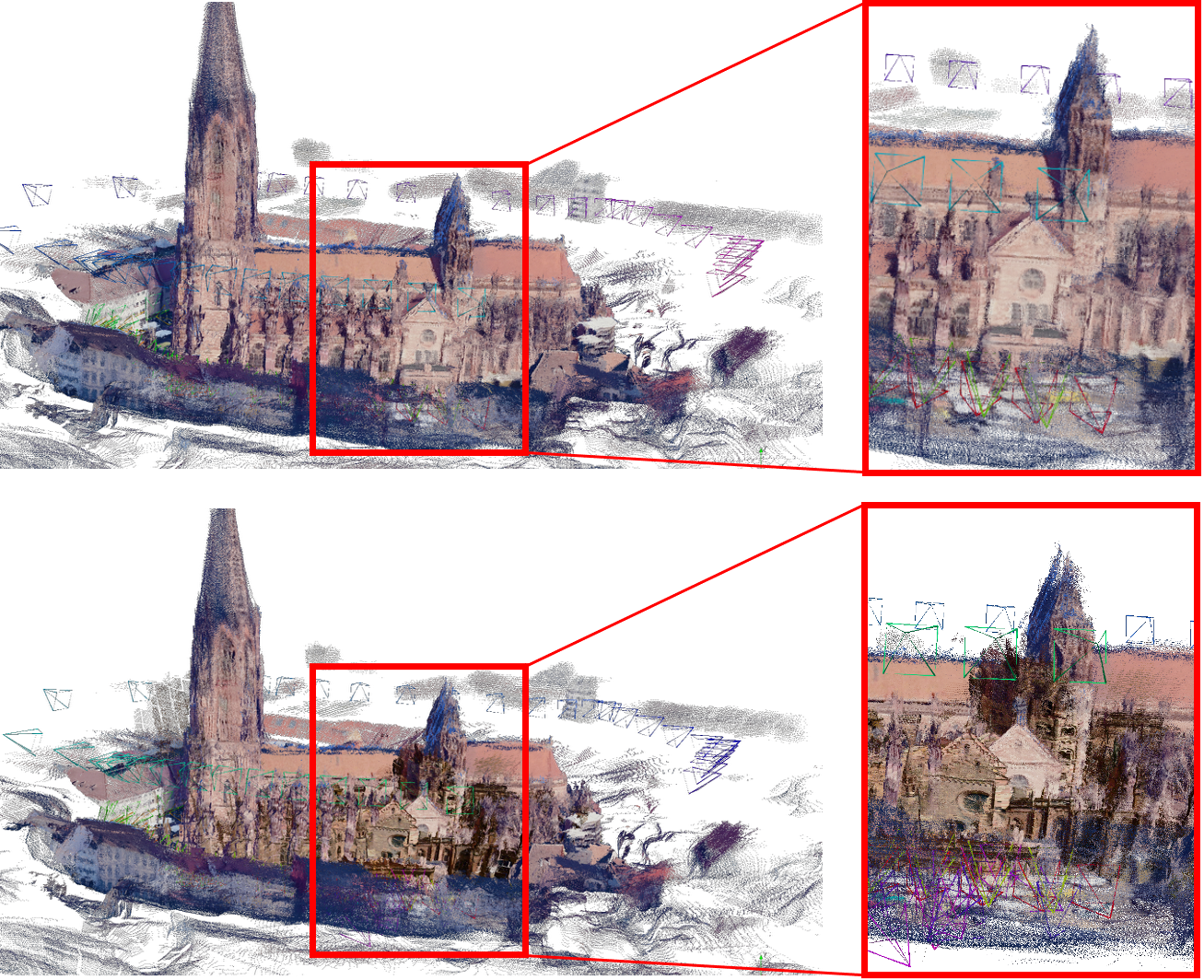}}}{}
    \vspace{-19pt}
    \caption{\textbf{Aligning a meta-image to a reference model with $\pi^3$.} As illustrated above, while $\pi^3$ successfully registers the Google Earth images, it struggles to correctly align the Internet images in this model; see, for instance, the \textbf{ghost structure} in the center of the red box on the bottom row. The top result, reconstructed from Google Earth images only, is shown for reference.}
    \label{fig:pi3}
\end{figure}

We compare our method against  %
DUSt3R~\cite{wang2024dust3r}, MASt3R~\cite{leroy2024mast3r}, VGGT~\cite{wang2025vggt} and $\pi^3$~\cite{wang2025pi3}, recent feed-forward reconstruction models. 
We conduct two sets of experiments to demonstrate that these models struggle in our problem setting: (i) meta-meta reconstructions, which reconstructs two meta-images belonging to the same physical scene without a reference model, and (ii) meta-to-reference reconstructions, which aligns a single meta-image to the reference model. We perform these experiments over the seven scenes in \dataset{} that contain multiple meta-images; this yields 16 and 32 meta-meta and meta-to-reference comparisons, respectively.  

As these methods cannot handle hundreds of images, we subsample 45 images each time (larger collections yield OOM on our A5000 GPU). For the alignment with the reference model, we subsample 35 from the reference model and 10 from each meta-image. As $\pi^3$ is more memory efficient, for this baseline we subsample 180 images; for the alignment with the reference model we subsample 150 and 30 images from the reference model and meta-image respectively. 
We sample evenly from each meta-image for the experiment without the reference model. Each run is repeated five times. To quantify performance, we measure geodesic rotation error~\cite{bezalel2025extremerotationestimationwild} between image pairs: ${\Delta R}_{\cal I\leftrightarrow\cal I}$ denotes the average error for meta-meta reconstructions and ${\Delta R}_{\cal I\leftrightarrow\cal M}$ denotes the average error for meta-to-reference alignment.

\begin{table}
\setlength{\tabcolsep}{6pt}
 \def\arraystretch{0.95}
\centering
\caption{\textbf{Comparison with Feed-Forward 3D Models.} We report geodesic rotation errors over two settings, comparing two meta-images directly (${\Delta R}_{\cal I\leftrightarrow\cal I}$) and comparing a single meta-image with a reference model (${\Delta R}_{\cal I\leftrightarrow\cal M}$). As illustrated below, our method significantly outperforms all recent feed-forward 3D models by an order of magnitude in both settings.}
\resizebox{0.4\textwidth}{!}{%
\begin{tabular}{llcc}
\toprule
Methods & ${\Delta R}_{\cal I\leftrightarrow\cal M}^\circ$ $\downarrow$ & ${\Delta R}_{\cal I\leftrightarrow\cal I}^\circ$ $\downarrow$\\ 
\midrule
DUSt3R & 54.40 & 29.27 \\
MASt3R & 24.18 & 12.52 \\
VGGT & 51.69 & 24.63 \\
$\pi^3$ & 68.46 & 45.80 \\
\midrule
Ours (COLMAP Init) & \textbf{2.59} & \textbf{1.48} \\

\bottomrule
\end{tabular}}
\label{table:geodesic_comparison}
\end{table}

Results are reported in Table~\ref{table:geodesic_comparison}. As shown, our method outperforms all baselines by an order of magnitude across both metrics, demonstrating that these models cannot cope with our challenging problem setting. Figure~\ref{fig:pi3}  illustrates a typical failure mode of meta-to-reference alignment. %
Additional qualitative results are provided in the supplementary. %

\subsection{Generalization to Different Reference Model}
\label{sec:generalization}
\begin{figure}
    \centering
    \jsubfig{\setlength{\fboxsep}{0pt}\fbox{\includegraphics[width=0.45\columnwidth]{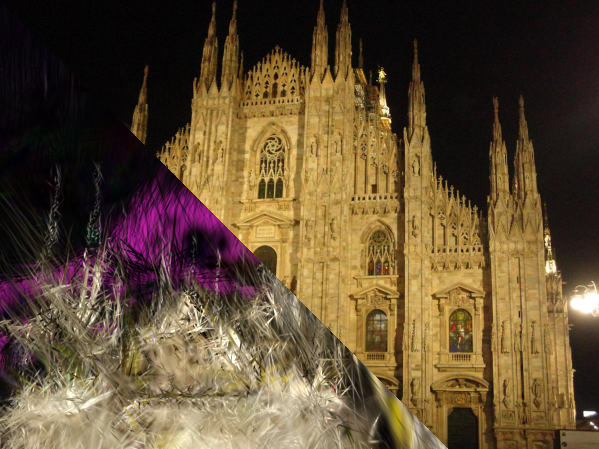}}}{}
    \jsubfig{\setlength{\fboxsep}{0pt}\fbox{\includegraphics[width=0.45\columnwidth]{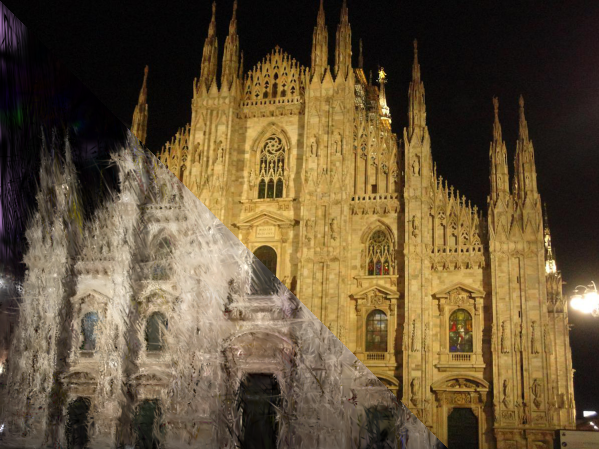}}}{} 
    \jsubfig{\setlength{\fboxsep}{0pt}\fbox{\includegraphics[width=0.45\columnwidth]{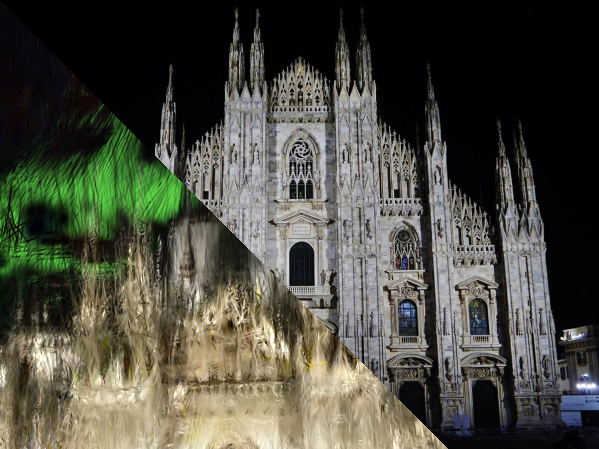}}}{COLMAP}
    \jsubfig{\setlength{\fboxsep}{0pt}\fbox{\includegraphics[width=0.45\columnwidth]{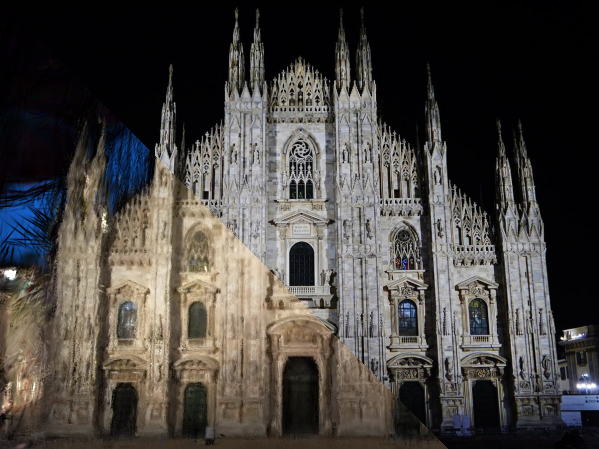}}}{Ours}
    \caption{\textbf{Generalization to Drone-Based Reference Models.} We evaluate our method using a reference model reconstructed from drone video frames sourced from YouTube. As illustrated above, our approach significantly improves the alignment over the COLMAP baseline, which serves as our initialization. %
    }
\label{drone_comparison}
\end{figure}

To demonstrate the generalization of our method to global reference models obtained from various sources, we created global reference
models from drone videos (downloaded from YouTube).
We constructed the reference models automatically by sampling the frames and recovering the trajectory using SfM. Results are presented in Figure \ref{drone_comparison}.

\subsection{Ablation Study}
\label{sec:ablation}
To assess the contributions of different components in our method, we conduct ablation studies on both the semantic features and robust optimization techniques.   These studies help isolate the effects of key optimizations and demonstrate the importance of each component. We conduct these ablations using the COLMAP initialization. The results of these ablations are summarized in \Cref{table:registration_results_table}. Additional ablations are provided in the supplementary.

\smallskip \noindent \textbf{Semantic Features Ablations:} We explore the impact of different photometric and feature-based optimization methods by replacing our primary semantic features with alternative approaches. This includes a photometric loss and other semantic feature extractors.
Photometric loss, commonly used in inverse optimization methods like iNeRF~\cite{yen2020inerf} and VF-NeRF~\cite{segre2024vfnerfviewshedfieldsrigid}, effectively aligns scenes through differentiable rendering in controlled environments. However, this method struggles on our “in-the-wild” dataset, as our results show it significantly underperforms, with high rotation error ($\Delta R$ of 6.48) and translation error ($\Delta T$ of 0.38), likely due to large color variations between meta-images and the low-quality reference model.

We also evaluate LSeg~\cite{li2022lseg}, which has been applied successfully in semantic segmentation tasks. While potentially beneficial in scenes with clear semantic structures, LSeg fails to outperform even the initialization (COLMAP), yielding a translation error of 0.34. This suggests that LSeg, though effective for segmentation, lacks the robustness needed for large-scale scene alignment in this challenging setting. 
Finally, we experiment with DINOv2 + DVT~\cite{yang2024denoisingvisiontransformers}, a variant of DINO designed to reduce grid-like artifacts in feature maps. While DINOv2 + DVT performs better than photometric loss and provides reasonable alignment results, we find that it falls short of the performance achieved using standard DINOv2 features, and therefore we did not utilize these semantic features in our pipeline.

\begin{table}
\setlength{\tabcolsep}{5pt}
 \def\arraystretch{0.95}
\centering
\caption{\textbf{Ablations}. We provide several experiments ablating our semantic features (middle) and robust optimization scheme (bottom). All the ablations where initialized with the COLMAP baseline. Best results are marked in bold. As illustrated below, the full method consistently outperforms all alternatives.
}

\resizebox{\linewidth}{!}{%
\begin{tabular}{llcccccccccccc}
\toprule
Methods & ${\Delta R^\circ}$ $\downarrow$ & ${\Delta T}$ $\downarrow$ & MTA\%$\uparrow$ & $O\%\downarrow$ \\ 
\midrule
Ours (COLMAP init) & \textbf{2.48} & \textbf{0.12} & \textbf{81} & \textbf{0}\\
\midrule
\emph{Semantic Features Ablations} \\
\quad Photometric Optimization & 6.48 & 0.38 & 72 & 22 \\
\quad LSeg ~\cite{li2022lseg}  & 4.78 & 0.34 & 62 & 19 \\
\quad DINOv2 + DVT ~\cite{yang2024denoisingvisiontransformers} & 2.86 & 0.14 & 78 & \textbf{0} \\
\midrule
\emph{Robust Optimization Ablations} \\
\quad w/o LTS & 3.78 & 0.19 & 69 & 3 \\
\quad Fixed LTS & 2.78 & 0.14 & 72 & \textbf{0} \\
\quad IRLS~\cite{daubechies2008irls} & 3.51 & 0.18 & 72 & 3 \\
\quad L2 & 4.21 & 0.20 & 75 & 3 \\
\bottomrule

\end{tabular}
}
\label{table:registration_results_table}
\end{table}

\smallskip \noindent \textbf{Robust Optimization Ablations:}
To evaluate the effect of our robust optimization scheme, we compare it to several baselines: (i) removing the robust optimization technique (w/o LTS), (ii) fixing the set of ignored images according to the first iteration in LTS (fixed LTS)  (iii) replacing it with an alternative robust optimization method (IRLS~\cite{daubechies2008irls}), and (iv) using $L2$ loss instead of $L1$. The ablation results are presented in the lower part of \Cref{table:registration_results_table}.

Both the IRLS and w/o LTS ablations show higher translation errors than the COLMAP baseline, highlighting the importance of LTS in optimizing alignment. Notably the ablation methods show up to 3\% outliers, indicating that most scenes converge, though not as effectively as with the full method.

The fixed LTS ablation also outperforms COLMAP but is slightly surpassed by the full approach. This indicates that a soft selection of images for optimization is more effective than a fixed cutoff, adjusting the ignored image set based on the current loss landscape.

We further analyze this phenomenon in the supplementary. In particular, we show histograms depicting the number of times images are ignored throughout the optimization procedure. Our analysis reveals that a majority of the images are consistently ignored (or not), while the relative loss of some images perturbs across iterations, further illustrating that our robust optimization scheme indeed yields a soft selection mechanism allowing for achieving stable convergence with improved performance. 

Our ablation studies highlight the importance of both semantic features and robust optimization techniques in achieving superior alignment. The results confirm that our full method consistently outperforms the ablation alternatives, with key optimizations such as LTS and semantic features based approaches playing critical roles in improving accuracy and convergence stability.

\section{Conclusion}
In this work, we proposed an approach for grounding partial 3D reconstructions \textit{in the wild} to a reference Gaussian Splatting model. Technically, this amounts to aligning in-the-wild images to a model obtained from pseudo-synthetic renderings. %
To solve it, we frame grounding as an inverse optimization problem and introduce a semantic feature-based robust optimization solution that is capable of handling outliers. Additionally we created \dataset{}, a new benchmark dataset that contains 3D reconstructions of landmarks associated internet photo collections previously assembled in the Wikiscenes dataset, registered with reference models obtained via images rendered from Google Earth Studio. 

As with other optimization-based approaches, our method remains sensitive to initialization, particularly when the initial alignment is far from the correct solution. Performance also becomes less reliable for very small image collections, which fall outside the regime our framework is designed to address. A more detailed analysis of these limitations is provided in the supplementary material. Future works can leverage the non overlapping aligned meta-images in \dataset{} to train sparse reconstruction pipelines. Another promising direction is to combine the global coverage of pseudo-synthetic imagery with the rich appearance details of Internet photographs to construct stronger hybrid 3D scene representations. Finally, incorporating language-based semantic features could unlock downstream applications such as language-guided scene grounding and navigation across large-scale environments. %

\paragraph{Acknowledgments}
This work was partially supported by the Israel Science Foundation grants 2510/23 and 2132/23.

{
    \small
    \bibliographystyle{ieeenat_fullname}
    \bibliography{main}
}

\setcounter{page}{1}
 \maketitlesupplementary

\setcounter{tocdepth}{2}

\medskip
\medskip
We refer readers to the interactive visualizations at the accompanying \url{Interactive Viewer/index.html} 
that show results for all presented baseline models (before and after our inverse optimization scheme) on the \dataset{} test set.

\section{Additional Results and Comparisons}
\label{sec:additional_results}

\subsection{Additional Quantitative Results}
In addition to the averaged $\Delta R$, $\Delta T$  reported in the main paper (Table 1), in Tables \ref{table:supp_results}, \ref{table:supp_results_gdls} \ref{table:supp_results_super_point}, we report a per meta-image performance breakdown for all the initializations. 
From this breakdown, we observe that our method successfully registers meta-images where the baseline exhibits large $\Delta R_{\cal I}$, $\Delta T_{\cal I}$ errors, such as lines 8, 13 and 14 in the COLMAP baseline table. However, there are cases where both our method and the baseline fail to register the images, as seen in lines 10, 31.

\begin{table}[th]
\centering
\resizebox{\linewidth}{!}{
\begin{filecontents*}{tables/supp/results.csv}
ID,Cathedral Number,Name,Meta Image ID,Mean Baseline Translation Error superglue,Mean Baseline Angle Error superglue,Mean Baseline Translation Error colmap,Mean Baseline Angle Error colmap,Mean Baseline Translation Error gpnps,Mean Baseline Angle Error gpnps,Mean Transform Translation Error dino,Mean Transform Angle Error dino,Mean Transform Translation Error gpnps_init,Mean Transform Angle Error gpnps_init,Mean Transform Translation Error superglue_init,Mean Transform Angle Error superglue_init
1,0,Milano Cathedral,3,,,0.216358316798243,4.11421341049569,0.227665864552565,2.87368067371237,0.347852415028894,4.8220063564493,0.359861105881194,4.99508898215909,,
2,0,Milano Cathedral,30,,,0.198009163455291,4.01917188483366,0.197109348306186,4.10176346775937,0.293317903196259,5.60960370223077,0.299200961537058,5.69082502025948,,
3,10,Sanfrancisco Cathedral,0,0.145980032064985,3.36736895361243,0.329484083974084,6.22383084838275,0.236717308420837,4.7103991147709,0.157311417602345,2.56201335762885,0.158858819964514,2.55149990501526,0.160618744961397,2.61170800510599
4,18,Lisbon Cathedral,0,0.0878058233109126,1.65899634902493,0.0642495092081175,1.83466582044221,0.083870263607884,1.21089288353757,0.135250795662792,1.9727154148324,0.13086560189801,1.84518556288894,0.13524618483155,1.98467342890003
5,20,Brussels Cathedral ,16,0.0190893620884898,1.25008394891762,0.00840022098439381,1.68691577731998,0.0163789009865442,0.394055968084616,0.0239689310633823,0.517518801403581,0.0239892203901215,0.519224281354939,0.0304213565958745,1.52066678290121
6,20,Brussels Cathedral ,15,0.0588375588729854,3.37618047391426,0.0509960964182475,1.66949972934355,0.0764139156479129,0.9527366797677,0.0756388926176158,0.924019223219032,0.075329415765612,0.896658818392434,0.0874843961828354,8.21744996186042
7,20,Brussels Cathedral ,0,0.0831137987086741,0.475768062730144,0.0853213312735731,0.631455167595809,0.0850311387907281,0.501940109357706,0.104659796464793,0.877261302077115,0.102542163988938,0.826826505490573,0.104587368853457,0.848079331531895
8,20,Brussels Cathedral ,13,0.0396758741032343,2.14513581222069,0.182148182046108,15.7225237422841,0.0320113244449913,0.599216365943644,0.0632057966599173,1.22819167401059,0.0566373397250532,1.14357498829834,0.0713423316775599,0.651820293637952
9,20,Brussels Cathedral ,14,0.103524181211809,11.5965204335975,0.0472161075028122,1.70789822606077,0.0709705050031603,0.925563097134666,0.075539921094314,0.909587458971953,0.075127923021714,0.900733092099661,0.0752225151750138,0.908161438721798
10,21,Salamanca Cathedral,1,2.00463491505404,17.9494176816405,0.515594744816829,11.6094167772306,0.873971256025639,21.296622258523,0.481014450016211,9.72794079510335,0.870182866749055,21.2630910209849,2.00625392778799,17.9970296889508
11,27,St John the Divine Cathedral,0,0.0114936981967714,0.477881159382589,0.00484519782184318,0.515430077587911,0.00515301713407466,0.224164811452952,0.0325071924067003,1.3412064317726,0.0337029138244526,1.37191928158232,0.0328795463990795,1.27568307817455
12,36,Burgos Cathedral,10,2.08053251644409,10.5030786702946,0.0194362577427925,0.921492378367739,0.0191750651738276,0.900511298176752,0.188627594282429,3.00003674363593,0.19653749267777,3.1271587346563,2.12405300597257,9.32329902245172
13,37,Aachen Cathedral,7,0.128378163466927,5.29905816506164,0.0855546827404453,41.4166589212809,0.0593877640984639,1.70467236325393,0.145612446371374,3.94430007981565,0.141915231828068,3.8069637758442,0.141149294908356,3.9362021234739
14,39,Basel Minster,4,0.156486847761008,4.46811050322039,0.558382070499572,24.0678399279591,0.643350142714106,17.0560123415533,0.212164041731982,3.74235992253467,0.186336022769736,2.94503072819499,0.211013296816892,4.29605680735052
15,43,Oviedo Cathedral,1,,,0.00550005266686554,0.254792083001718,0.00564196569345545,0.180895626539248,0.0889400434641493,1.38517260367445,0.0870366679975538,1.33980518746533,,
16,46,Palermo Cathedral,1,0.0258582517245635,1.06694053311471,0.0264209438407345,0.695895366866272,0.026671726239686,0.661026415669635,0.0892034117418579,1.32049071505612,0.0848271695312083,1.29683292869119,0.0841852120509176,1.23325692238286
17,46,Palermo Cathedral,0,0.0337954268819452,0.63986001464715,0.0180056210644412,0.378997045911703,0.0179655895643687,0.325352898344055,0.0443932650380741,0.716375787575054,0.0480351533437469,0.82104557021757,0.0475224083988689,0.805783120764957
18,50,Wells Cathedral,1,0.0210030094421142,0.510165519107975,0.21917400785558,3.0719700496761,0.0105531211060799,0.26847476303464,0.0950663211162648,2.13763102859182,0.0964263606935207,2.18084659521692,0.0879800885617104,2.07928439509692
19,52,Lincoln Cathedral,2,,,0.0238321897339948,0.493337702810936,0.0254026009974689,0.390155526177866,0.0505983891363982,0.515710634123812,0.0497420633725449,0.500748713489269,,
20,52,Lincoln Cathedral,11,,,0.0171758675948895,1.07612557823963,,,0.0409790697626613,1.36751555341252,,,,
21,53,Monreale Cathedral,0,0.0247543072747605,0.577850192870948,0.00708205755722037,0.518284373016173,0.00629573368748365,0.178879134891993,0.0324837957376671,0.47566097412318,0.0333271311981705,0.507656812541103,0.0324827921881756,0.463764507341906
22,54,Rouen Cathedral,0,0.0725408249282044,1.88135658204826,0.0214233969707272,1.09929149586496,0.0335297285878717,1.36957697453334,0.0824478770175709,2.62211799469075,0.0812816554935736,2.59206533643909,0.07437887175225,2.50108925752725
23,55,Geneva Cathedral,2,0.0745991872110189,2.50927254312039,0.0259075067437855,3.80808885900385,0.0177979180196075,0.748491769035192,0.251730302132927,6.61669896862396,0.0146059092537628,0.708548532983512,0.0742654464699074,2.59653001680939
24,56,Ulm Minster,1,0.128216328588135,2.1815136639952,0.224749787243675,4.46788028384545,0.224263062798458,4.24003461428754,0.188574839218059,3.96487759448817,0.218399228983465,4.37240427410063,0.185066335488459,3.7523135893627
25,61,Bordeaux Cathedral,3,0.0697154977097381,3.51693293578884,0.0112636850847856,0.254375559700337,0.0115336397982292,0.303256289101021,0.0889213668114478,2.71338902447984,0.088061473028021,2.67672450758092,0.0939788447916331,2.71505640846903
26,61,Bordeaux Cathedral,0,0.568804194204074,4.97602475940496,0.291018720006044,8.80996596176937,0.28927549085204,8.11820497340698,0.0948513159034171,2.5994831856449,0.100467553062686,2.58631583459732,0.102028822869184,2.61553931288946
27,75,Murcia Cathedral,0,0.0555597378918607,0.926220740079754,0.0514362657961406,1.39847073869726,0.051400730941202,1.29805891931836,0.0390811930045624,0.649852786739108,0.0410635188916447,0.686677261147469,0.0445044799735608,0.696428422835781
28,85,Metz Cathedral,2,0.368978741463697,5.80085176526128,0.159494733096143,2.88296504641805,0.15883363966066,2.88195241399907,0.0671853512035776,0.479758821966797,0.0662773527484703,0.486196599722713,0.0858542608525127,0.63611680921037
29,85,Metz Cathedral,1,0.0718845871942682,1.24958359162221,0.142675886308392,5.19615982407133,0.0654026633820166,0.703691101699191,0.0979339263891534,1.21366122185692,0.0977085798505228,1.22650960023258,0.0933351906692028,1.00310162958218
30,90,Avila Cathedral,1,0.0981633443700937,4.52298969121592,0.0354732989284034,1.32540515735571,0.0362268756175619,1.34901175530291,0.0601988766662346,2.43722895959179,0.0590618633700383,2.42794435809762,0.0741010634818441,2.94684749124843
31,93,salisbury cathedral,4,0.18597016866658,7.06354397274379,0.17312856165264,6.77050730069608,0.213706443033471,7.25851339110549,0.214714507424215,6.53720105601399,0.214491912383582,6.55756691509872,0.213823578331941,6.53046304085619
32,97,Napoli Cathedral,1,0.0208176077619591,0.921612794819947,0.0175067759007795,0.940577507292396,0.0195758935989031,0.862202172400653,0.0233131031483681,0.476414530600155,0.0245744379385118,0.472709974803804,0.0237096334814811,0.471959295845195
\end{filecontents*}
\pgfplotstabletypeset[
    col sep=comma, %
    header=true, %
    every head row/.style={before row=\hline, after row=\hline}, %
    every last row/.style={after row=\hline}, %
    columns={ID, Cathedral Number,Name, Meta Image ID, Mean Baseline Translation Error colmap, Mean Baseline Angle Error colmap, Mean Transform Translation Error dino, Mean Transform Angle Error dino}, %
    columns/Name/.style={string type},
    columns/Cathedral Number/.style={column name=WikiScenes ID, string type},
    columns/Mean Baseline Translation Error colmap/.style={column name=Baseline ${\Delta T_{\cal I}}$, fixed, fixed zerofill, precision=3},
    columns/Mean Baseline Angle Error colmap/.style={column name=Baseline ${\Delta R_{\cal I}}$, fixed, fixed zerofill, precision=3 },
    columns/Mean Transform Translation Error dino/.style={column name=Our ${\Delta T_{\cal I}}$, fixed, fixed zerofill, precision=3},
    columns/Mean Transform Angle Error dino/.style={column name=Our ${\Delta R_{\cal I}}$, fixed, fixed zerofill, precision=3},
]{tables/supp/results.csv}
}
\caption{\textbf{Performance Breakdown Per Meta-Image}. Performance of our method (initialized with COLMAP) and the COLMAP baseline per meta-image, considering only meta-images where the baseline did not fail.}
\label{table:supp_results}
\end{table}

\begin{table}[th]
\centering
\resizebox{\linewidth}{!}{
\pgfplotstabletypeset[
    col sep=comma, %
    header=true, %
    every head row/.style={before row=\hline, after row=\hline}, %
    every last row/.style={after row=\hline}, %
    columns={ID, Cathedral Number,Name, Meta Image ID, Mean Baseline Translation Error gpnps, Mean Baseline Angle Error gpnps, Mean Transform Translation Error gpnps_init, Mean Transform Angle Error gpnps_init}, %
    columns/Name/.style={string type},
    columns/Cathedral Number/.style={column name=WikiScenes ID, string type},
    columns/Mean Baseline Translation Error gpnps/.style={column name=Baseline ${\Delta T_{\cal I}}$, fixed, fixed zerofill, precision=3},
    columns/Mean Baseline Angle Error gpnps/.style={column name=Baseline ${\Delta R_{\cal I}}$, fixed, fixed zerofill, precision=3 },
    columns/Mean Transform Translation Error gpnps_init/.style={column name=Our ${\Delta T_{\cal I}}$, fixed, fixed zerofill, precision=3},
    columns/Mean Transform Angle Error gpnps_init/.style={column name=Our ${\Delta R_{\cal I}}$, fixed, fixed zerofill, precision=3},
]{tables/supp/results.csv}
}
\caption{\textbf{Performance Breakdown Per Meta-Image gDLS+++ initialization}. Performance of our method (initialized with gDLS+++) and the gDLS+++ baseline per meta-image}
\label{table:supp_results_gdls}
\end{table}

\begin{table}[th]
\centering
\resizebox{\linewidth}{!}{
\pgfplotstabletypeset[
    col sep=comma, %
    header=true, %
    every head row/.style={before row=\hline, after row=\hline}, %
    every last row/.style={after row=\hline}, %
    columns={ID, Cathedral Number,Name, Meta Image ID, Mean Baseline Translation Error superglue, Mean Baseline Angle Error superglue, Mean Transform Translation Error superglue_init, Mean Transform Angle Error superglue_init}, %
    columns/Name/.style={string type},
    columns/Cathedral Number/.style={column name=WikiScenes ID, string type},
    columns/Mean Baseline Translation Error superglue/.style={column name=Baseline ${\Delta T_{\cal I}}$, fixed, fixed zerofill, precision=3},
    columns/Mean Baseline Angle Error superglue/.style={column name=Baseline ${\Delta R_{\cal I}}$, fixed, fixed zerofill, precision=3 },
    columns/Mean Transform Translation Error superglue_init/.style={column name=Our ${\Delta T_{\cal I}}$, fixed, fixed zerofill, precision=3},
    columns/Mean Transform Angle Error superglue_init/.style={column name=Our ${\Delta R_{\cal I}}$, fixed, fixed zerofill, precision=3},
]{tables/supp/results.csv}
}
\caption{\textbf{Performance Breakdown Per Meta-Image SP+LG initialization}. Performance of our method (initialized with SP+LG) and the SP+LG baseline per meta-image}
\label{table:supp_results_super_point}
\end{table}

As mentioned in the main paper, there are cases where the baselines fails and does not output any transformation. Empty lines in the results tables indicate baseline failures. Upon a closer look of the COLMAP and SP+LG baseline failures, we found that it fails because it is unable to register the minimum of three images to the reference model, which is required for the global transform estimation.

Additionally \cref{table:mta_thresholds} \cref{table:outlier_thresholds} present the MTA and O\% scores across various thresholds. These tables show our method consistently outperforms the baseline across all thresholds, considering both the MTA and O\%. 

\begin{table}[t]
\setlength{\tabcolsep}{5pt}
 \def\arraystretch{0.95}
\centering
\resizebox{\linewidth}{!}{%
\begin{tabular}{llcccccccccccc}
\toprule
Methods & MTA$_{(3,0.1)}$ & MTA$_{(7,0.1)}$ & MTA$_{(10,0.1)}$ & MTA$_{(5, 0.09)}$ & MTA$_{(5, 0.15)}$ & MTA$_{(5,0.2)}$ \\ 
\midrule
Ours (Colmap init) & 62 & 62 & 62 & 53 & 72 & 81 \\
Colmap Baseline & 56 & 59 & 59 & 59 & 59 & 66 \\
\bottomrule
\end{tabular}
}
\caption{\textbf{Performance over Different MTA Thresholds}. Above, we report MTA scores over different threshold values for both the baseline and our model. Note that MTA$_{(r, t)}$ considers an image as accurately registered if $\Delta R < r$ and $ \Delta T < t$.  }
\label{table:mta_thresholds}
\end{table}

\begin{table}[t]
\setlength{\tabcolsep}{5pt}
\def\arraystretch{0.95}
\centering
\resizebox{\linewidth}{!}{%
\begin{tabular}{llcccccccccccc}
\toprule
Methods & $O\%_{(10,0.3)}$ & $O\%_{(10,0.4)}$ & $O\%_{(10,0.6)}$ & $O\%_{(15,0.5)}$ & $O\%_{(17,0.5)}$ & $O\%_{(20, 0.5)}$ \\ 
\midrule
Ours (Colmap init) & 6 & 3 & 0 & 0 & 0 & 0 \\
Colmap Baseline & 16 & 12 & 12 & 12 & 9 & 9 \\
\bottomrule
\end{tabular}
}
\caption{\textbf{Performance over Different O\% Thresholds}. Above, we report outlier scores over different threshold values for both the baseline and our model. Note that $O\%_{(r, t)}$ considers an image as an outlier if $\Delta R > r$ or $ \Delta T > t$. }
\label{table:outlier_thresholds}
\end{table}

\subsection{Additional Qualitative Comparison} 

\label{sec:master_comparison_supp}

\begin{figure}
\rotatebox{90}{\whitetxt{aaa} DUSt3R}
\jsubfig{\includegraphics[height=2.2cm, width=0.3\linewidth]{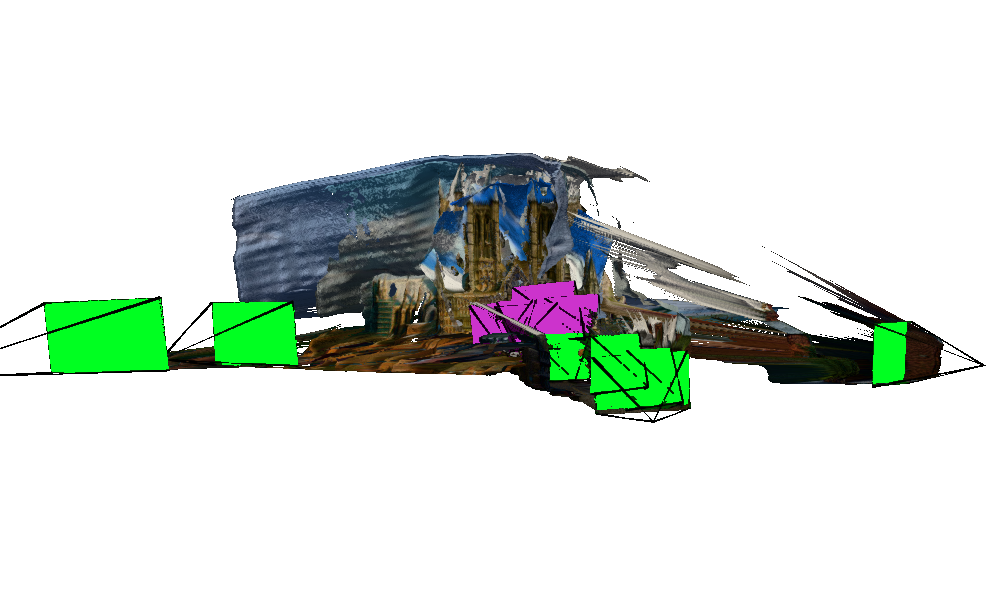}}{}
\jsubfig{\includegraphics[height=2.2cm, width=0.3\linewidth, trim=600 0 0 200, clip]{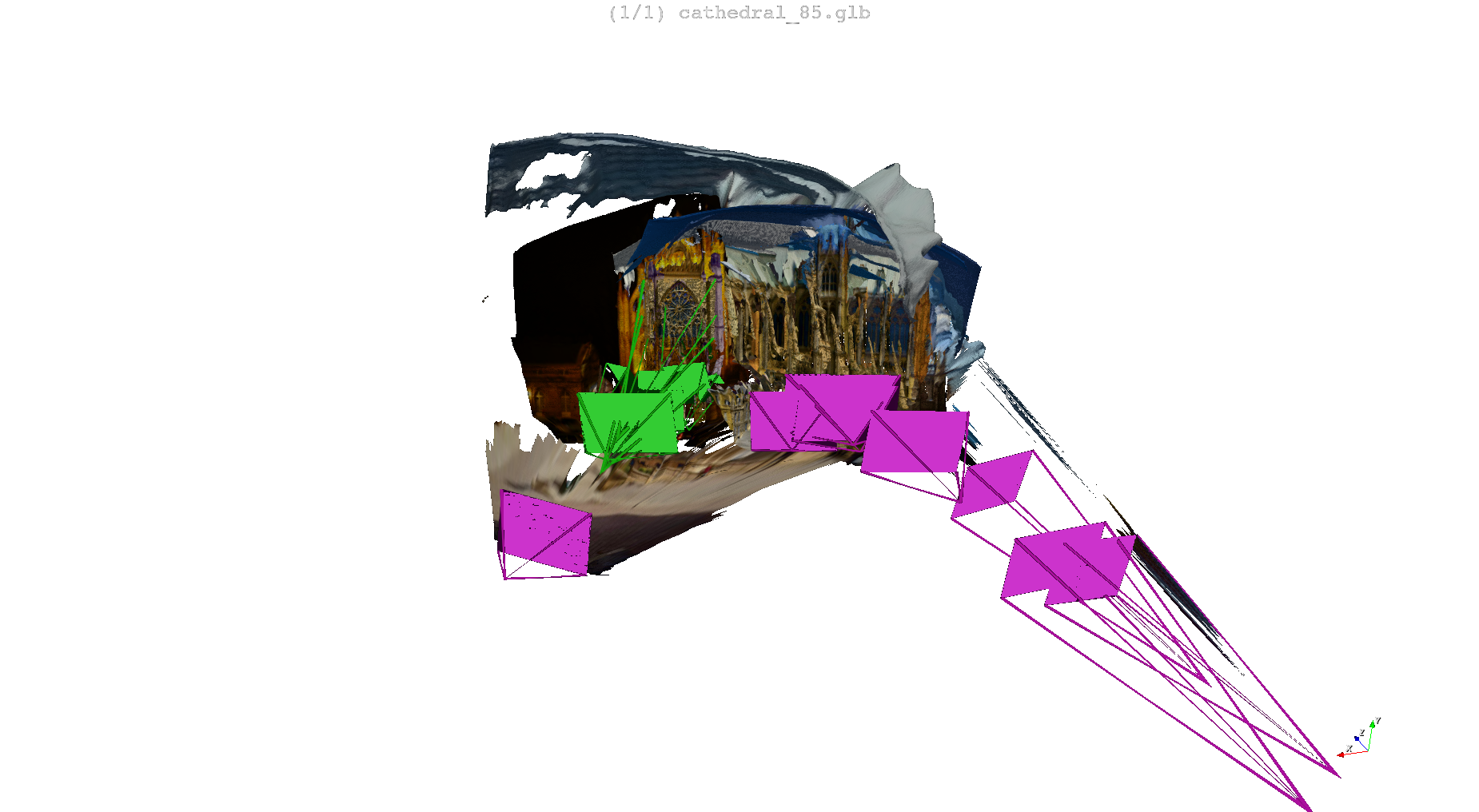}}{}
\jsubfig{\includegraphics[height=2.2cm, width=0.3\linewidth, trim=400 200 400 0, clip]{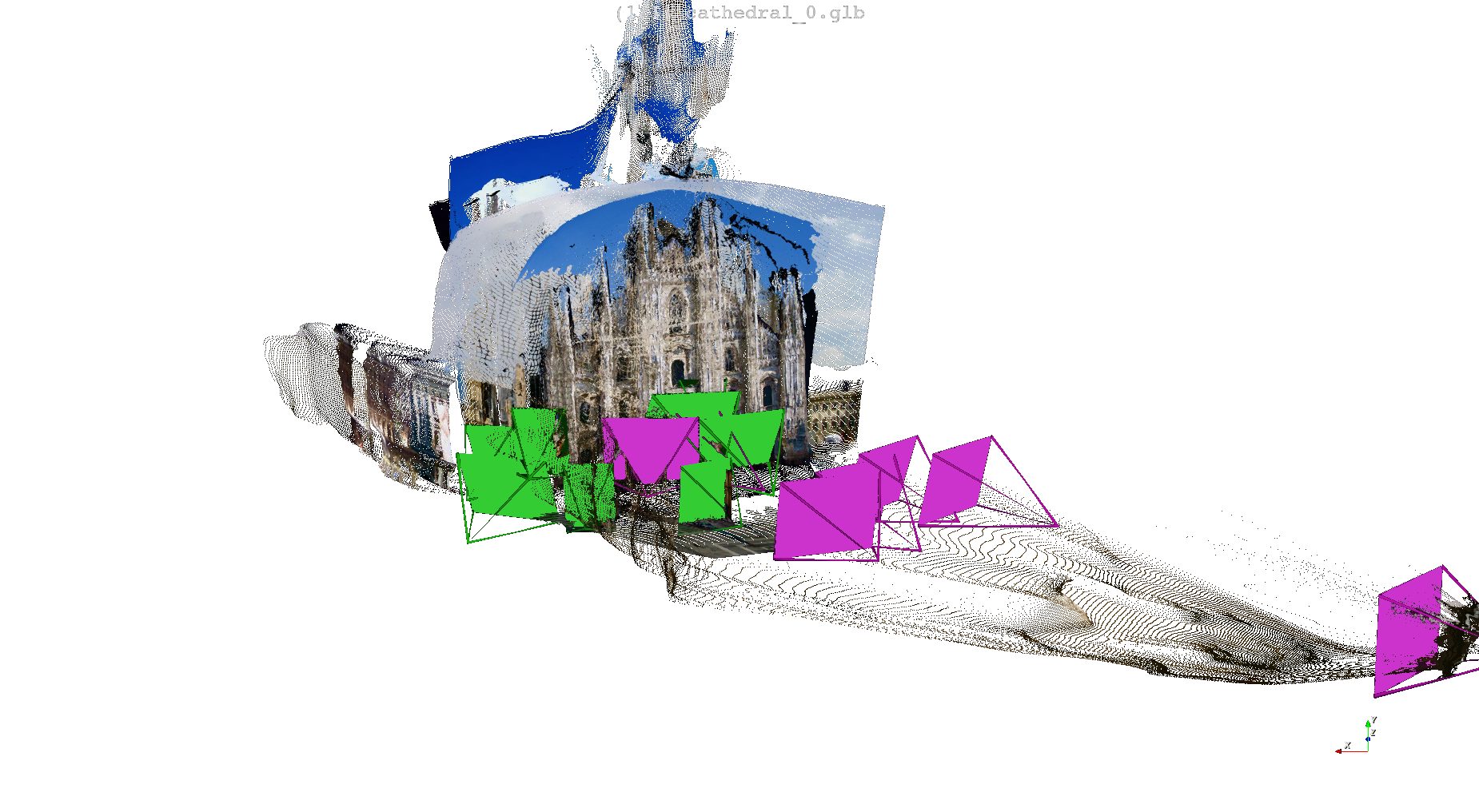}}{}
\\
\rotatebox{90}{\whitetxt{aaa}MASt3R}
\jsubfig{\includegraphics[height=2.1cm, width=0.3\linewidth]{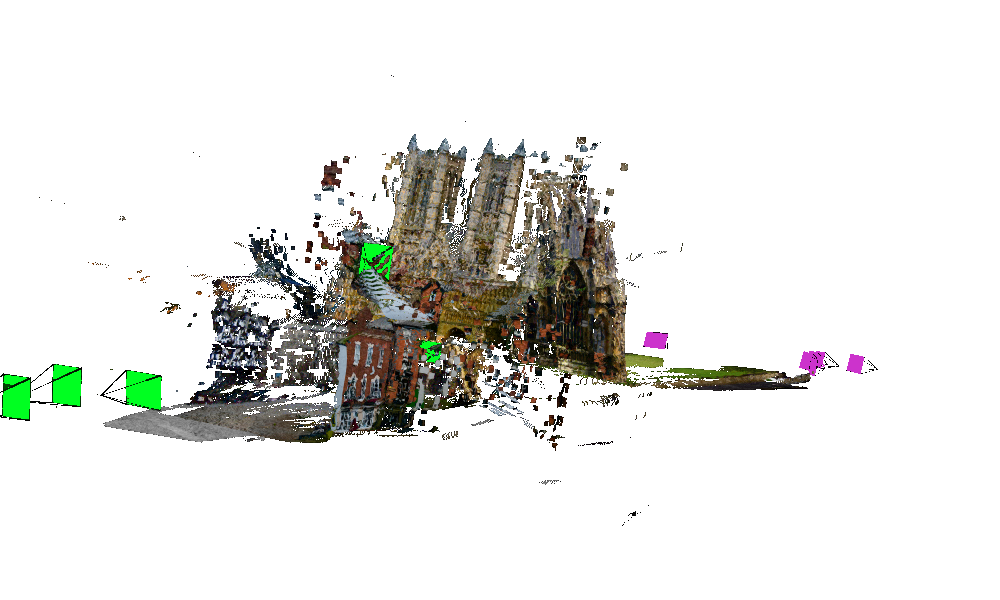}}{}
\jsubfig{\includegraphics[height=2.1cm, width=0.3\linewidth, trim=200 0 0 50, clip]{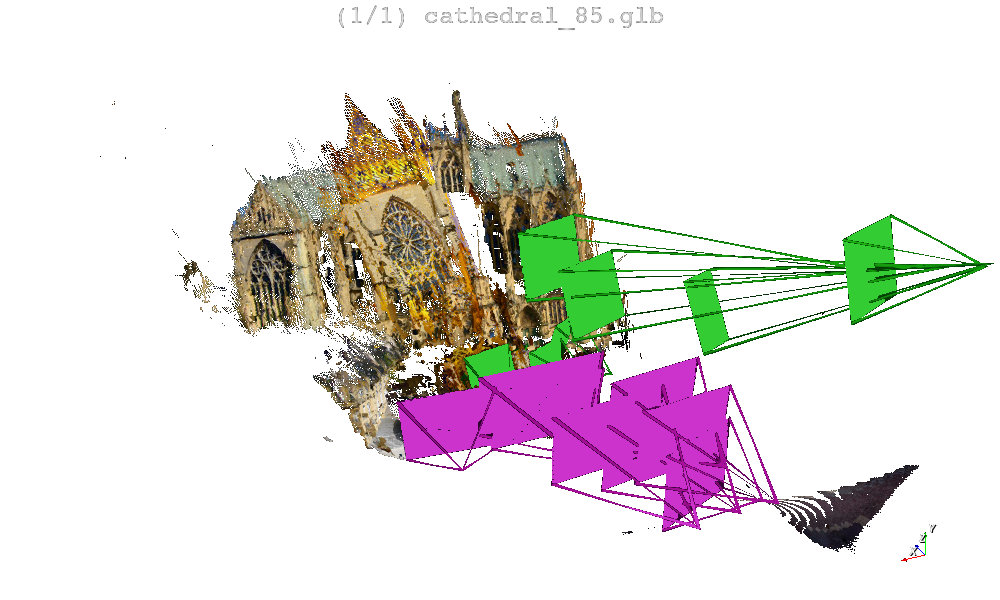}}{}
\jsubfig{\includegraphics[height=2.1cm, width=0.3\linewidth, trim=400 0 200 50, clip]{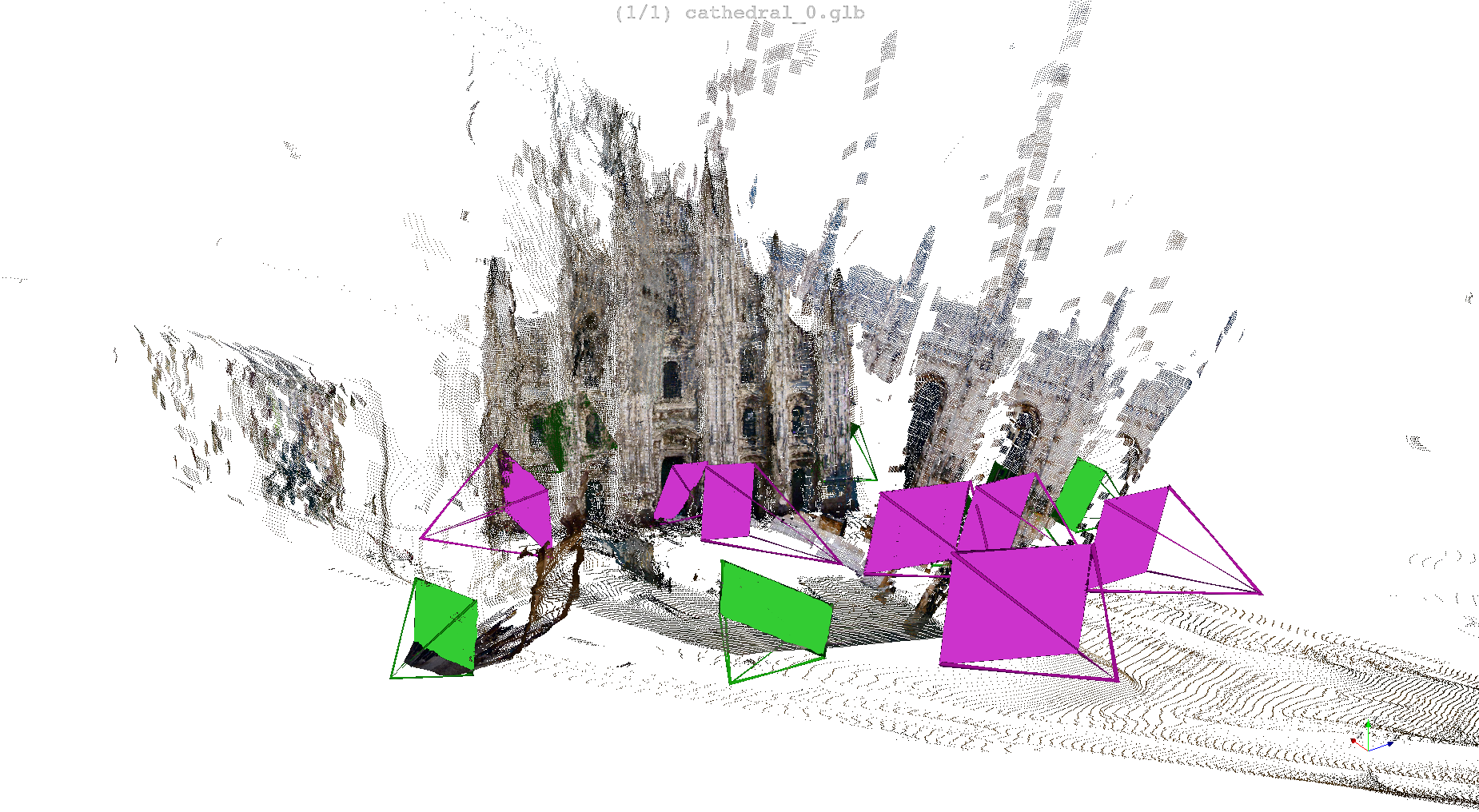}}{}
\\
\rotatebox{90}{\whitetxt{aaa}VGGT}
\jsubfig{\includegraphics[height=1.65cm, width=0.3\linewidth]{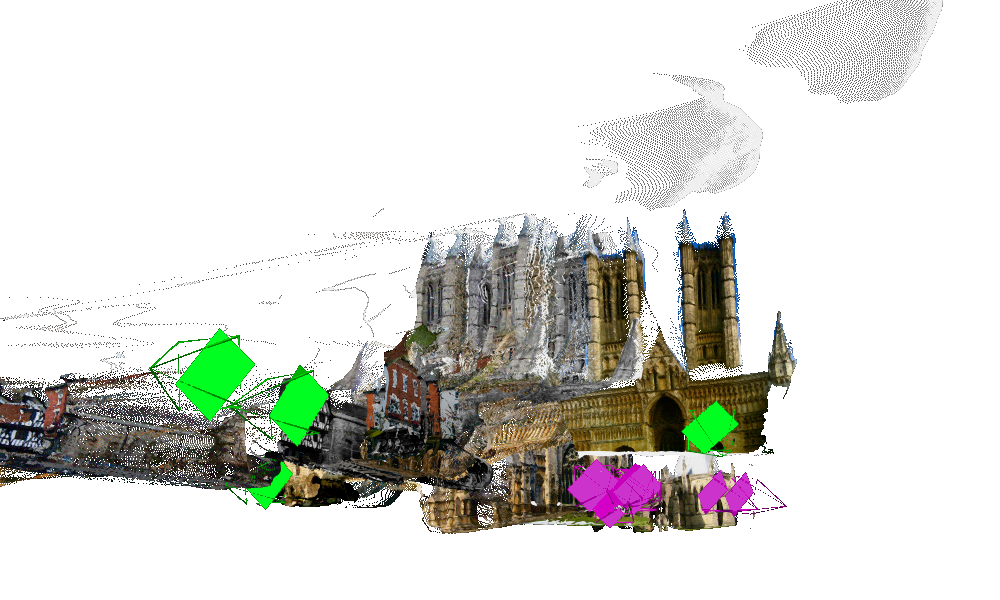}}{\small{Lincoln Cathedral}}
\jsubfig{\includegraphics[height=1.65cm, width=0.3\linewidth]{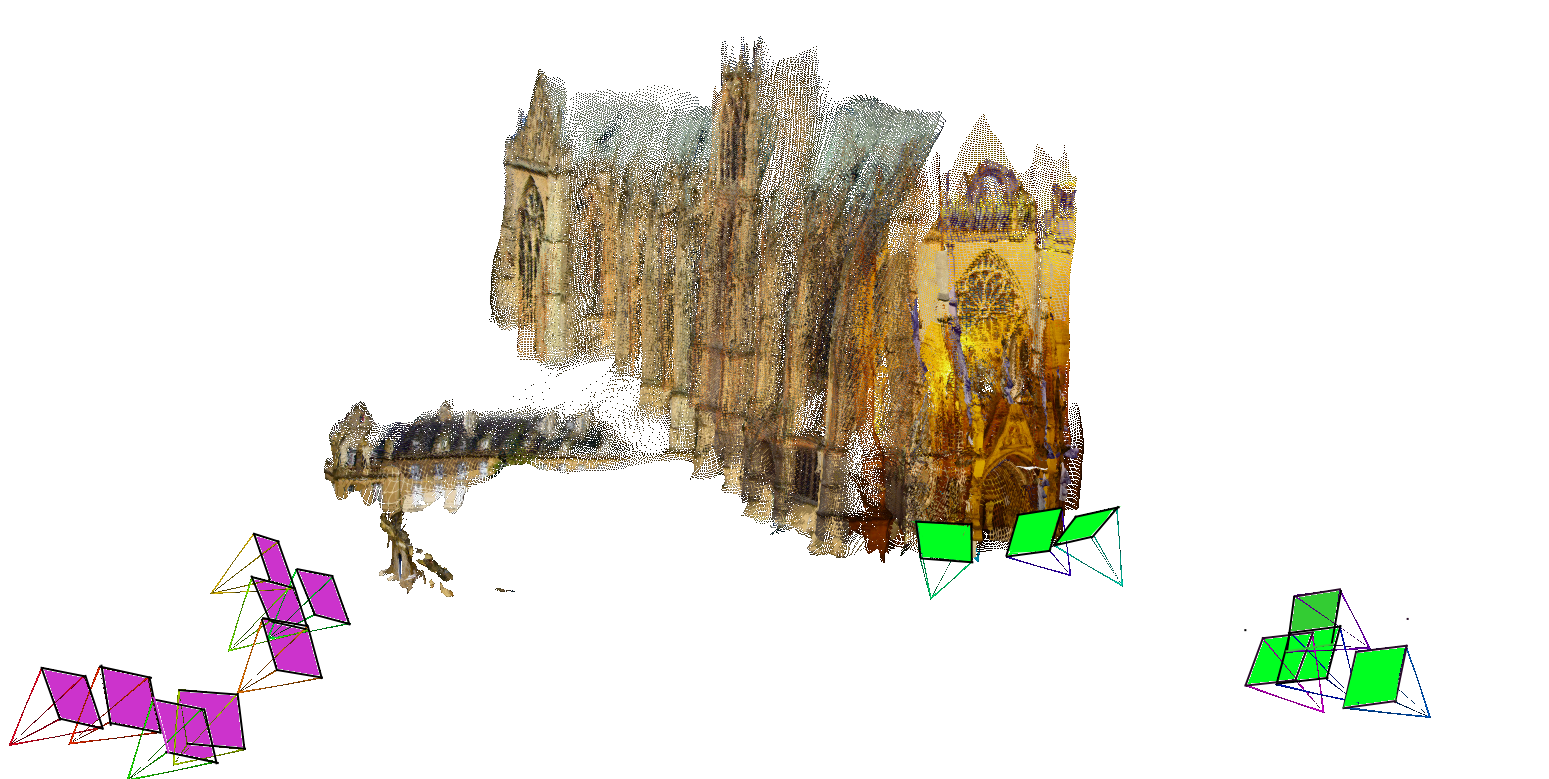}}{\small{Metz Cathedral}}
\jsubfig{\includegraphics[height=1.65cm, width=0.3\linewidth]{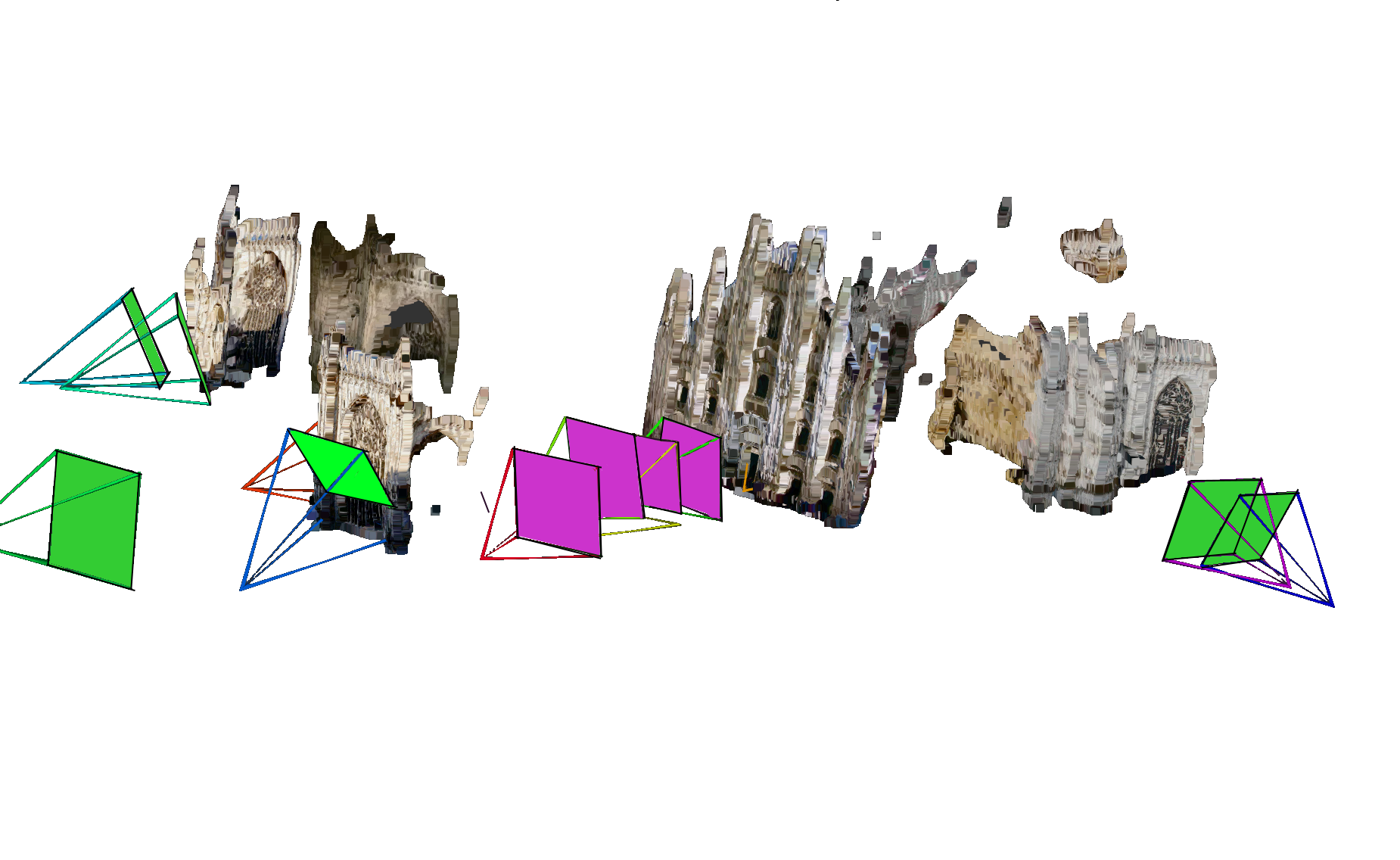}}{\small{Milan Cathedral}}

\caption{\textbf{Reconstructions of Feed-Forward Models}. We visualize three reconstructions obtained by running DUSt3R ~\cite{wang2024dust3r} MASt3R ~\cite{leroy2024mast3r} and VGGT ~\cite{wang2025vggt} over images sampled from two meta-images (visualized in \textcolor{green_teaser}{\bf green} and \textcolor{magenta_teaser}{\bf purple}). %
As illustrated above, all methods fail to reconstruct the Milan and Lincoln Cathedral, showing either broken or overlapping meta-images, while these cameras capture non-overlapping regions as as seen in the ground truth reconstructions (see illustration in the main paper).}
\label{fig:master_comparison}
\end{figure}

\begin{figure*}
    \jsubfig{\includegraphics[width=0.3\textwidth]{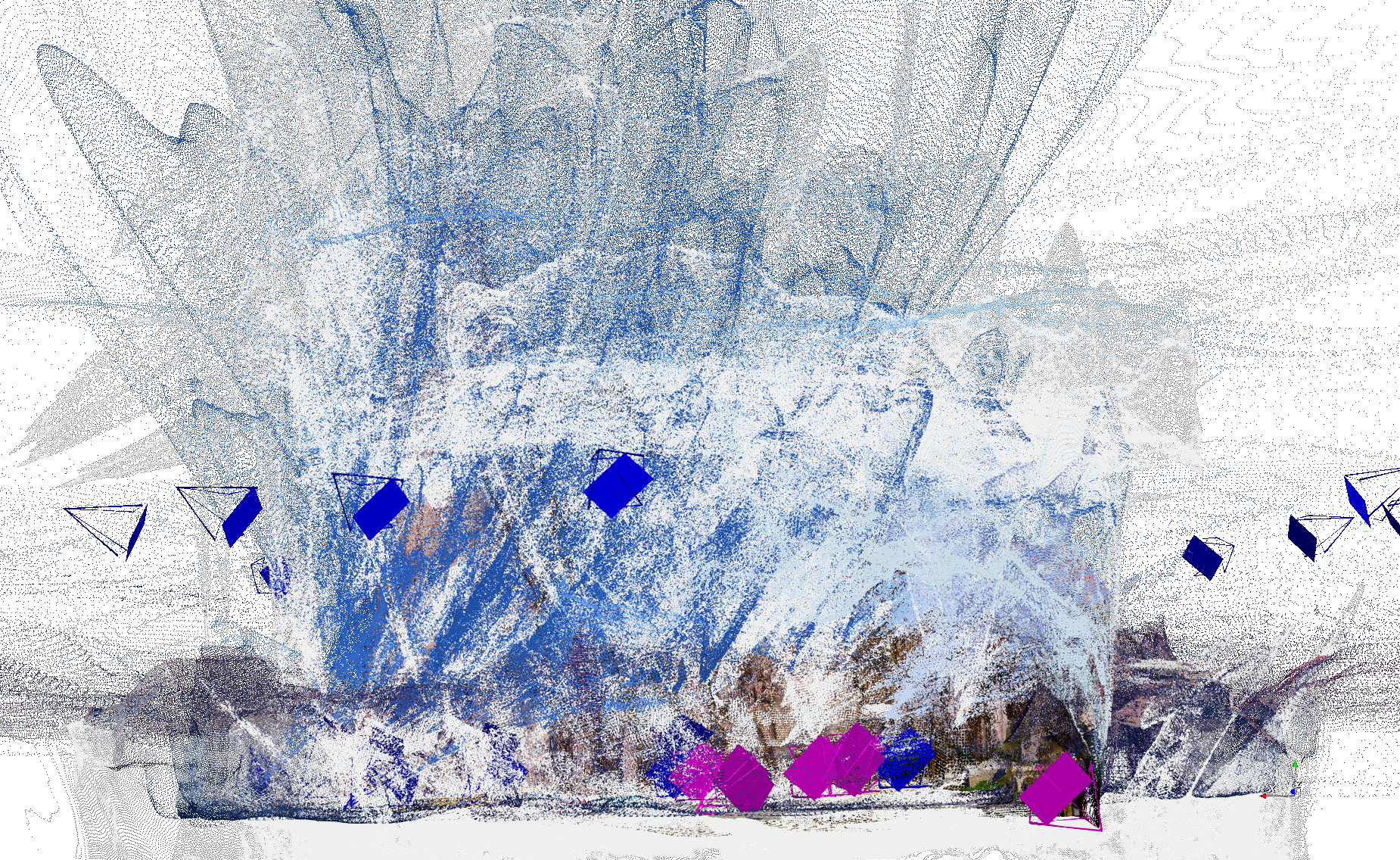}}{Freiburg Cathedral}
    \jsubfig{\includegraphics[width=0.3\textwidth]{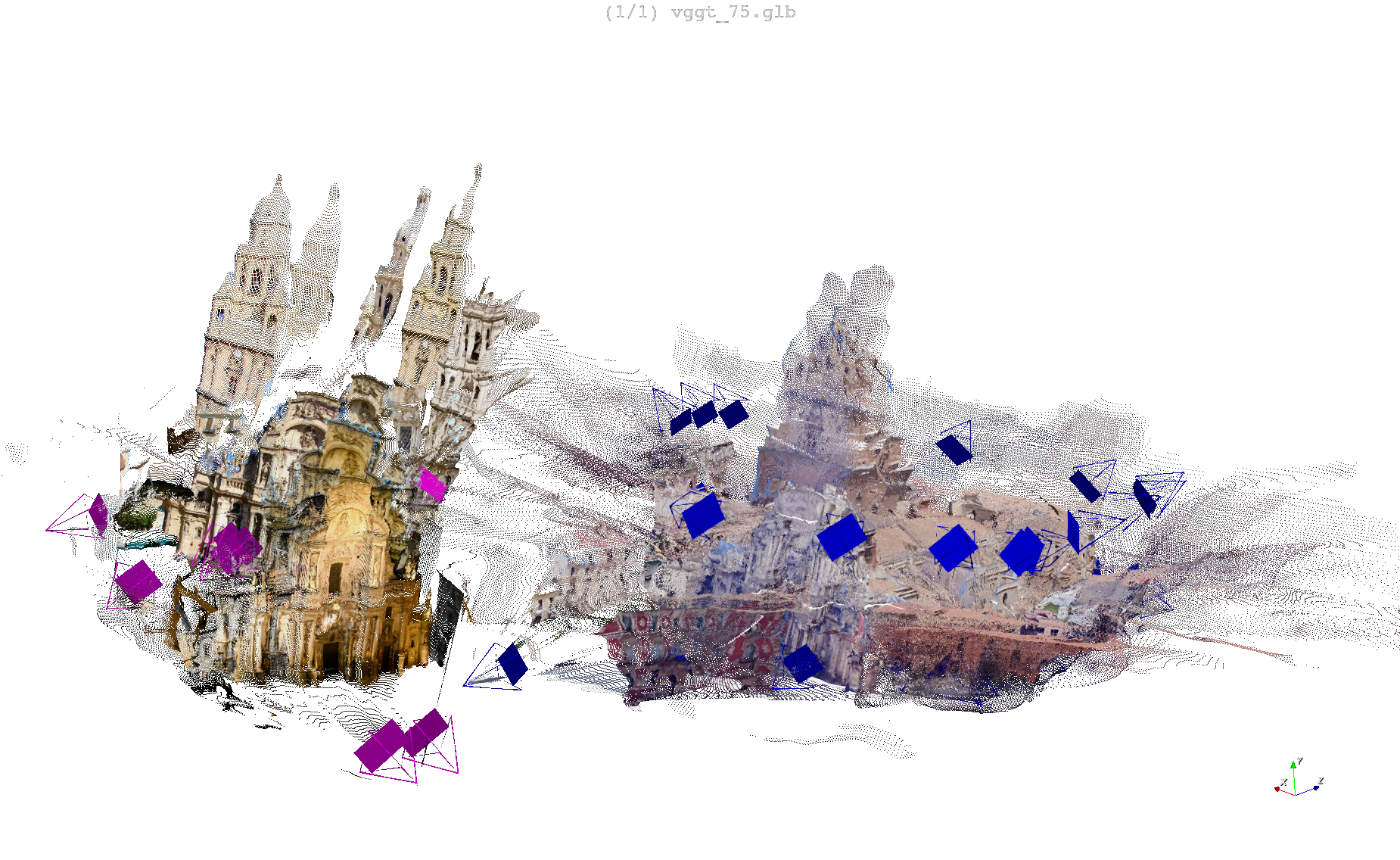}}{Murcia Cathedral}
    \jsubfig{\includegraphics[width=0.3\textwidth]{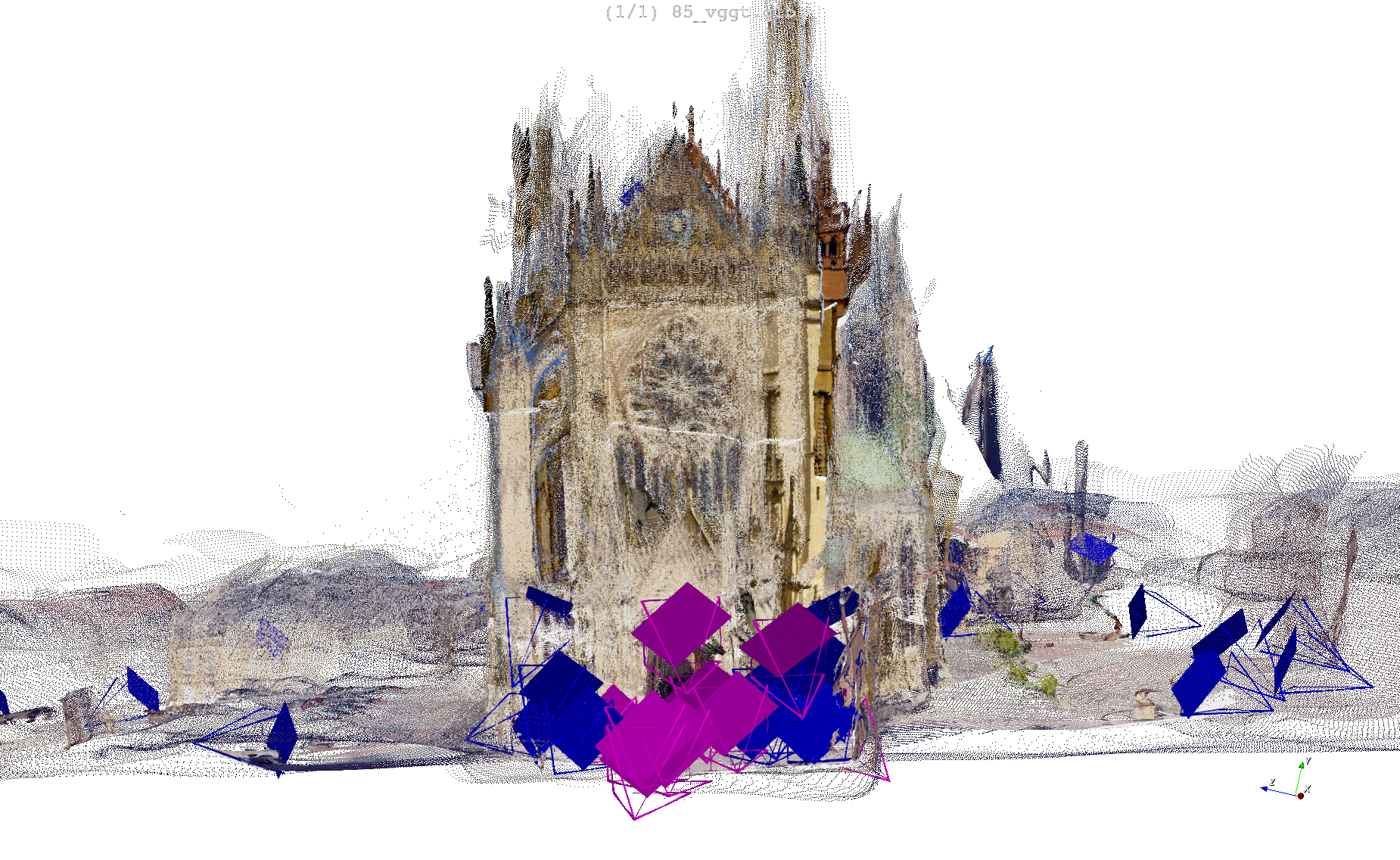}}{Metz Cathedral}

\caption{\textbf{Aligning a meta-image to a reference model with VGGT}. Meta-image cameras are visualized in purple, while Google Earth images from the reference model are in blue. As illustrated above, VGGT failed to register the Murcia meta-image (\emph{i.e.}, the output contains two distinct regions, one for the Internet images and one for the Google Earth images) and also failed to reconstruct the reference model in the Freiburg Cathedral. For the Metz Cathedral, VGGT produced a slightly misaligned registration, as evident by the ghost structures near the top of the building.}
\label{fig:master_comparison_sup}
\end{figure*}

In the paper, we qualitatively show that $\pi^3$ struggles to register meta-image to reference model, illustrating that existing sparse reconstruction approaches cannot address the task we propose. In \cref{fig:master_comparison_sup}, we present typical failures cases of VGGT when registering a meta-image to a reference model. In \cref{fig:master_comparison}, we qualitatively show failure cases of the feed-forward methods when registering non-overlapping meta-images.

\begin{figure}
    \centering
    \jsubfig{\setlength{\fboxsep}{0pt}\fbox{\includegraphics[width=0.45\columnwidth]{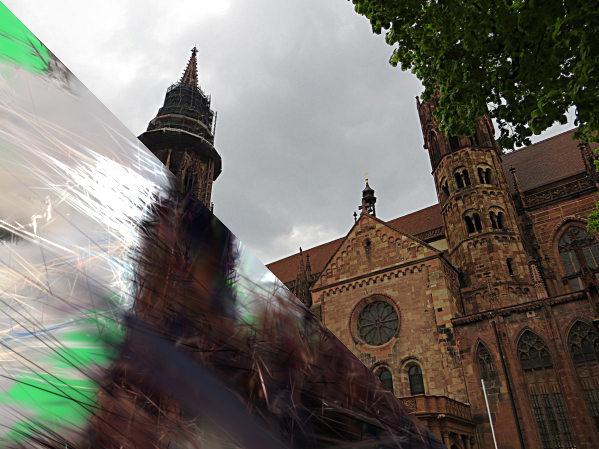}}}{}
    \jsubfig{\setlength{\fboxsep}{0pt}\fbox{\includegraphics[width=0.45\columnwidth]{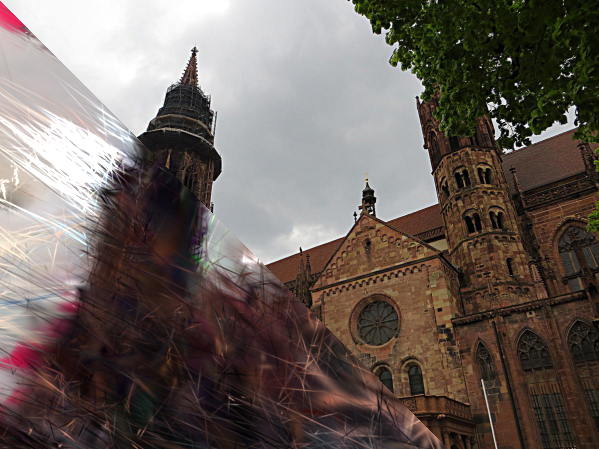}}}{} 
    \jsubfig{\setlength{\fboxsep}{0pt}\fbox{\includegraphics[width=0.45\columnwidth]{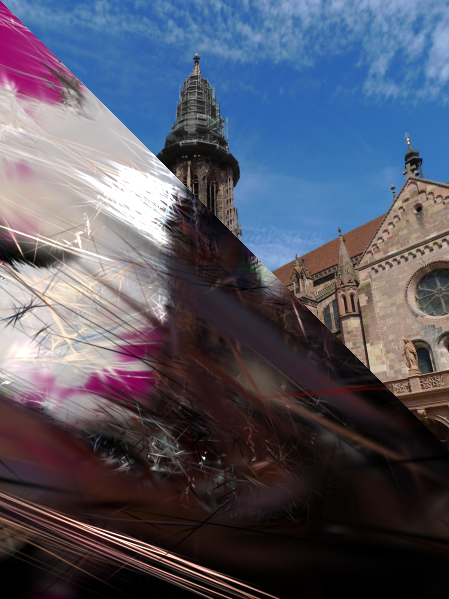}}}{COLMAP}
    \jsubfig{\setlength{\fboxsep}{0pt}\fbox{\includegraphics[width=0.45\columnwidth]{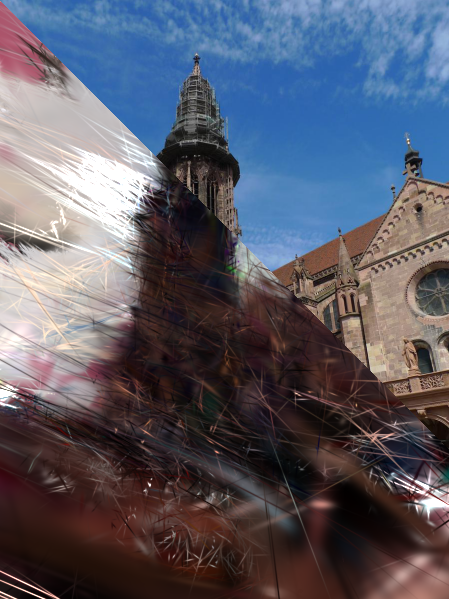}}}{Ours}
    \caption{\textbf{Drone Reference model - Additional examples} Results using a reference
model from drone video frames are depicted above. The drone videos of the Freiburg Cathedral are taken from Youtube. As illustrated our approach significantly improves the alignment, in comparison to the COLMAP baseline, which serves as our initialization.}
\label{drone_comparison_supp}
\end{figure}

\begin{figure}
    \centering
    \rotatebox{90}{\whitetxt{g}Gate}
    \jsubfig{\setlength{\fboxsep}{0pt}\fbox{\includegraphics[width=0.45\columnwidth]{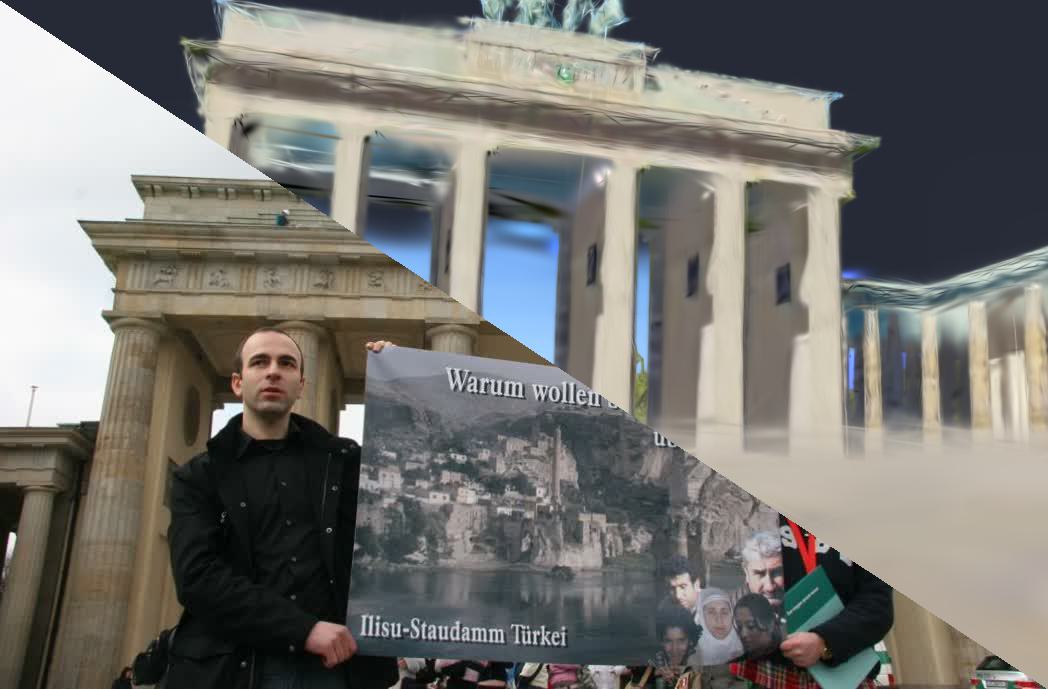}}}{}
    \jsubfig{\setlength{\fboxsep}{0pt}\fbox{\includegraphics[width=0.45\columnwidth]{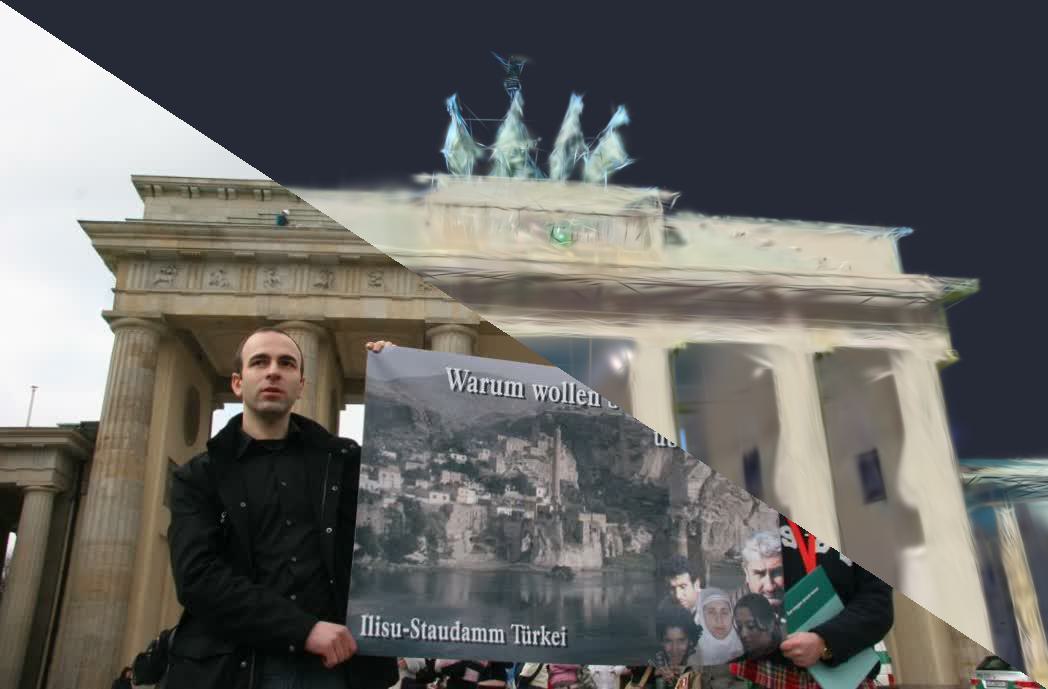}}}{} 
    \rotatebox{90}{\whitetxt{aaaa}Brandenburg}
    \jsubfig{\setlength{\fboxsep}{0pt}\fbox{\includegraphics[width=0.45\columnwidth]{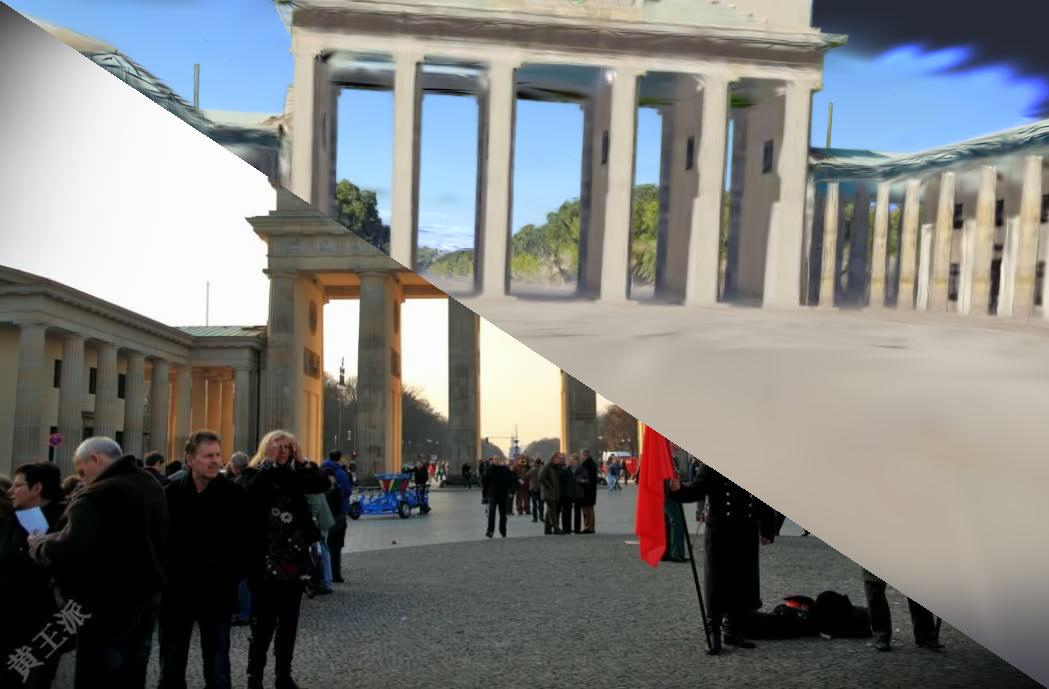}}}{}
    \jsubfig{\setlength{\fboxsep}{0pt}\fbox{\includegraphics[width=0.45\columnwidth]{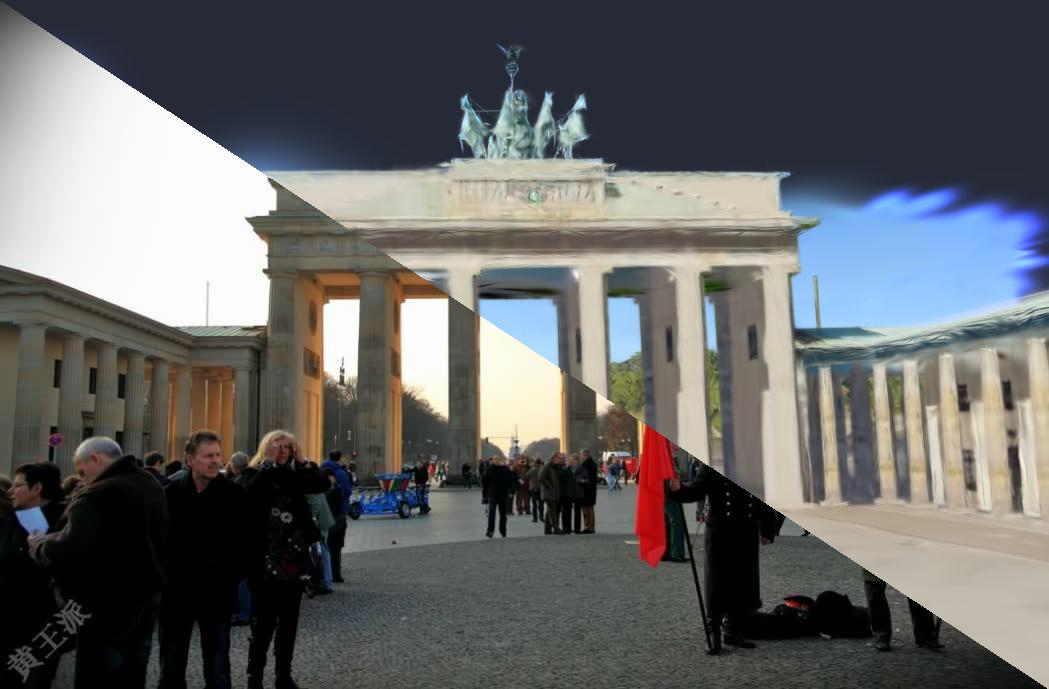}}}{}{}
    \rotatebox{90}{\whitetxt{g}Palace}
    \jsubfig{\setlength{\fboxsep}{0pt}\fbox{\includegraphics[width=0.45\columnwidth]{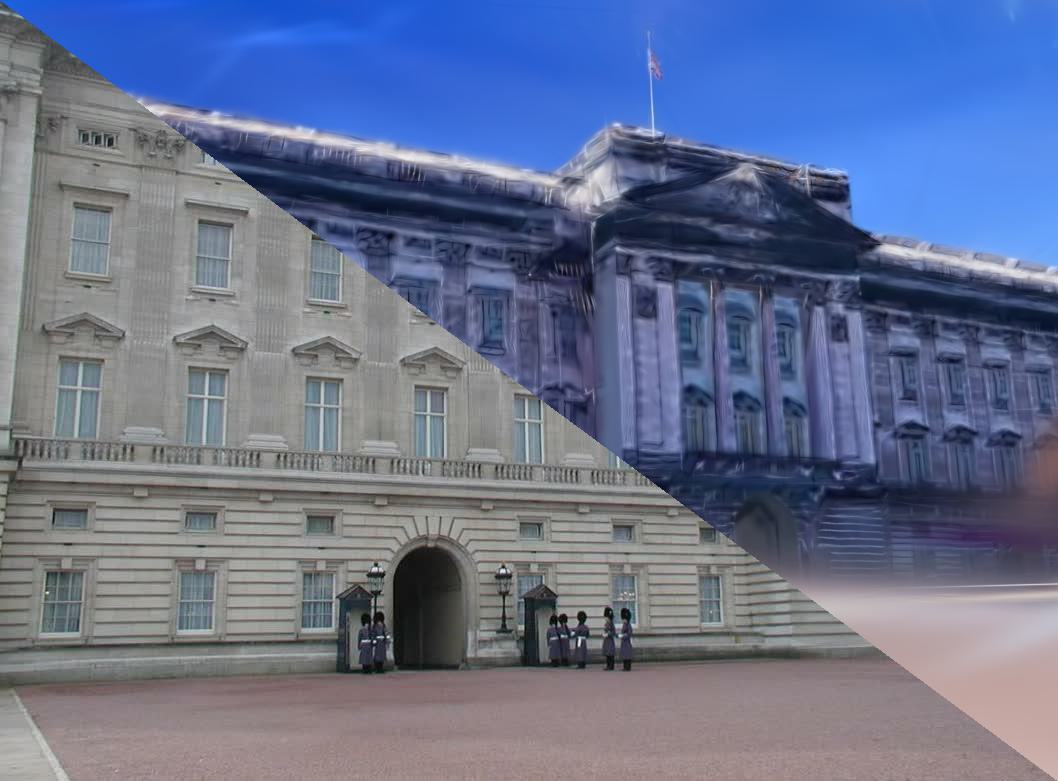}}}{}
    \jsubfig{\setlength{\fboxsep}{0pt}\fbox{\includegraphics[width=0.45\columnwidth]{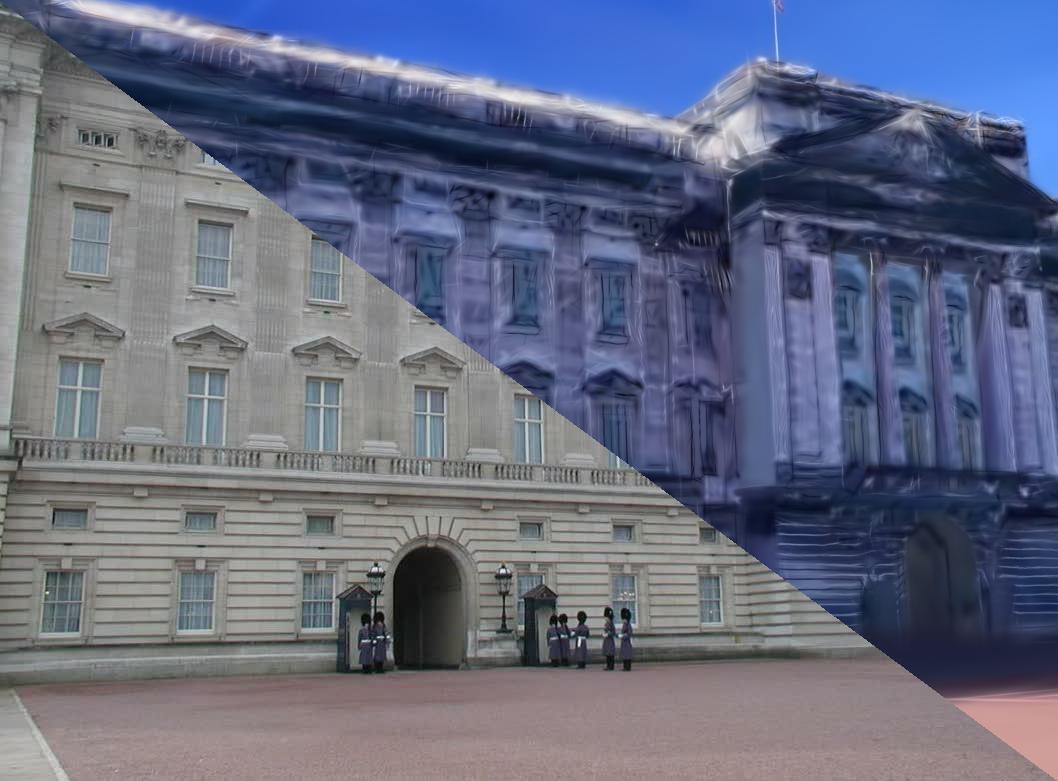}}}{} 
    \rotatebox{90}{\whitetxt{aaaaaa}Buckingham}
    \jsubfig{\setlength{\fboxsep}{0pt}\fbox{\includegraphics[width=0.45\columnwidth]{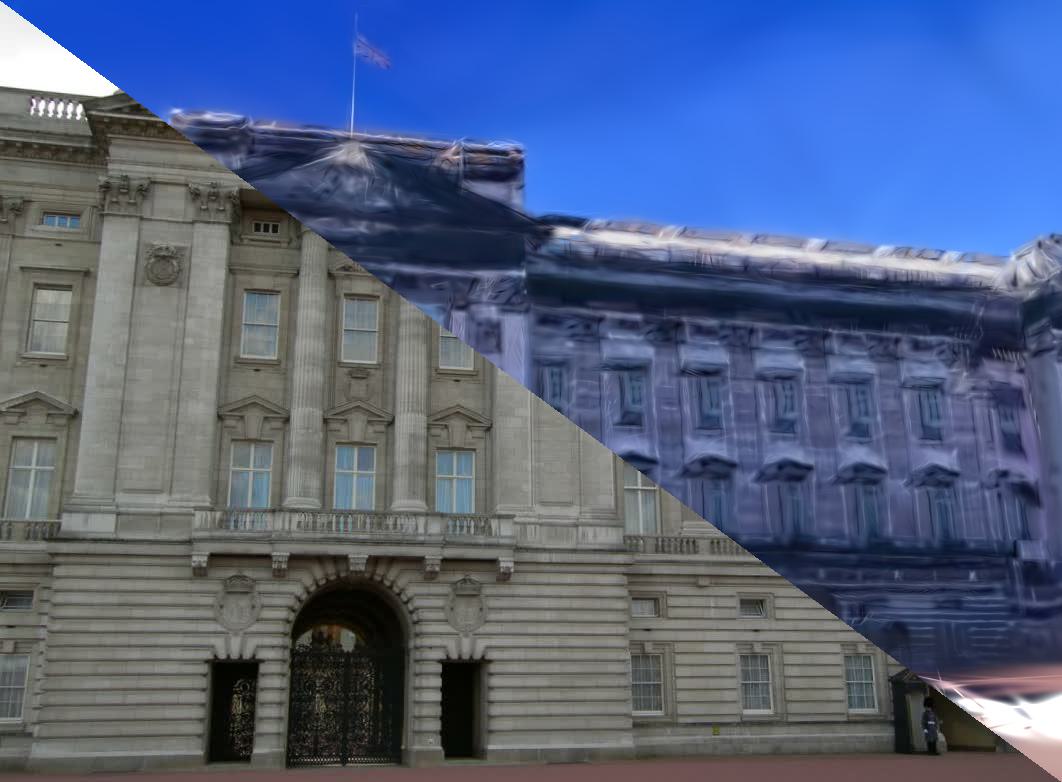}}}{Baseline}
    \jsubfig{\setlength{\fboxsep}{0pt}\fbox{\includegraphics[width=0.45\columnwidth]{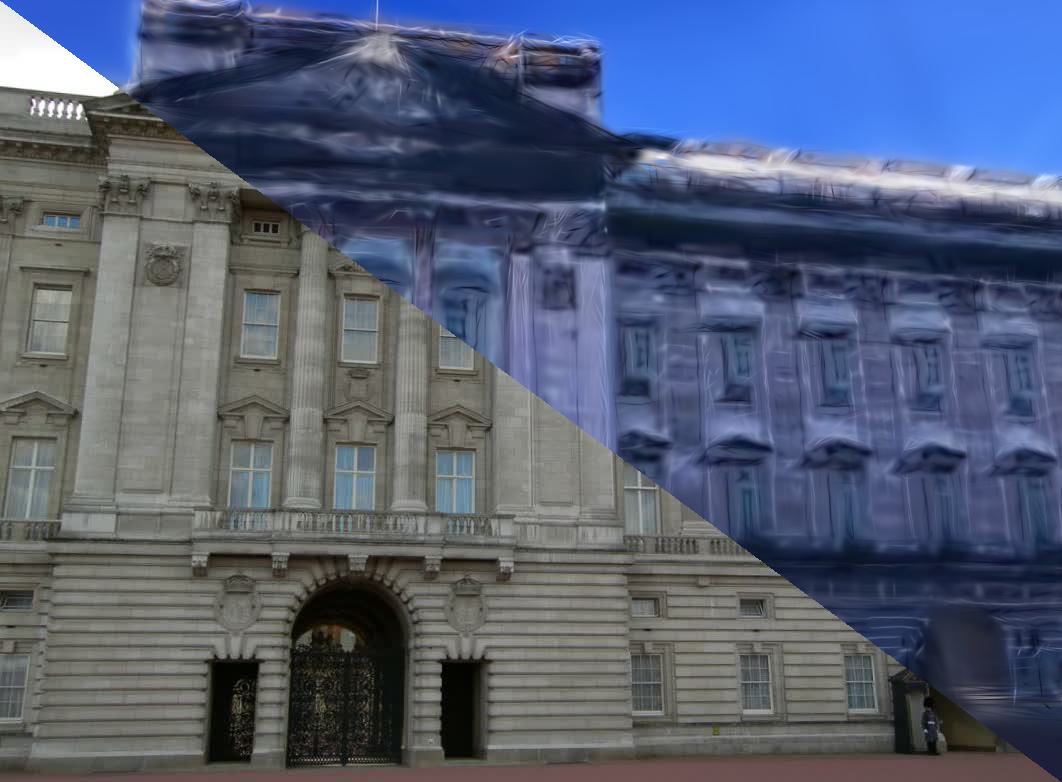}}}{Ours}

    \caption{\textbf{Qualitative Comparison on Different Scene Types}  A visualization of the alignment results for our method and the COLMAP baseline. The scenes are taken from the IMC-PT \cite{IMC-PT} dataset.}
    \label{imc_pt_comparison}
\end{figure}

In \cref{drone_comparison_supp} we show additional qualitative comparison comparison to the COLMAP baseline using a reference model created from drone videos.
In \cref{imc_pt_comparison} we show the results of our method on the IMC-PT dataset compared to the COLMAP baseline.
In \cref{baseline_comparison} we show additional qualitative comparison to the COLMAP baseline.

\begin{figure*}
    \centering

    \rotatebox{90}{Bordeaux Cathedral}
    \hspace{0pt}    
    \jsubfig{\setlength{\fboxsep}{0pt}\fbox{\includegraphics[width=2.5cm, trim=0 225 200 200, clip]{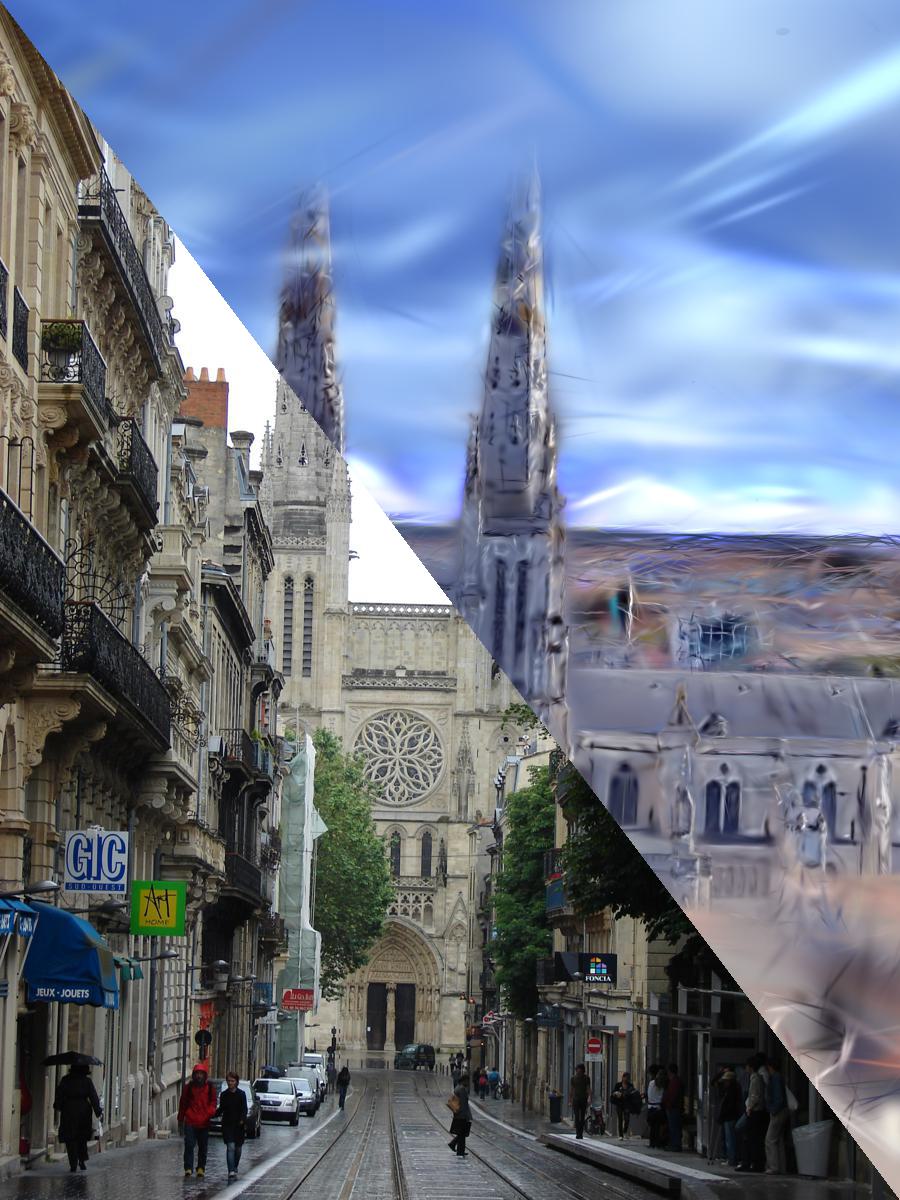}}}{}
    \jsubfig{\setlength{\fboxsep}{0pt}\fbox{\includegraphics[width=2.5cm, trim=0 225 200 200, clip]{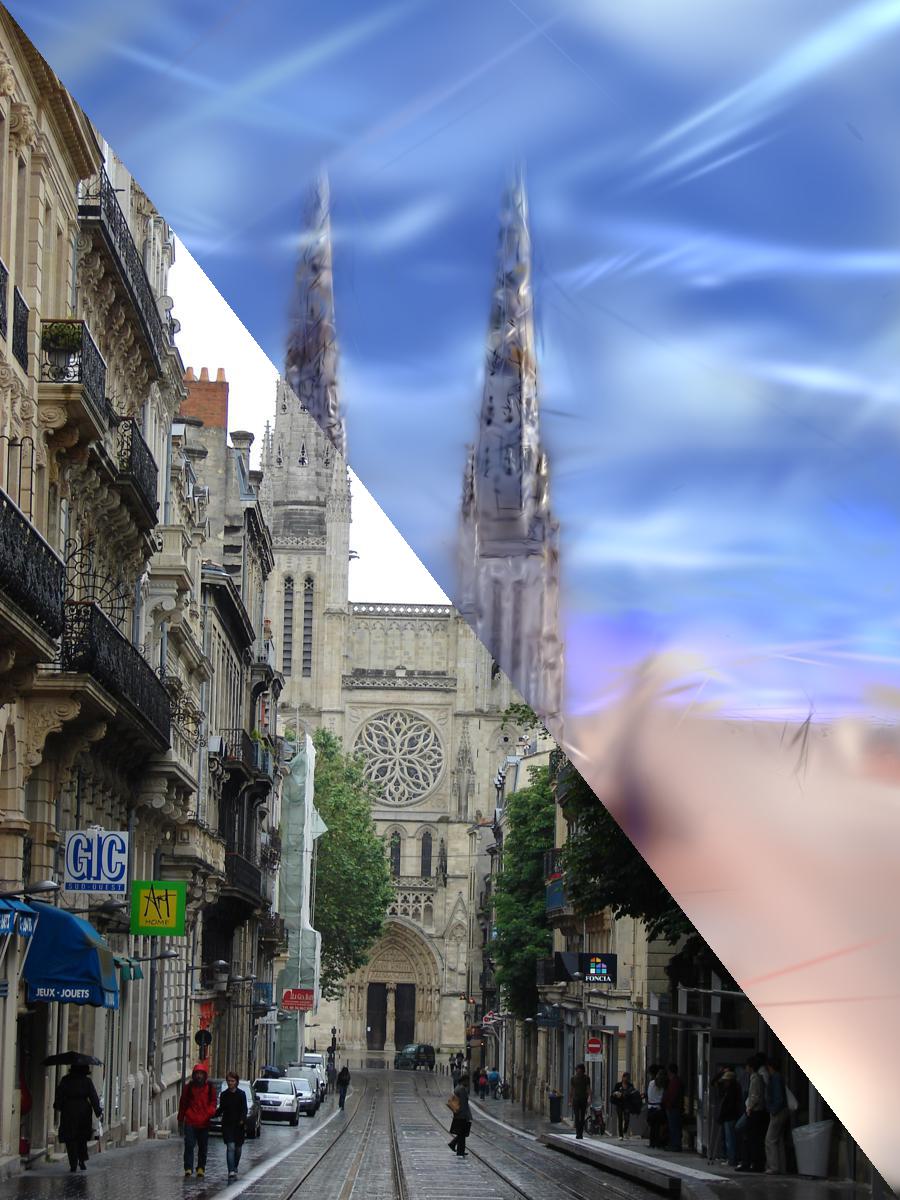}}}{} 
    \hspace{0.5cm}
    \jsubfig{\setlength{\fboxsep}{0pt}\fbox{\includegraphics[width=2.5cm, trim=200 0 200 0, clip]{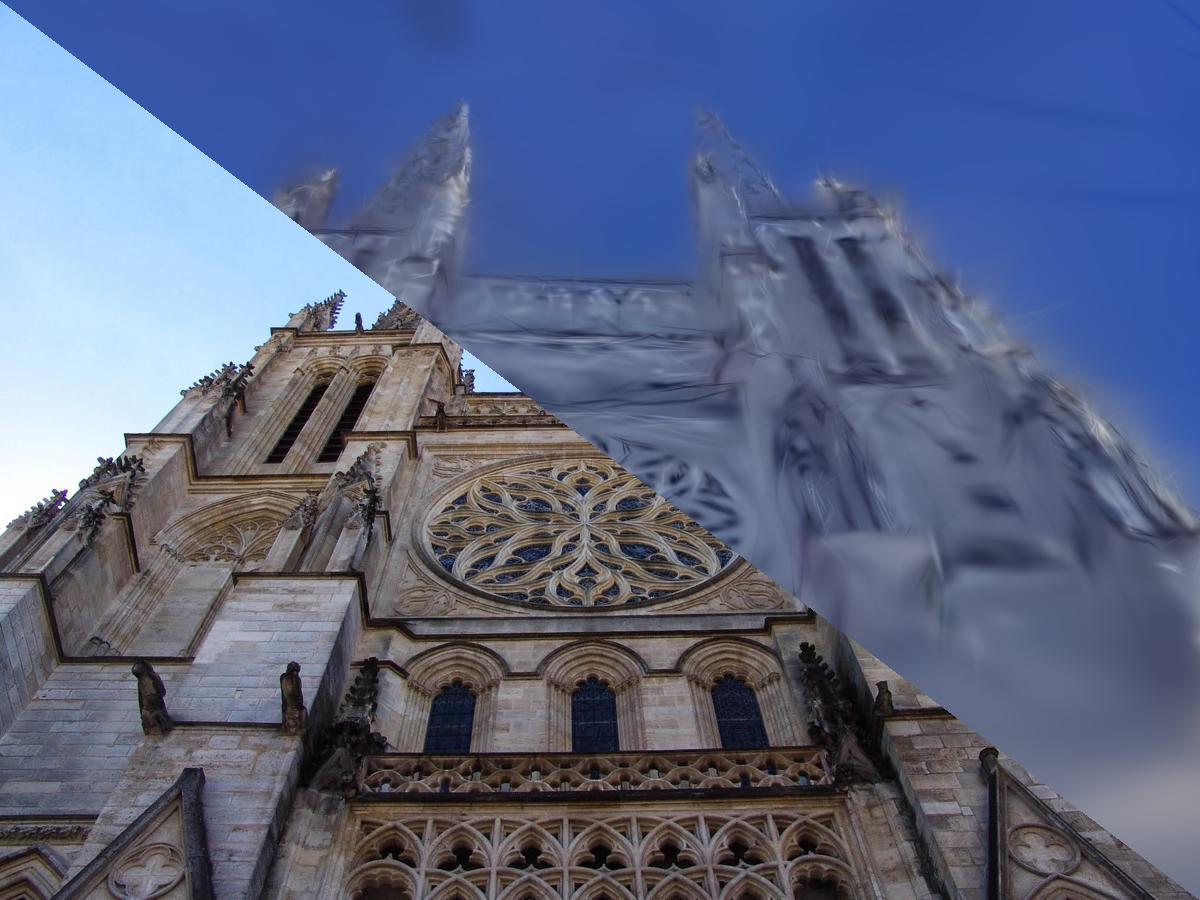}}}{}
    \jsubfig{\setlength{\fboxsep}{0pt}\fbox{\includegraphics[width=2.5cm, trim=200 0 200 0, clip]{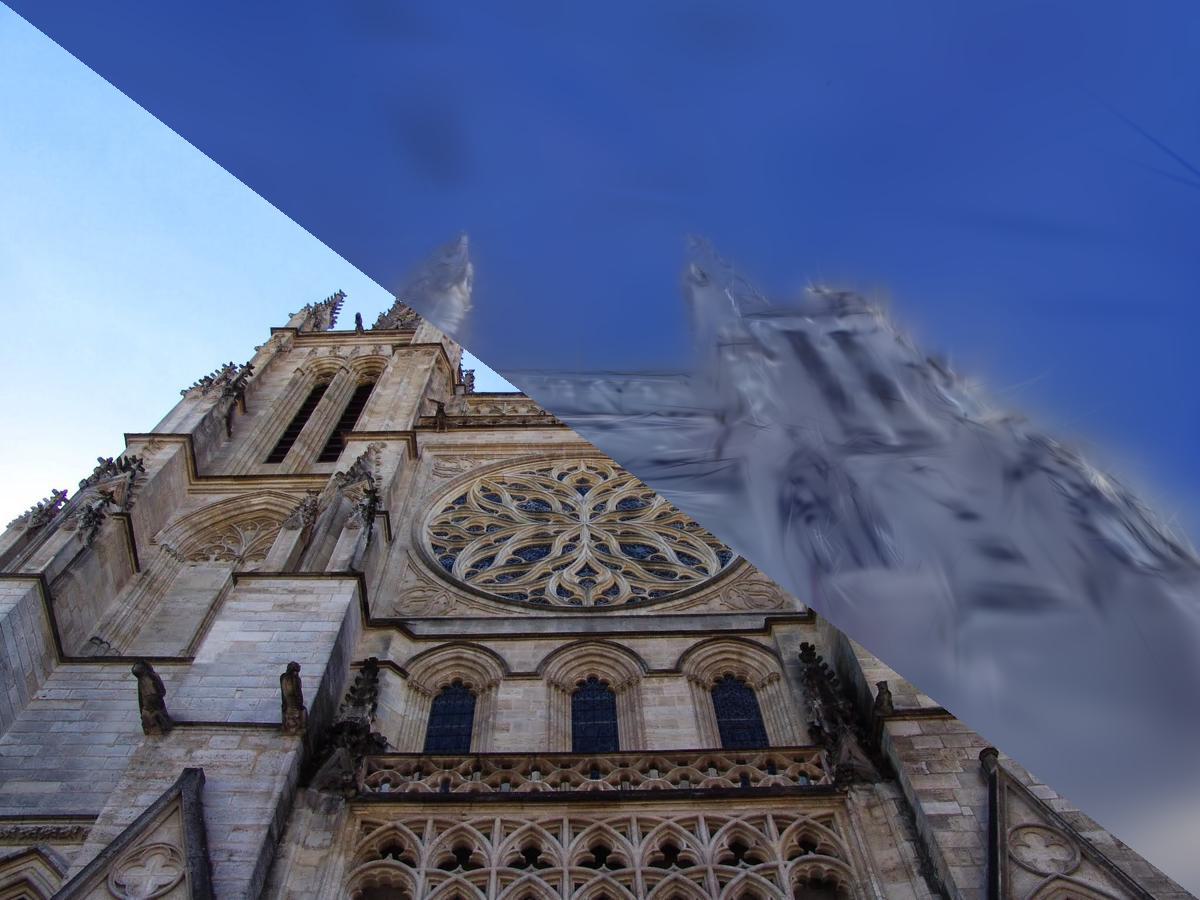}}}{}
    \hspace{0.5cm}
    \jsubfig{\setlength{\fboxsep}{0pt}\fbox{\includegraphics[width=2.5cm, trim=0 325 200 200, clip]{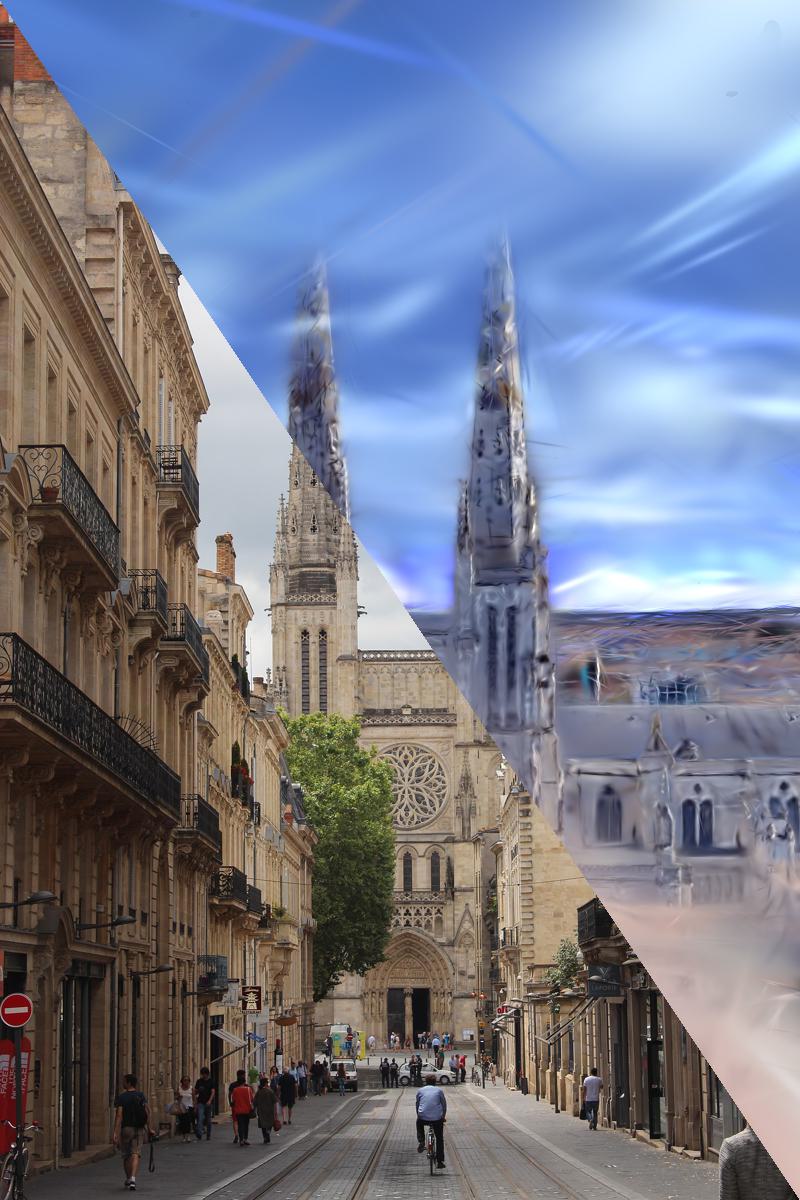}}}{}
    \jsubfig{\setlength{\fboxsep}{0pt}\fbox{\includegraphics[width=2.5cm, trim=0 325 200 200, clip]{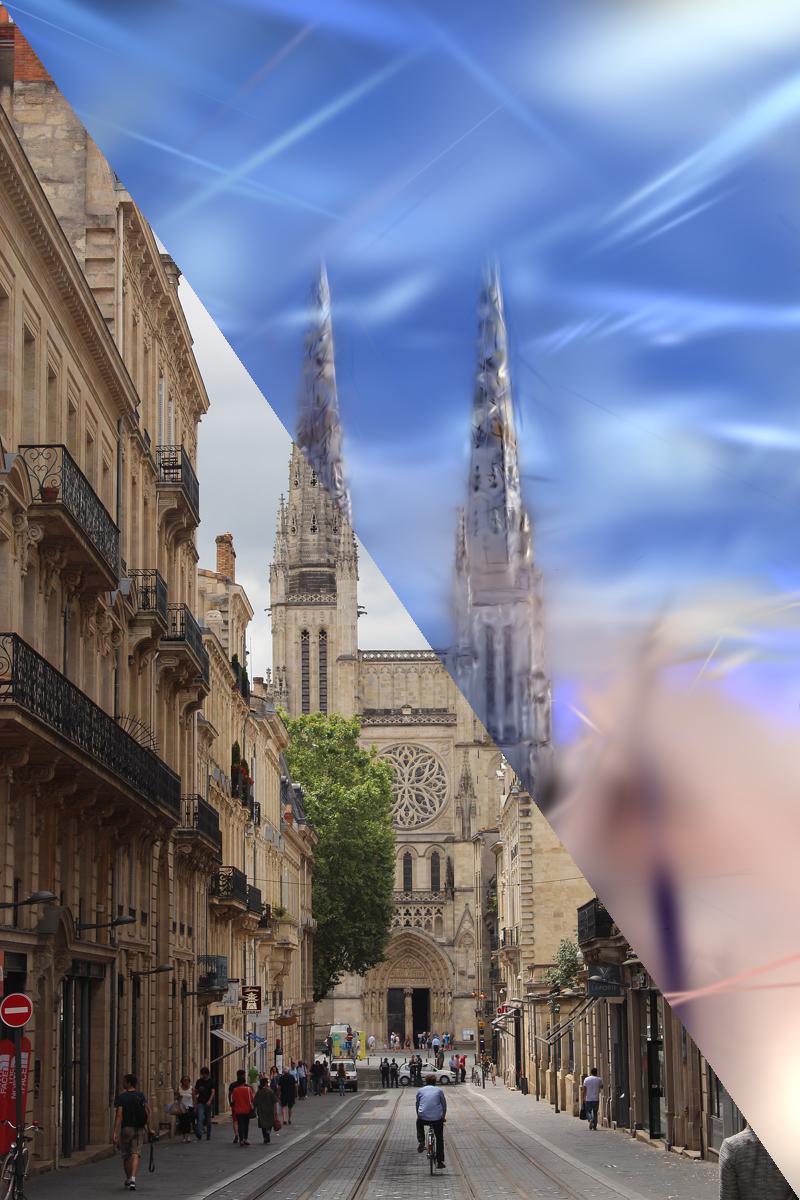}}}{}

    \vspace{0.2cm}

    \rotatebox{90}{Naples Cathedral}
    \jsubfig{\setlength{\fboxsep}{0pt}\fbox{\includegraphics[width=2.5cm, trim=300 0 200 0, clip]{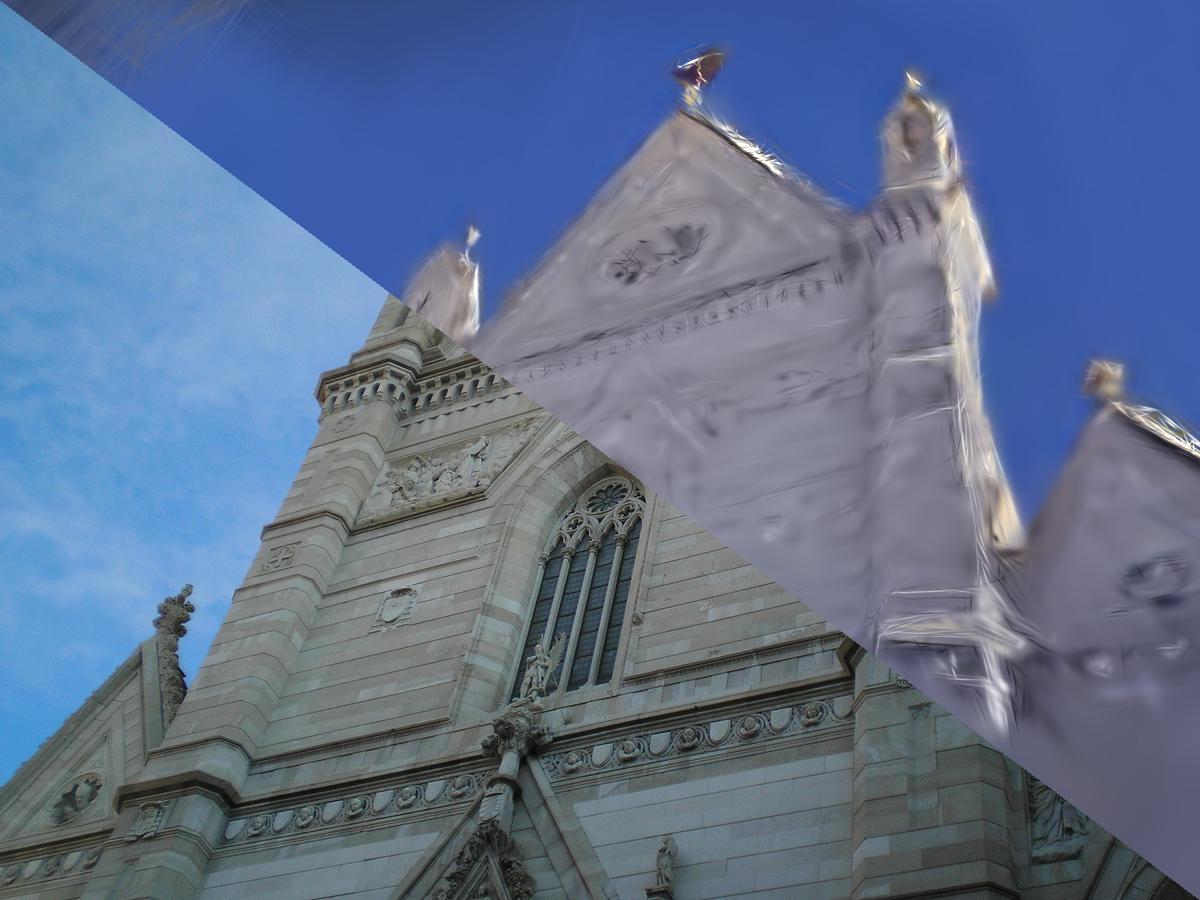}}}{}
    \jsubfig{\setlength{\fboxsep}{0pt}\fbox{\includegraphics[width=2.5cm, trim=300 0 200 0, clip]{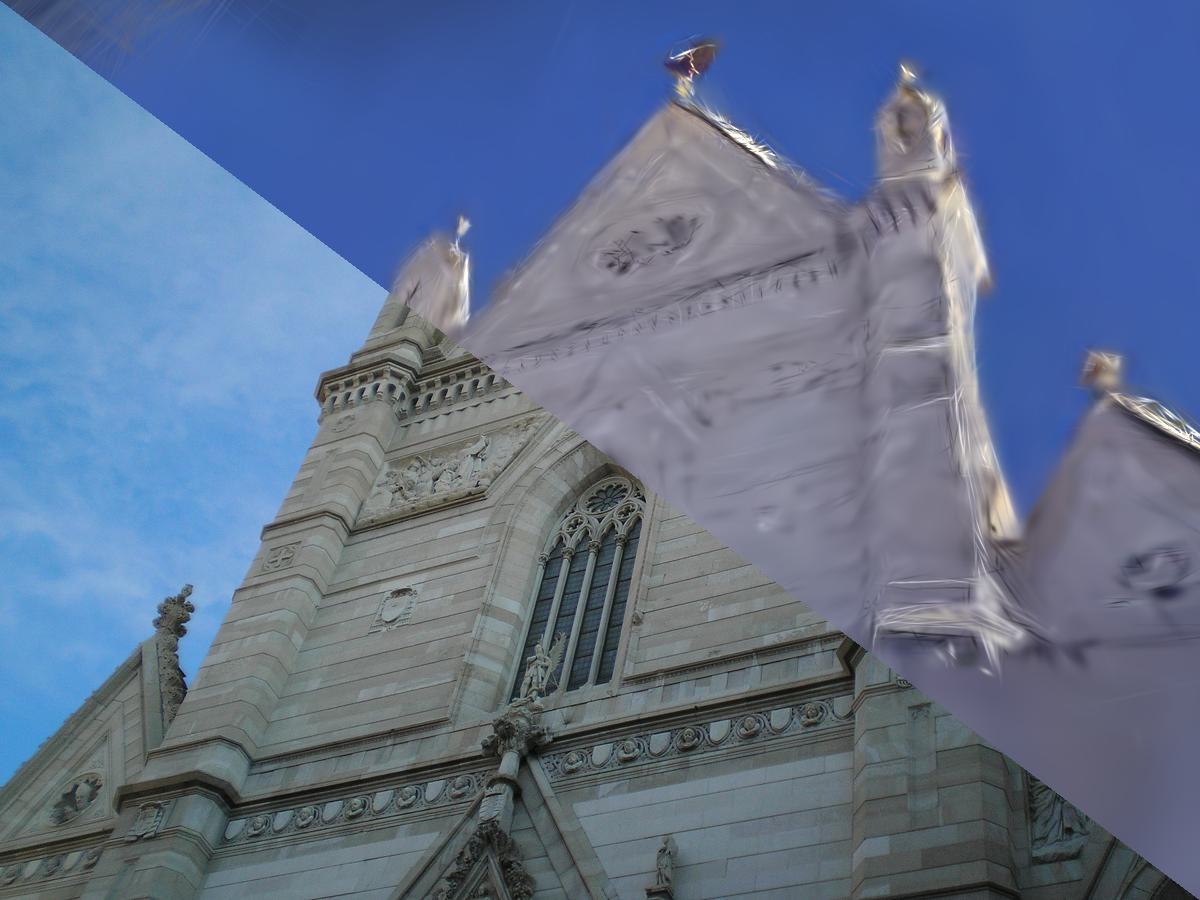}}}{}
    \hspace{0.5cm}
    \jsubfig{\setlength{\fboxsep}{0pt}\fbox{\includegraphics[width=2.5cm, trim=0 150 0 50, clip]{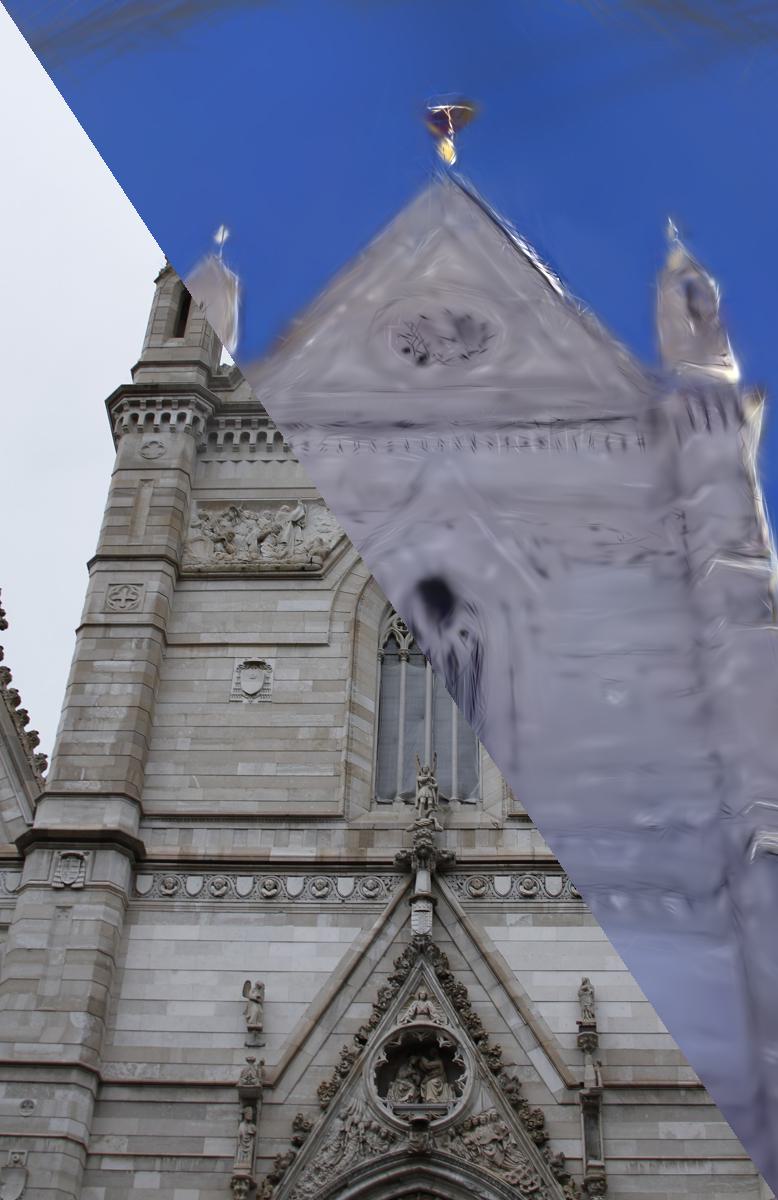}}}{}
    \jsubfig{\setlength{\fboxsep}{0pt}\fbox{\includegraphics[width=2.5cm, trim=0 150 0 50, clip]{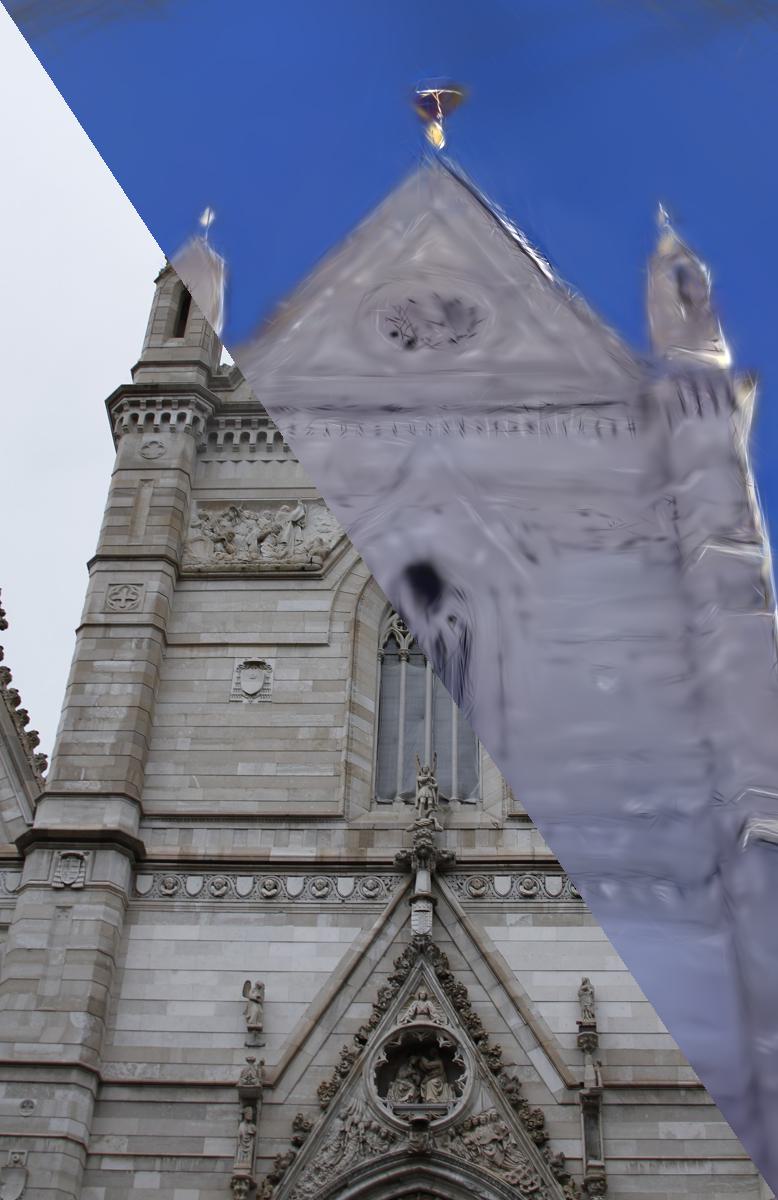}}}{}
    \hspace{0.5cm}
    \jsubfig{\setlength{\fboxsep}{0pt}\fbox{\includegraphics[width=2.5cm, trim=100 175 0 0, clip]{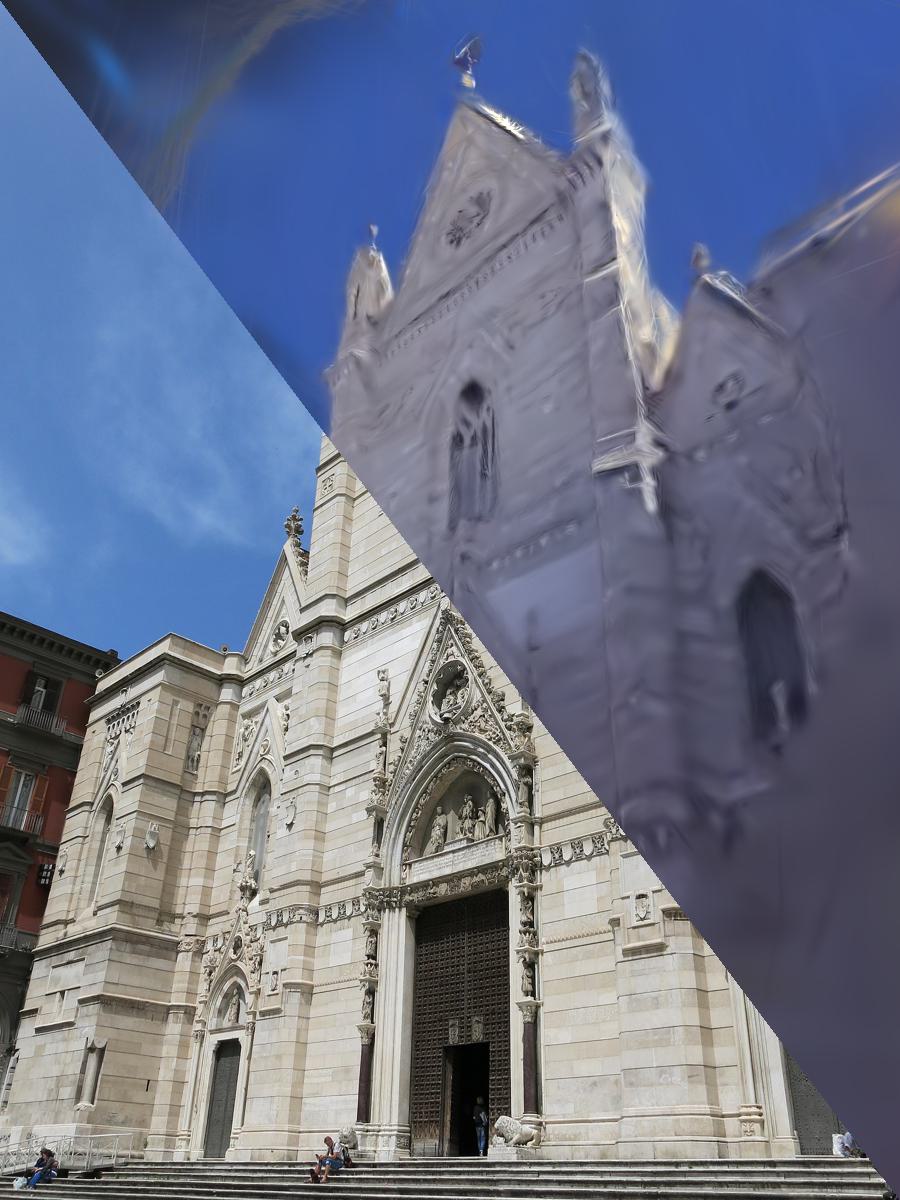}}}{}
    \jsubfig{\setlength{\fboxsep}{0pt}\fbox{\includegraphics[width=2.5cm, trim=100 175 0 0, clip]{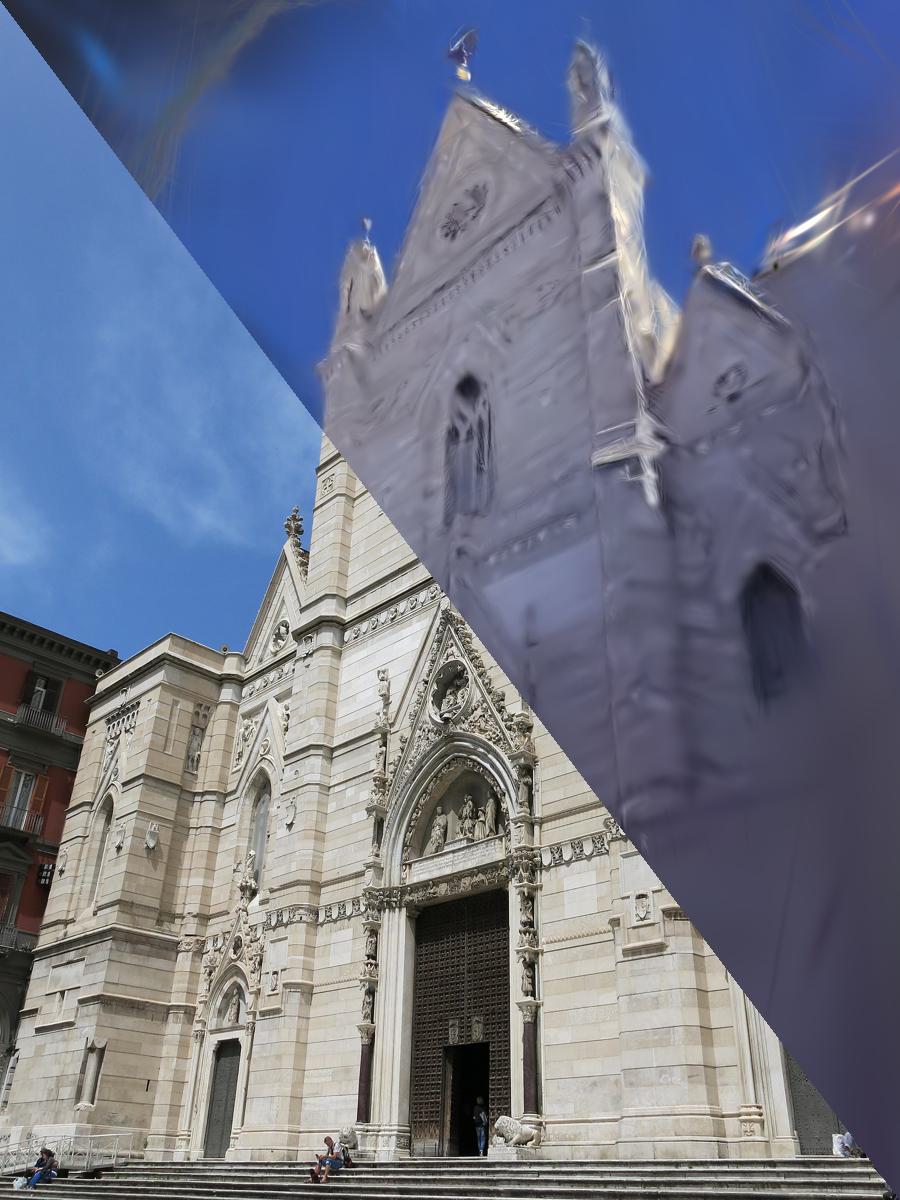}}}{}

    \vspace{0.2cm}
    
    \rotatebox{90}{Metz Cathedral}
    \hspace{0pt}    
    \jsubfig{\setlength{\fboxsep}{0pt}\fbox{\includegraphics[width=2.5cm, trim=0 50 0 0, clip]{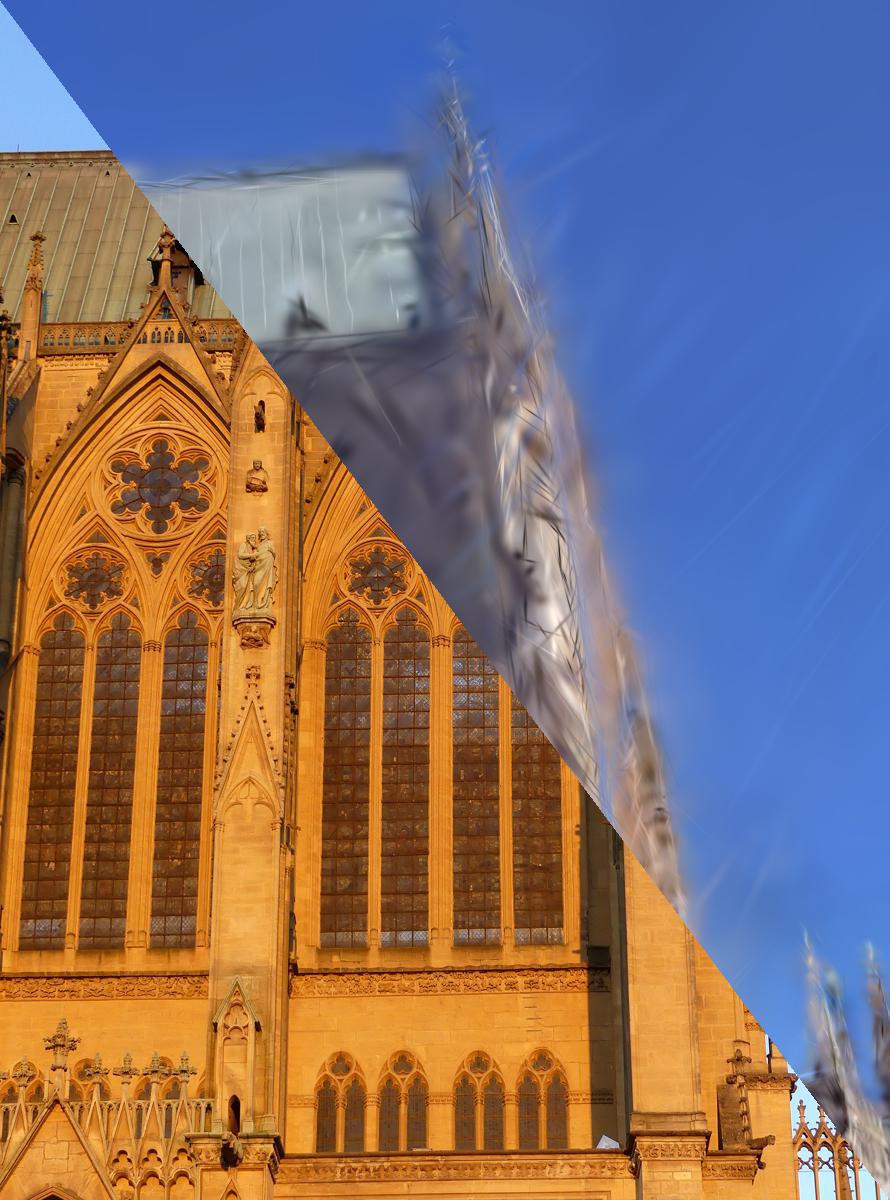}}}{}
    \jsubfig{\setlength{\fboxsep}{0pt}\fbox{\includegraphics[width=2.5cm, trim=0 50 0 0, clip]{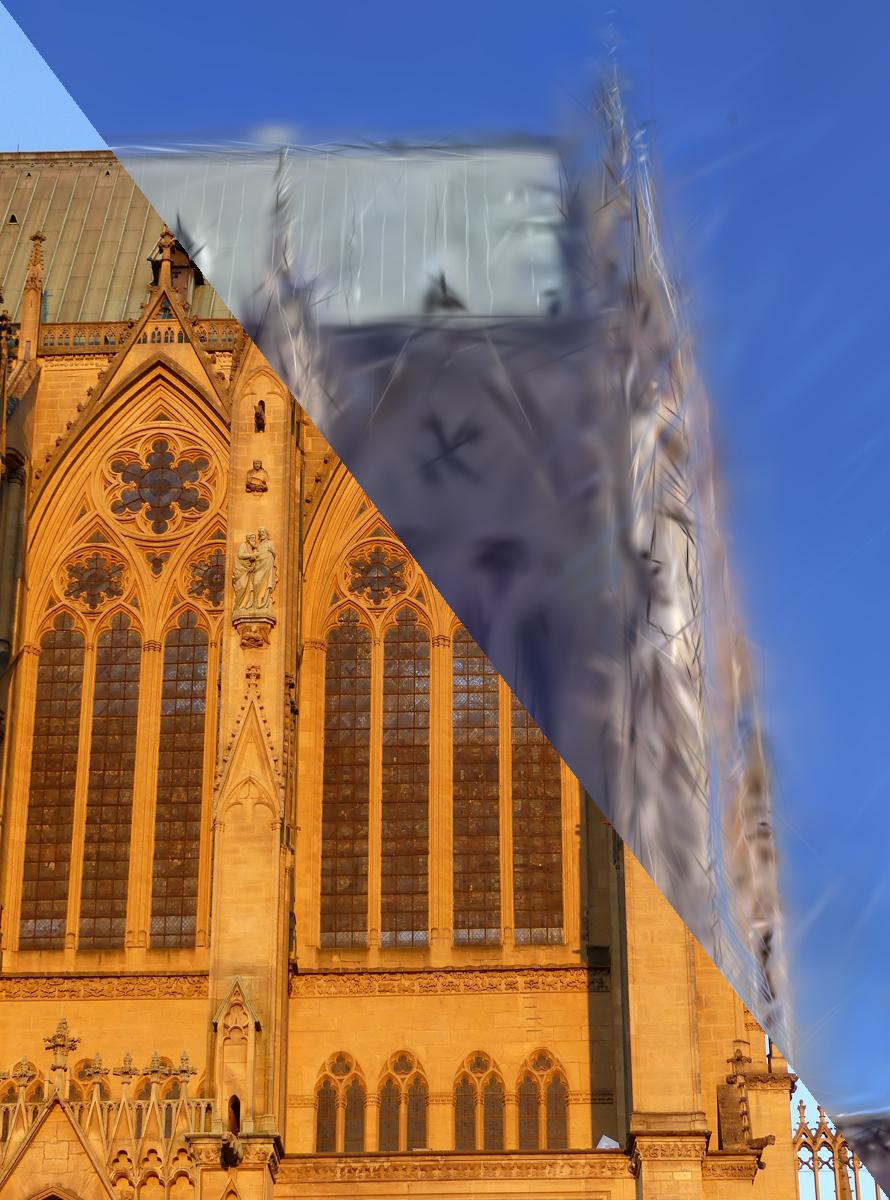}}}{}
    \hspace{0.5cm}
    \jsubfig{\setlength{\fboxsep}{0pt}\fbox{\includegraphics[width=2.5cm, trim=0 25 0 0, clip]{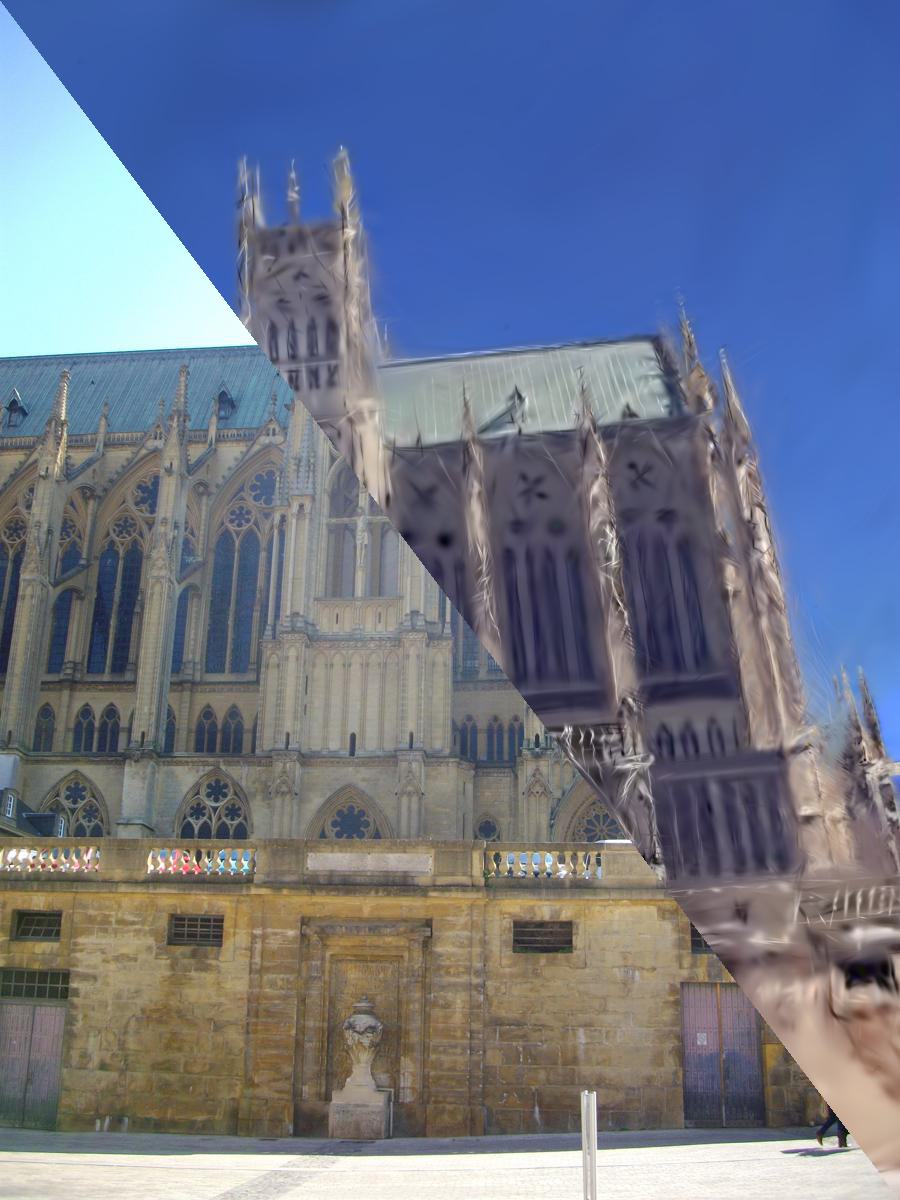}}}{}
    \jsubfig{\setlength{\fboxsep}{0pt}\fbox{\includegraphics[width=2.5cm, trim=0 25 0 0, clip]{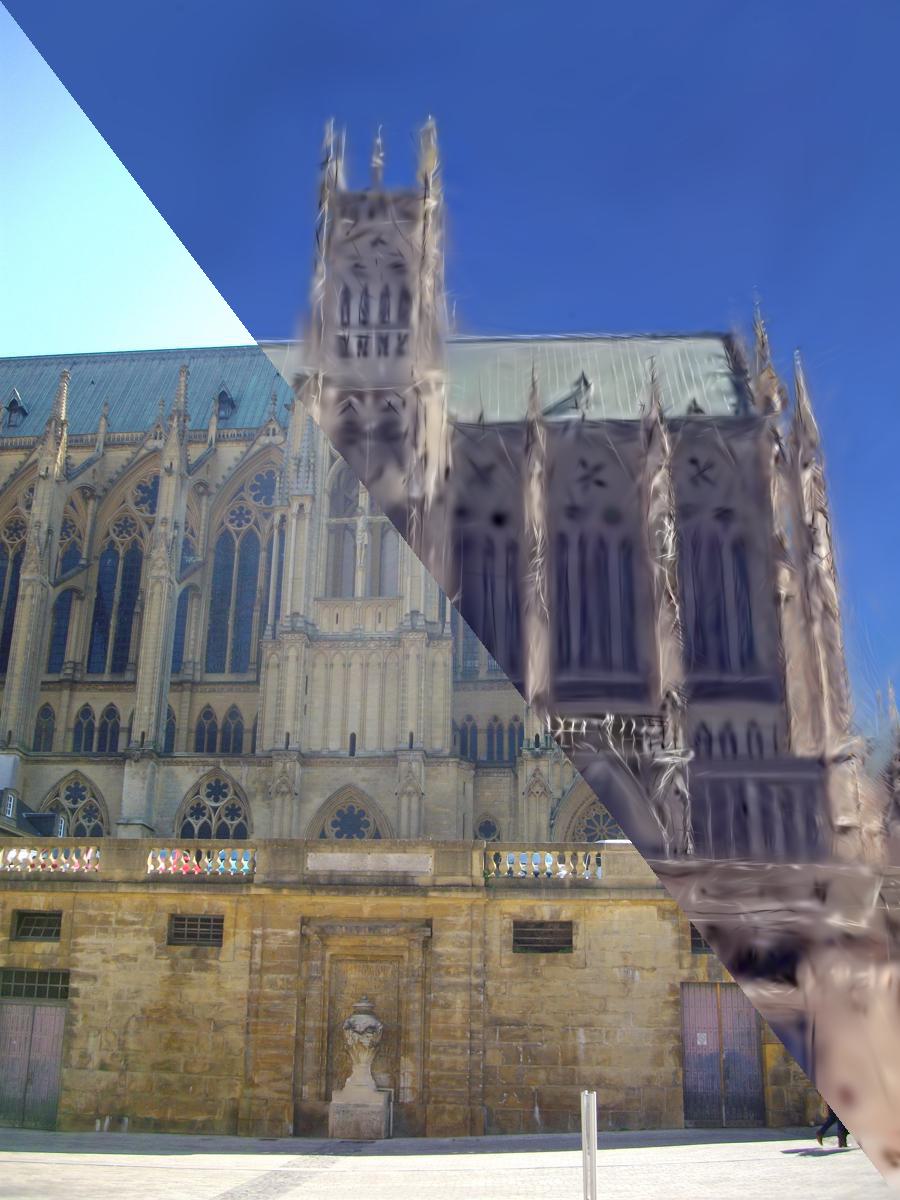}}}{}
    \hspace{0.5cm}
    \jsubfig{\setlength{\fboxsep}{0pt}\fbox{\includegraphics[width=2.5cm, trim=50 25 0 0, clip]{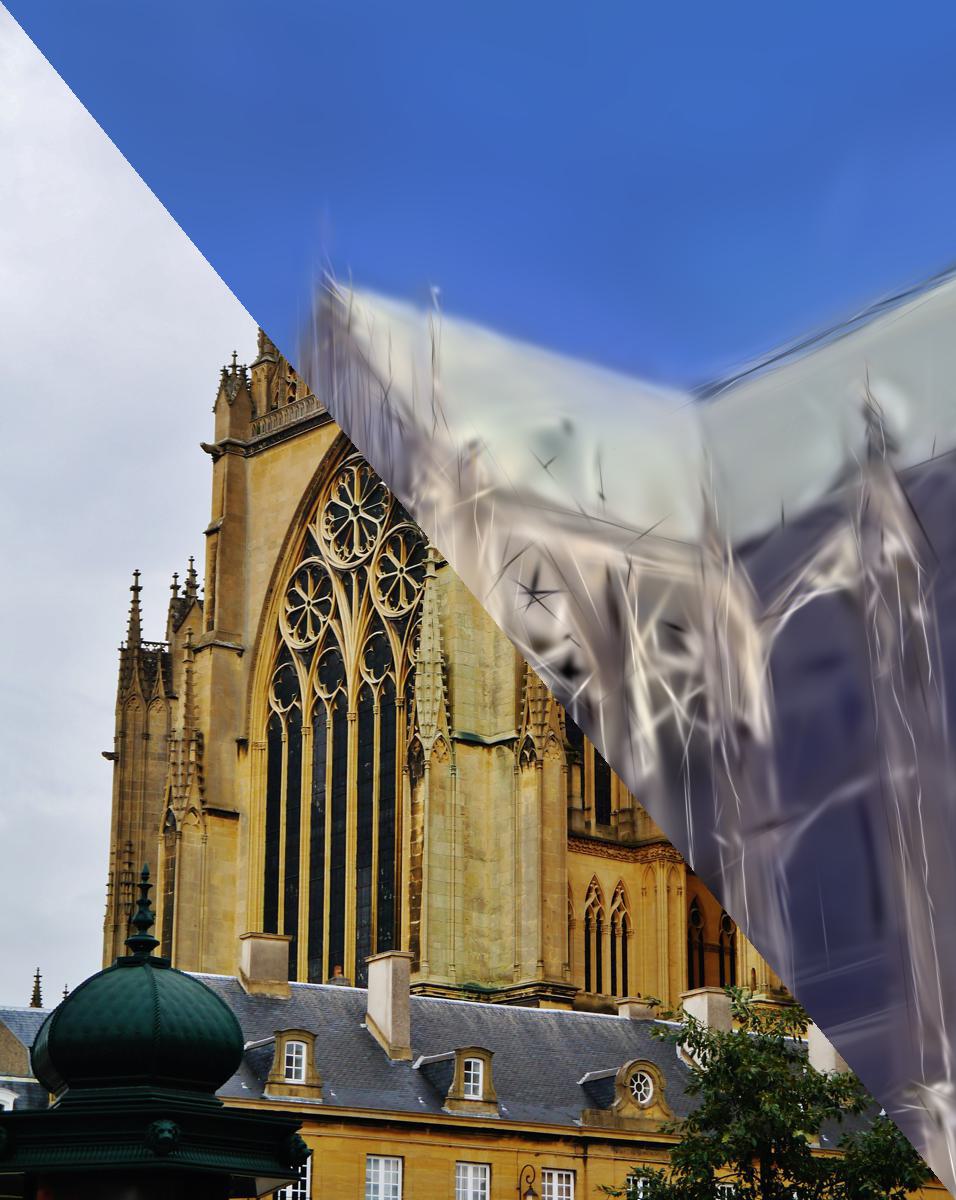}}}{}
    \jsubfig{\setlength{\fboxsep}{0pt}\fbox{\includegraphics[width=2.5cm, trim=50 25 0 0, clip]{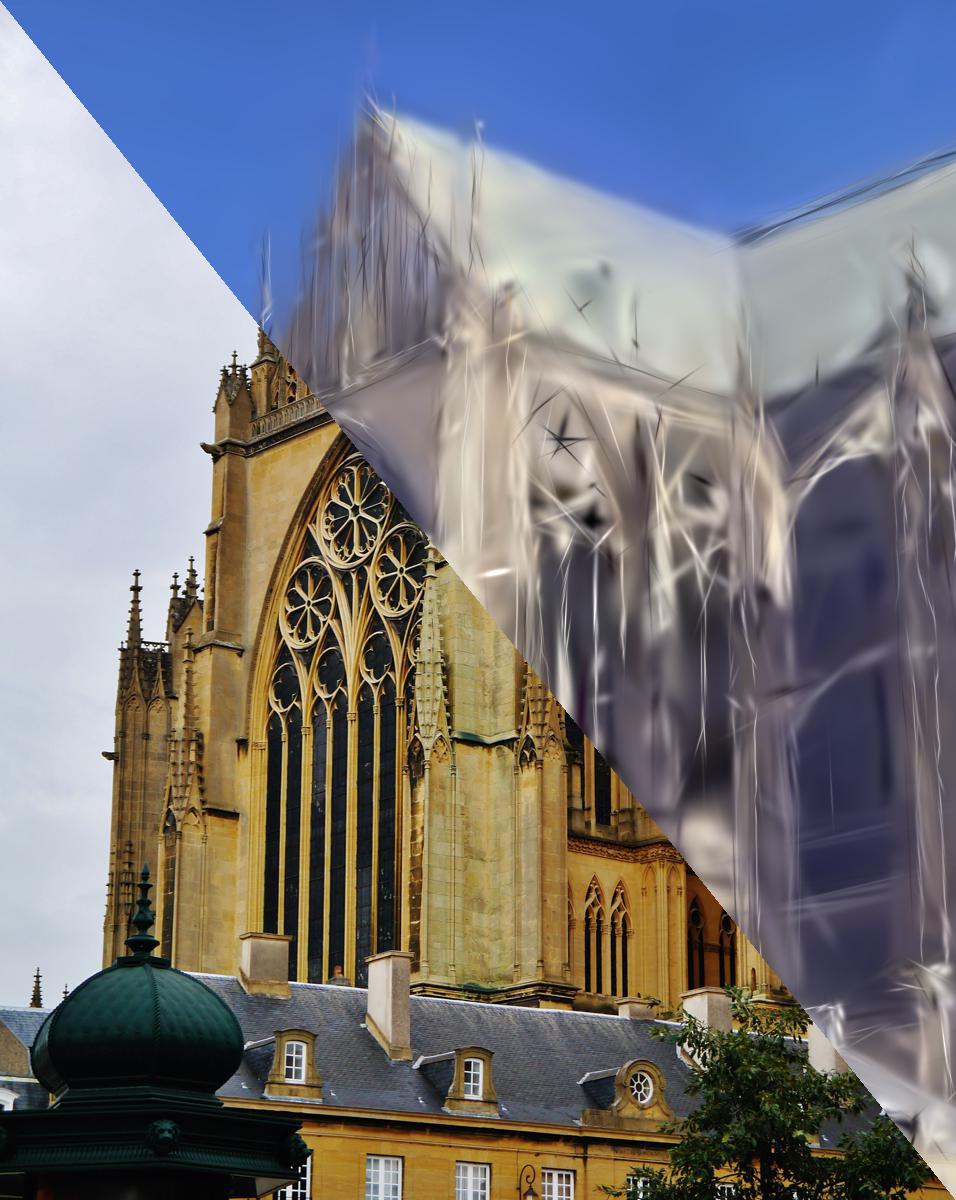}}}{}

    \caption{\textbf{Additional Qualitative Comparison:} A visualization of the alignment results for our method and the COLMAP baseline. Each image shows the ground truth in the lower half and the rendered image from the reference model $\mathcal{M}$ after alignment in the top half. As demonstrated, our inverse optimization-based approach predicts precise transformations, even in the presence of challenging, inaccurate initializations.}
    \label{baseline_comparison}
\end{figure*}

\subsection{Method Analysis}
\label{sec:method_analysis}
\begin{figure*}
    \jsubfig{\includegraphics[width=0.5\linewidth, trim=40 0 60 0, clip]{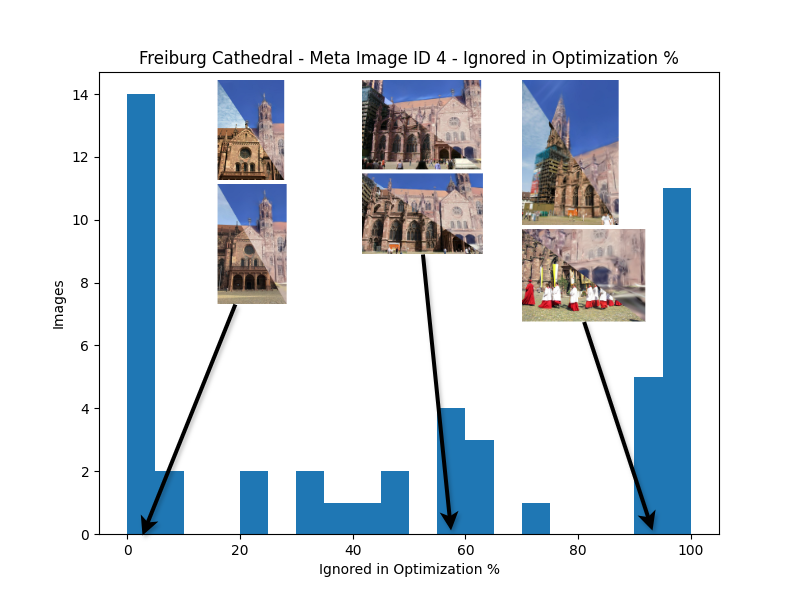}}{}
    \jsubfig{\includegraphics[width=0.5\linewidth, trim=40 0 60 0, clip]{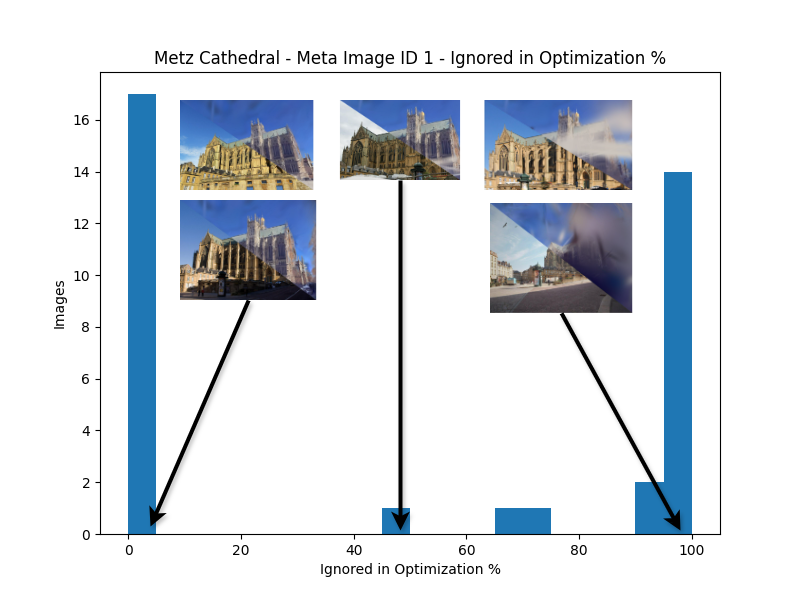}}{}
    \jsubfig{\includegraphics[width=0.5\linewidth, trim=40 0 60 0, clip]{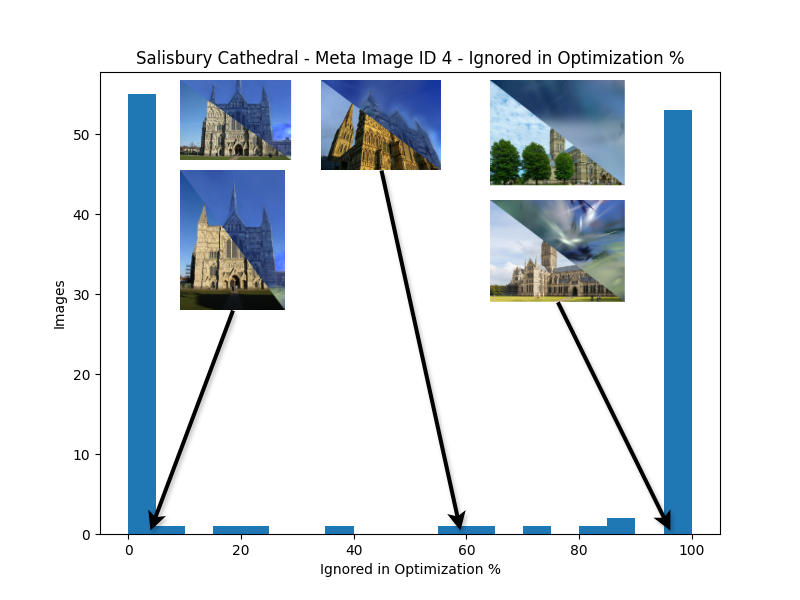}}{}
    \jsubfig{\includegraphics[width=0.5\linewidth, trim=40 0 60 0, clip]{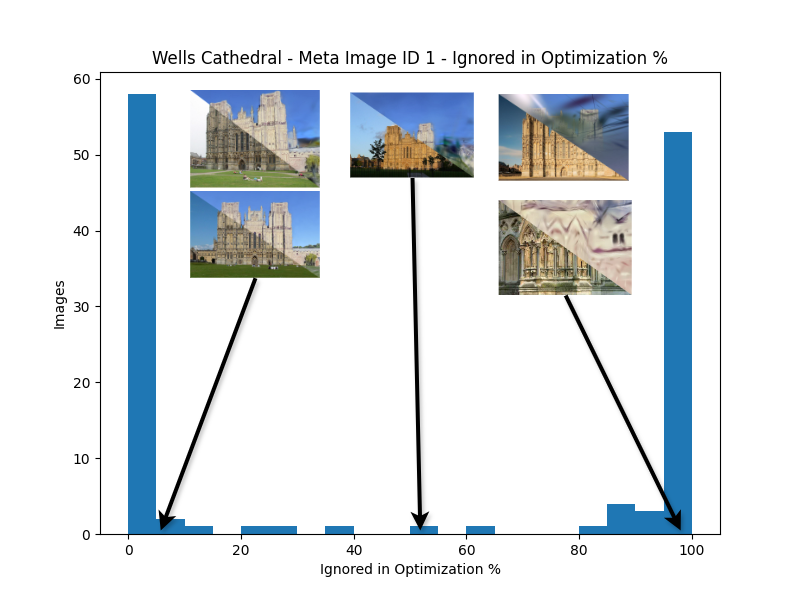}}{}
    \caption{\textbf{Additional Analysis of our Robust Optimization Framework}. Our method uses LTS to ignore the images with loss values that are higher than the median image loss. In the figure, we show a sample of images in several bins. The image below the diagonal are the real-world Internet images, and above is the rendered image from the reference model (rendered at the end of the optimization). As illustrated by the rightmost bin, images there are typically outliers, \emph{i.e.}, images with occlusions and images that resides behind floaters in the reference model. Further analysis is provided in \cref{sec:method_analysis}.
    }
    \label{fig:ignored_images_histogram}
    \end{figure*}

As mentioned in the main paper, to handle image outliers we use robust optimization method (LTS \cite{least_trimmed_squares}). In each optimization iteration the LTS ignores images with loss higher than the median. We visualize the ignored images distribution in \cref{fig:ignored_images_histogram}. As discussed in Section 5.4 of the main paper, the histograms reveal that most images are either constantly ignored (in the rightmost bin) or never ignored (in the leftmost bin). However, some image fall into the middle bins where they are sometimes ignored and sometimes not. The histograms further illustrate that our robust optimization scheme provides a soft selection mechanism enabling stable convergence. 

Additionally, we visualize images from several bins. These demonstrate that the images that are always ignored by our method are indeed outliers. Specifically, these are images with occlusion and images that resides behind floaters in the reference model.

\subsection{Comparison with Feed-Forward 3D Models}
\label{sec:feed_forward_comparisons}
In the paper, we compare our method with three baselines: VGGT, MASt3R, $\pi^3$ and DUSt3R. We ran MASt3R and DUSt3R using the master-sfm code provided in their official repository, we ran $\pi^3$ from the official repository and we ran VGGT using the demo-colmap setup from its official repository. Due to GPU memory limitations on our A5000 GPU, we were unable to run these baseline (except $\pi^3$) methods with more than 45 input images. We were able to run $\pi^3$ with up to 180 images.

For the meta-to-meta experiment, we sampled 22 random images from each meta image and ran the experiment 5 times.

For the meta-to-reference experiment, we sampled 35 images from the reference model and 10 random images from the meta model. We chose to sample 35 images from the reference model to ensure sufficient coverage of the entire scene. We did not sample random images from the reference model because random sampling often failed to cover the scene, causing the Feed-Forward methods to fail. Instead, we selected evenly spaced images, since the images were captured sequentially. We repeated this experiment 5 times, each with a different image offset.

\section{Limitations}
\label{sec:limitations_supp}
\begin{figure}
    \jsubfig{{\includegraphics[width=1\linewidth] {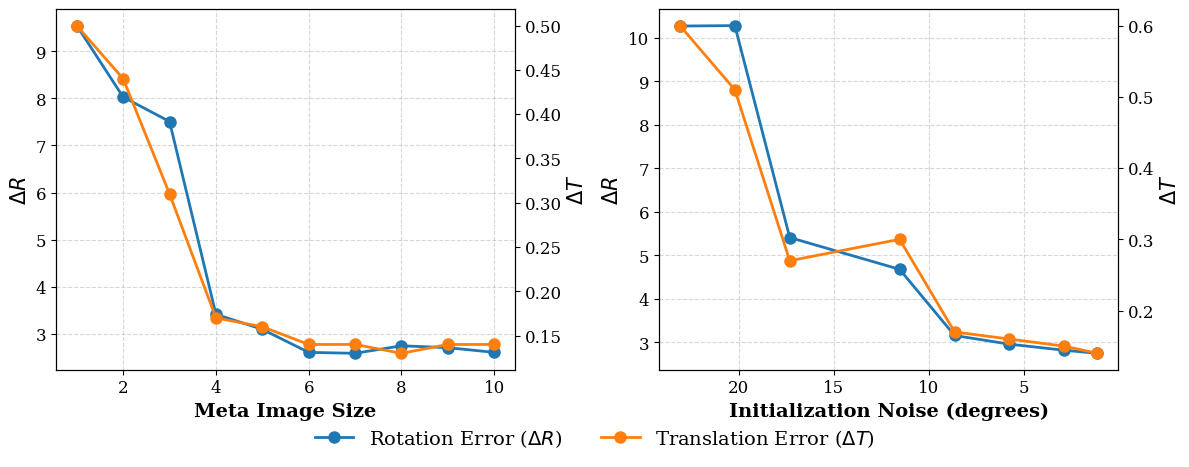}}}{}
    \vspace{-20pt}
    \caption{\textbf{Robustness Analysis}. Average errors $\Delta R$ and $\Delta T$ across the benchmark as a function of the meta-image size and the initialization noise (rotation). The graphs indicate that the error increases with smaller meta-images, reaching a plateau at a size of approximately $6$. Furthermore, the initialization graph demonstrate that the method aligns the images successfully once the noise is below a specific threshold ($10^\circ$) for each of the rotation parameters; see Section \ref{sec:limitations_supp} for additional details.
    }
    \label{limitation_analysis}
\end{figure}

While our method is not specifically designed for single-shot scenarios, we evaluate its reliability with fewer images per meta-image in \cref{limitation_analysis} (left). We evaluate performance by randomly sub-sampling subsets of varying sizes from each meta-image, reporting the average error across five independent samples Performance drops over very small meta-images due to an insufficient number of informative images to guide our alignment scheme. However, results remain largely stable for small collections containing at least six images. Furthermore, our method is challenged by very noisy initializations. As illustrated in \cref{limitation_analysis} (right), adding noise to the rotation parameters leads to a significant drop in performance.

\section{Implementation Details}
\label{sup:implementation}

\subsection{The reference model}
First we extract DINOv2 \cite{oquab2024dinov2learningrobustvisual} dense features per rendered landmark image from Google Earth Studio. We resize each image to $1400X1400$ and then use the pretrained backbone \textit{dinov2\_vits14}, which outputs dense feature map $100X100$. We chose DINOv2 with embedding size of 384.
We use the DINO implementation \textit{facebookresearch/dinov2} in Github.

We use the landmark images rendered from Google Earth, the COLMAP model of those images from the benchmark, and the extracted DINOv2 feature to train 3DGS. We follow the implementation of Feature 3DGS\cite{zhou2024feature}.
We chose feature vector per Gaussian with size 128. We apply the Speedup Model, a Conv2d network with input 128 output 384 and kernel size 1 to decode the features of the Gaussian to the DINOv2 Features.
We integrated the implementation of Feature3DGS to Nerf Studio, specifically to Splatfacto - the Gaussian Splatting implementation in Nerf Studio.
In each training iteration we choose one image, render its feature (size 128) pass it through the speedup module which translated it to size 384. Then we use bilinear interpolation to resize the rendered feature image to 100X100 to match the DINOv2 features which was extracted beforehand. Similarly to \cite{zhou2024feature}, our Loss function is: 

\begin{equation}
L = L_{rgb} + \left| F_t(I) - F_s(\hat{I}) \right| 
\label{eq:loss}
\end{equation}
Where $L_{rgb}$ is the regular 3DGS, $I$ is the Image, $F_t(I)$ is the extracted features of the image in the pre-process, and $F_s(\hat{I})$ is the rendered images after the a pass through the speed up module and the bilinear interpolation.

For the 3DGS parameters we use the parameters of the Spaltfacto method in NerfStudio \cite{nerfstudio}. We use Adam Optimizer for the gaussian feature vectors with l2=0.05, eps=1e-15, and exponential decay scheduler with lr-final=1.6e-6. For the Feature Speedup Module we use Adam optimizer with lr=0.001, eps=1e-15. We train the model for 12000 steps.

\subsection{Initialization}
\label{sec:supp_initalization}
We use COLMAP, SP+LG and gDLS+++ to initialize the global transformation of meta-image $\cal I$. To find the global transformation using COLMAP we first register the images in the meta-image one by one. We give COLMAP as an input the landmark model we previously built from Google Earth Studio images. We fix the input model (using the flag fix-existing-images)
when running the COLMAP exhaustive matcher and the mapper. To find the global best transform we align the meta-image to the registered images using COLMAP model aligner. COLMAP model aligner uses point set registration and RANSAC. For the COLMAP aligner we used alignment-max-error=3, alignment-type="custom" and ref-is-gps=0. For the SP+LG initialization we perform the same steps as the COLMAP initialization, but replacing the feature extractor from SIFT to SuperPoint and the feature matching to Light Glue.

\subsection{Registration Implementation Details}
For the registration vector we used Adam Optimizer with lr=1e-3, eps=1e-8. 

We noticed that in some 3DGS models there were floaters around the ground, which impede the convergence of our registration in some cases. To mitigate it we set the near plane of all the cameras 0.7, so the cameras will not render the gaussians near them, especially the floaters on the ground.

\subsection{Runtime} 

 The optimization (creating the reference model, performed once per reference model) takes roughly 15 minutes on a Nvidia A5000 GPU. The registration (of the meta image to the reference model) is performed over 12000 steps, taking about 5 minutes on a Nvidia A5000 GPU.

\begin{figure}
\includegraphics[width=0.4\textwidth]{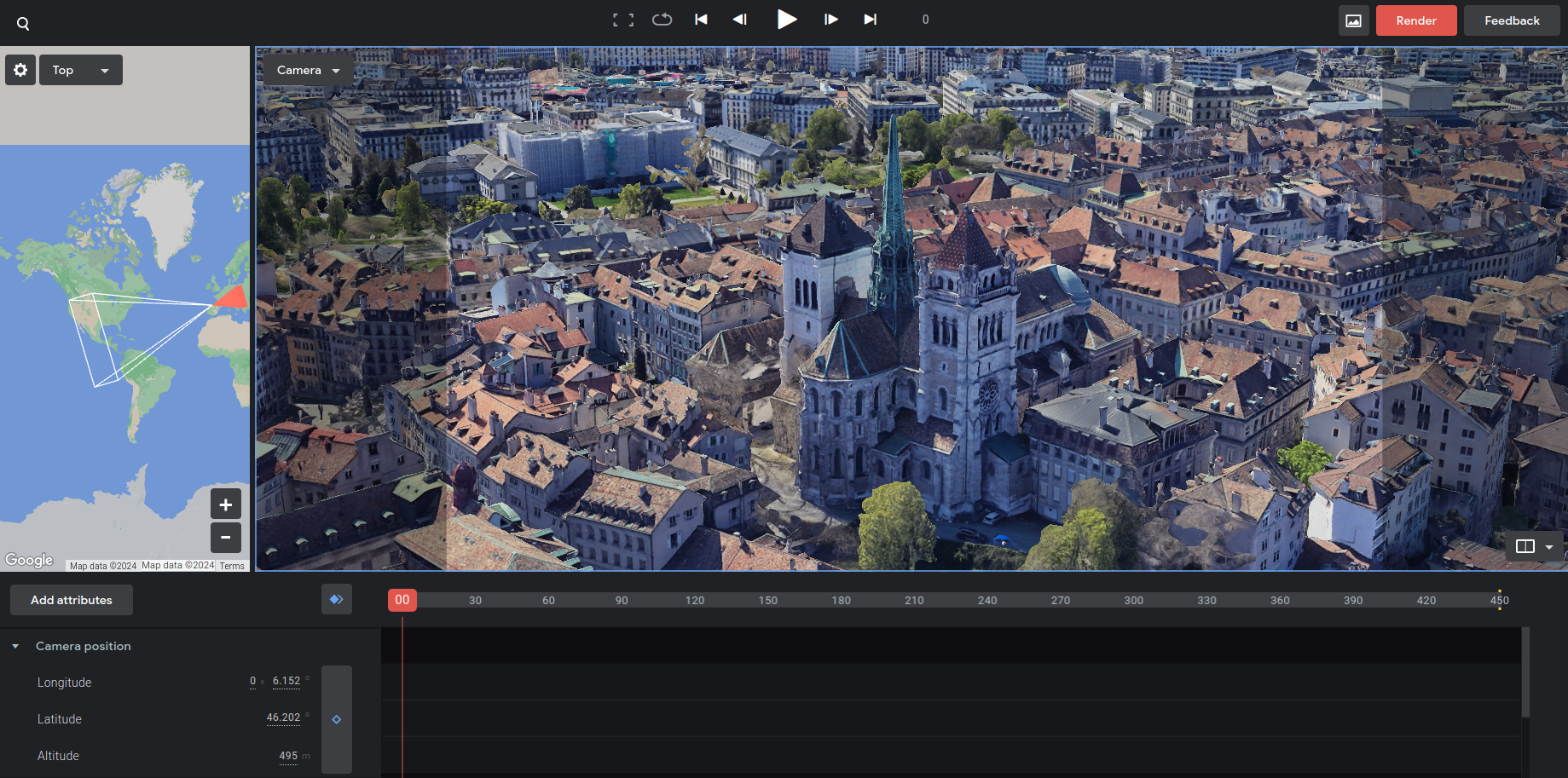}
\caption{\textbf{Google Earth Studio UI:} Screenshot of Google Earth Studio, showing the 3D model of the Geneva Cathedral. For each landmark we create a camera trajectory and rendered the images on the trajectory using the program }
\label{fig:google_earth_studio_screen}
\end{figure}

\section{The \dataset Benchmark}
\label{sup:dataset}

We rendered images around each landmark using Google Earth Studio, the camera trajectories for each landmark will be published with the benchmark. 
The Google Earth Studio rendering UI is presented at \cref{fig:google_earth_studio_screen}.

After rendering the images, we create a COLMAP using the rendered images of the landmark from Google Earth Studio. We use COLMAP spatial matcher, utilizing the GPS coordinates saved in the rendered image by Google Earth Studio and we configure the mapper with the flag "ignore\_watermarks".

We aligned this model images from the WikiScenes dataset, we chose only images in the exterior category for each landmark. The images are mostly not registered correctly with the default COLMAP parameters, for each landmark we manually found the best parameters, presented in \cref{table:benchmark_parameters}. Then we manually removed images that were not registered correctly.
The benchmark is described in \cref{table:benchmark_description}.
\begin{table}
 \def\arraystretch{0.95}
\centering
\resizebox{\linewidth}{!}{%
\begin{tabular}{llcccccccccccc}
\toprule
WikiScenes ID & Name & Required Matches \\
\midrule
0 & Milano Cathedral & 13 \\
10 & San Francis & 13 \\
18 & Lisbon Cathedral & 30 \\
20 & Brussels Cathedral  & 12 \\
21 & Salamanca Cathedral & 17 \\
27 & St John the Diving Cathedral & 15 \\
36 & Burgos Cathedral & 17 \\
37 & Aachen Cathedral & 13 \\
39 & Freiburg Cathedral & 15 \\
43 & Oveido Cathedral & 13 \\
46 & Palermo Cathedral & 13 \\
50 & Wells Cathedral & 13 \\
52 & Lincoln Cathedral & 13 \\
53 & Monreale Cathedral & 13 \\
54 & Rouen Cathedral & 19 \\
55 & Geneva Cathedral & 14 \\
61 & Bordeaux Cathedral & 14 \\
75 & Murcia Cathedral & 14 \\
85 & Metz Cathedral & 18 \\
90 & ávila Cathedral & 15 \\
93 & Salisbury Cathedral & 30 \\
97 & Napoli Cathedral & 25 \\

\bottomrule
\end{tabular}
}
\caption{\textbf{COLMAP Parameters for \dataset Benchmark Creation}. 
We run COLMAP mapper with two-view tracks, increased triangulation max transitivity (3), increased absolute pose maximal error (36), and increased bundle adjustment maximal refinement range (0.0015). Additionally we run the mapper with varying number of Required Matches described in the table.
}
\label{table:benchmark_parameters}
\end{table}

\begin{table}[ht]
\centering
\resizebox{\linewidth}{!}{%
    \begin{filecontents*}{tables/supp/benchmark_description.csv}
WikiScenes ID,Name,Meta Image ID,Meta Image Size
0,Milano Cathedral,3,19
0,Milano Cathedral,30,713
10,Sanfrancisco Cathedral,0,21
18,Lisbon Cathedral,0,289
20,Brussels Cathedral ,16,8
20,Brussels Cathedral ,15,12
20,Brussels Cathedral ,0,168
20,Brussels Cathedral ,13,12
20,Brussels Cathedral ,14,12
21,Salamanca Cathedral,1,342
27,St John the Divine Cathedral,0,33
36,Burgos Cathedral,10,24
37,Aachen Cathedral,7,23
39,Basel Minster,4,53
43,Oviedo Cathedral,1,73
46,Palermo Cathedral,1,33
46,Palermo Cathedral,0,188
50,Wells Cathedral,1,141
52,Lincoln Cathedral,2,125
52,Lincoln Cathedral,11,14
53,Monreale Cathedral,0,35
54,Rouen Cathedral,0,184
55,Geneva Cathedral,2,26
56,Ulm Minster,1,122
61,Bordeaux Cathedral,3,20
61,Bordeaux Cathedral,0,39
75,Murcia Cathedral,0,52
85,Metz Cathedral,2,59
85,Metz Cathedral,1,40
90,Avila Cathedral,1,110
93,salisbury cathedral,4,131
97,Napoli Cathedral,1,43
\end{filecontents*}
\pgfplotstabletypeset[
        col sep=comma, %
        header=true, %
        string type, %
        every head row/.style={before row=\hline, after row=\hline}, 
        every last row/.style={after row=\hline} 
    ]{tables/supp/benchmark_description.csv}
}
\caption{\textbf{Benchmark Scenes}. The \dataset{} benchmark consists of 32 meta-images from 23 different landmarks found in the WikiScenes~\cite{wu2021towers} dataset, as detailed above. }
\label{table:benchmark_description}
\end{table}

To compare meta-image alignment to the benchmark, the meta-image camera intrinsics must match the intrinsic on the benchmark. To enable evaluation, we forced the intrinsics of the benchmark's cameras on the meta images by rebuilding the meta image.

\end{document}